\documentclass[10pt,twocolumn,letterpaper]{article}

\usepackage{cvpr}
\usepackage{times}
\usepackage{epsfig}
\usepackage{graphicx}
\usepackage{amsmath}
\usepackage{amssymb}
\usepackage{amsfonts}
\usepackage{booktabs}
\usepackage{paralist}
\usepackage{multirow}
\usepackage{array}

\usepackage[pagebackref=true,breaklinks=true,letterpaper=true,colorlinks,bookmarks=false]{hyperref}

\cvprfinalcopy 

\def\cvprPaperID{283} 

\ifcvprfinal\fi

\newcommand{\secref}[1]{Section~\ref{sec:#1}}
\newcommand{\figref}[1]{Figure~\ref{fig:#1}}
\newcommand{\tabref}[1]{Table~\ref{tab:#1}}

\newcommand{\specialcell}[2][c]{%
    \begin{tabular}[#1]{@{}c@{}}#2\end{tabular}}

\graphicspath{{figs/}}
\begin{document}

\title{Predicting Ground-Level Scene Layout from Aerial Imagery}

\author{Menghua Zhai \quad Zachary Bessinger \quad Scott Workman
\quad Nathan Jacobs\\ Computer Science, University of Kentucky\\
{\tt\small \{ted, zach, scott, jacobs\}@cs.uky.edu}}


\maketitle

\begin{abstract} 

  We introduce a novel strategy for learning to extract semantically
  meaningful features from aerial imagery.  Instead of manually
  labeling the aerial imagery, we propose to predict (noisy) semantic
  features automatically extracted from co-located ground imagery. Our
  network architecture takes an aerial image as input, extracts
  features using a convolutional neural network, and then applies an
  adaptive transformation to map these features into the ground-level
  perspective.  We use an end-to-end learning approach to minimize the
  difference between the semantic segmentation extracted directly from
  the ground image and the semantic segmentation predicted solely
  based on the aerial image.
  
  We show that a model learned using this strategy, with no additional
  training, is already capable of rough semantic labeling of aerial
  imagery. Furthermore, we demonstrate that by finetuning this model
  we can achieve more accurate semantic segmentation than two baseline
  initialization strategies. We use our network to address the task of
  estimating the geolocation and geoorientation of a ground image.
  Finally, we show how features extracted from an aerial image can be
  used to hallucinate a plausible ground-level panorama.

\end{abstract}

\section{Introduction}

\begin{figure}[t]
	\centering
	\includegraphics[width=\linewidth]{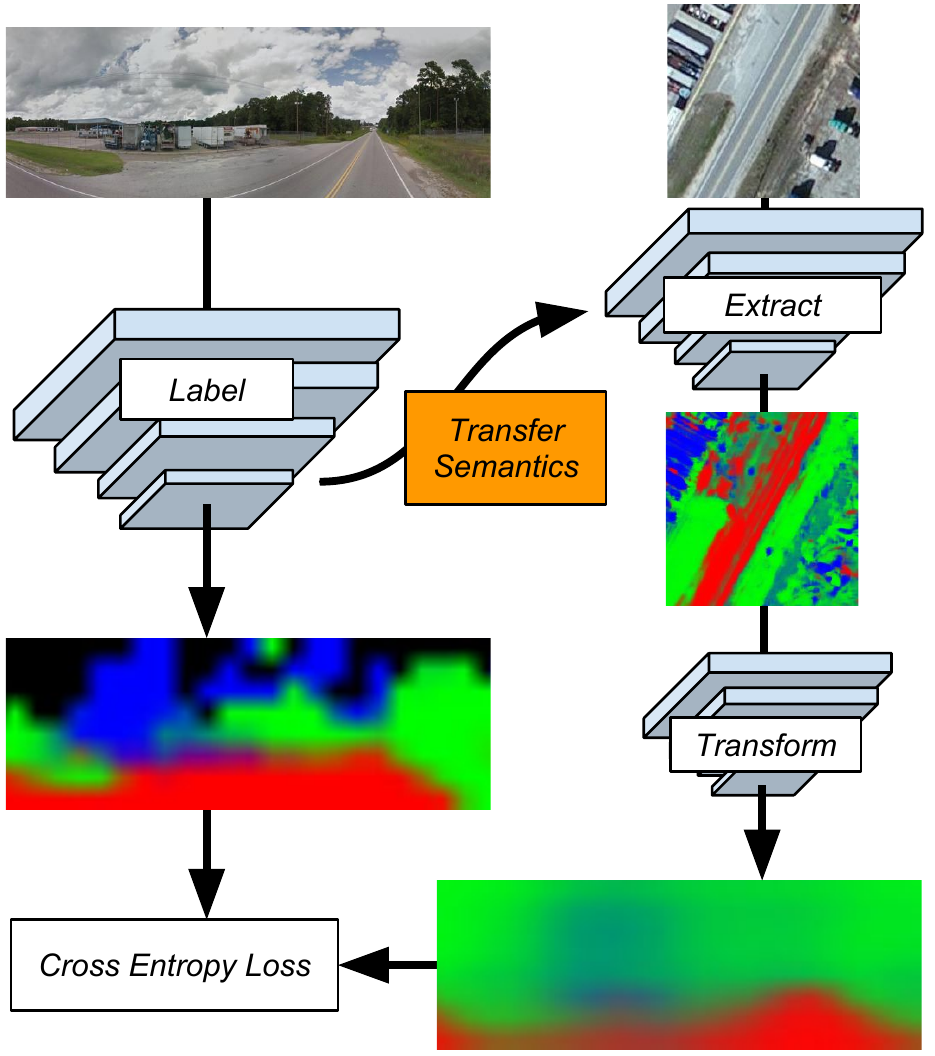}
  \caption{We learn to predict the ground-image
    segmentation directly from an aerial image of the same location,
  thereby transferring the semantics from the ground to the aerial
image domain.}
  %
	\label{fig:overview}
\end{figure}

%
%
%
%

Learning-based methods for pixel-level labeling of aerial imagery
have long relied on manually annotated training data. Unfortunately,
such data is expensive to create. Furthermore, its value is limited
because a method trained on one dataset will typically not perform
well when applied to another source of aerial imagery.  The
difficulty in obtaining datasets of sufficient scale for all
modalities has hampered progress in applying deep learning techniques
to aerial imagery.  There have been a few notable
exceptions~\cite{mnih2010learning,paisitkriangkrai2015effective}, but
these have all used fairly coarse grained semantic classes, covered a
small spatial area, and are limited to modalities in which human
annotators are able to manually assign labels.

We propose a novel strategy for obtaining semantic labels for
aerial image segmentation.  See \figref{overview} for a schematic
overview of the approach.  Our idea is to use existing methods for
semantic image segmentation, which are tailored for ground
images, and apply these to a large dataset of geo-tagged ground
images.  We use these semantically labeled images as a form of weak
supervision and attempt to predict these semantic labels from an
aerial image centered around the location of the ground
image.  We do not use a parametric transformation between the
aerial and ground-level viewpoints.  Instead, we use a dense
representation, similar in spirit to the general representation,
dubbed filter flow, described by Seitz and Baker~\cite{filterflow}.  

There has been significant interest recently in predicting
ground image features from aerial imagery for the task of
ground image geolocalization~\cite{workman2015wide}. Our work is
unique in that it is the first to attempt to predict a dense
pixel-level segmentation of the ground image. We demonstrate the
value of this approach in several ways. 

\paragraph{Main Contributions:} The main contributions of this work
are: (1)  a novel convolutional neural network (CNN) architecture that
relates the appearance of a aerial image appearance to the semantic
layout of a ground image of the same location, (2) demonstrating
the value of our training strategy for pre-training a CNN to
understand aerial imagery, (3) extensions of the proposed technique to
the tasks of ground image localization, orientation estimation,
and synthesis, and (4) an extensive evaluation of each of these
techniques on large, real-wold datasets.  Together these represent an
important step in enabling deep learning techniques to be extended to
the domain of aerial image understanding. 

\begin{figure*}
	\centering
	\includegraphics[width=\linewidth]{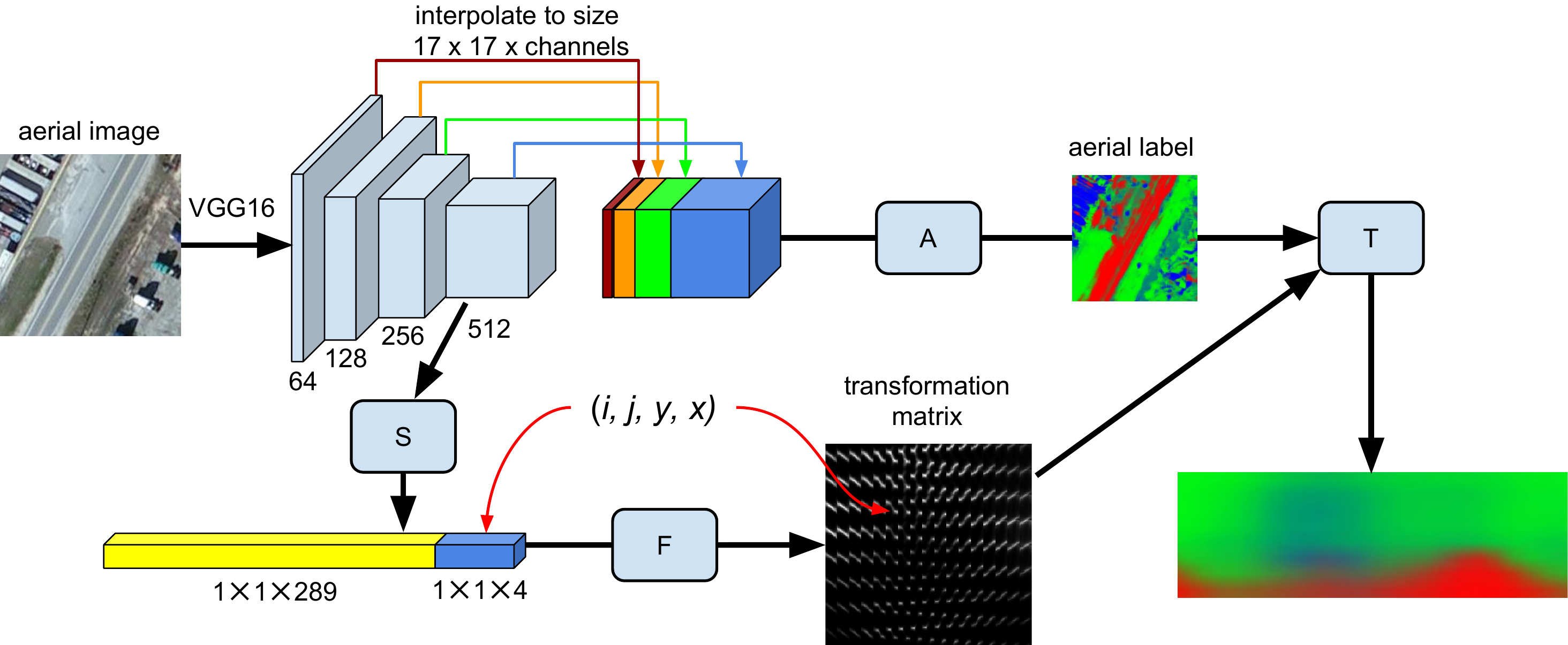}

  \caption{A visual overview of our network architecture. We extract
    features from an aerial image using the VGG16 architecture and
    form a hypercolumn using the PixelNet approach. These features are
    processed by three networks that consist of 1 $\times$ 1
    convolutions: network $A$ converts the hypercolumn into semantic
    features; network $S$ extracts useful features from the aerial
    image for controlling the transformation; and network $F$ 
    defines the transformation between viewpoints.  The
  transformation is applied, $T$, to the aerial semantic features to
create a ground-level semantic labeling.  }

	\label{fig:architecture}
\end{figure*}

\section{Related Work}

\paragraph{Learning Viewpoint Transformations}
Many methods have been proposed to represent the relationship between
the appearance of two viewpoints. Seitz and Baker~\cite{filterflow}
model image transformations using a space-variant linear filter,
similar to a convolution but varying per-pixel. They highlight that a
linear transformation of a vectorized representation of all the pixels
in an image is very general; it can represent all standard parametric
transformations, such as similarity, affine, perspective, and more.
More recently, Jaderberg \etal~\cite{jaderberg2015spatial} describe an
end-to-end learnable module for neural networks, the spatial
transformer, which allows explicit spatial transformations (\eg
scaling, cropping, rotation, non-rigid deformation) of feature maps
within the network that are conditioned on individual data samples.
Practically, including a spatial transformer allows a network to
select regions of interest from an input and transform them to a
canonical pose. Similarly, Tinghui \etal~\cite{tinghui2016flow}
address the problem of novel view synthesis. They observe that the
visual appearance of different views is highly correlated and propose
a CNN architecture for estimating appearance flows, a representation
of which pixels in the input image can be used for reconstruction. 

\paragraph{Relating Aerial and Ground-Level Viewpoints}
Several methods have been recently proposed to jointly reason about
co-located aerial and ground image pairs. Luo
\etal~\cite{luo2008event} demonstrate that aerial imagery can aid
in recognizing the visual content of a geo-tagged ground image.
M{\'a}ttyus \etal~\cite{mattyus2016hd} perform joint inference over
monocular aerial imagery and stereo ground images for fine-grained
road segmentation. Wegner \etal~\cite{wegner2016cataloging} build a
map of street trees. Given the horizon line and the camera intrinsics,
Ghouaiel and Lef{\`e}vre~\cite{ghouaiel2016coupling} transform
geo-tagged ground-level panoramas to a top-down view to enable
comparisons with aerial imagery for the task of change detection.
Recent work on cross-view image
geolocalization~\cite{lin2013cross,lin2015learning,workman2015geocnn,workman2015wide}
 has shown that convolutional neural
networks are capable of extracting features from aerial imagery
that can be matched to features extracted from ground imagery.
Vo \etal~\cite{vo2016localizing} extend this line of work,
demonstrating improved geolocalization performance by applying an
auxiliary loss function to regress the ground-level camera orientation
with respect to the aerial image. To our knowledge, our work is the
first work to explore predicting the semantic layout of a ground
image from an aerial image.

\paragraph{Semantic Segmentation of Aerial/Satellite Imagery}

There is a long tradition of using computer vision techniques for
aerial and satellite image
understanding~\cite{Huertas1988DetectingBI,Willuhn1996ARS,Cheng2016ASO}.
Historically these two domains were distinct.  Satellite imagery was
typically lower-resolution, from a strictly top-down view, and with a
diversity of spectral bands.  Aerial imagery was typically
higher-resolution, with a greater diversity of viewing angles, but
with only RGB and NIR sensors.  Recently these two domains have
converged; we will use the term aerial imagery as we are primarily
working with high-resolution RGB imagery.  However, our approach could
be applied to many types of aerial and satellite imagery.  Kluckner 
\etal~\cite{Kluckner2009SemanticCI} address the task of semantic
segmentation using a random forest to combine color and height
information.  More recent work has explored the use of CNNs for aerial
image understanding. Mnih and Hinton propose a CNN for detecting roads
in aerial imagery~\cite{mnih2010learning} using GIS data as ground
truth.  They extend their approach to handle omission noise and
misregistration between the imagery and the
labels~\cite{mnih2012learning}. These approaches require either
extensive pixel-level manual annotation or existing GIS data. Our work
is the first to demonstrate the ability to transfer a dense
pixel-level labeling of ground imagery to aerial imagery.

\paragraph{Visual Domain Adaptation}
Domain adaptation addresses the misalignment of source and target
domains~\cite{daume2006domain}. A significant amount of work has
explored domain adaptation for visual
recognition~\cite{patel2015visual}. Jhuo \etal~\cite{jhuo2012robust}
propose a low-rank reconstruction approach where the source features
are transformed to an intermediate representation in which they can be
linearly reconstructed by the target samples. Our work is most similar
to that of Sun \etal ~\cite{sun2016unsupervised}, who propose a method
for transferring scene categorizations and attributes from ground
images to aerial imagery. Similar to our approach, they learn a
transformation matrix which minimizes the distance between a source
feature and the target feature. Our work differs in several ways: 1)
we carry out the linear transformation not only in the semantic
dimensions but also in the spatial dimensions, 2) we constrain the
transformation matrix such that the semantic meaning of the source
feature and the target feature remains the same, 3) our transformation
matrix is input dependent, and 4) we learn the transformation matrix
as well as the source feature at the same time, in an end-by-end
manner, which simplifies training. 

\section{Cross-view Supervised Training}
\label{sec:method}

We propose a novel training strategy for learning to extract useful
features from aerial imagery.  The idea is to predict the semantic
scene layout, $L_g$, of a ground image, $I_g$, using only an aligned
aerial image, $I_a$, from the same location.  This strategy leverages
existing methods for ground image understanding at training time, but
does not require any ground imagery at testing time.  

We represent semantic scene layout, $L_g$, as a pixel-level
probability distribution over classes, such as {\em road}, {\em
vegetation}, and {\em building}.  We construct a training pair by
collecting a georegistered ground panorama and an aerial image of the
same location, orienting the panorama to the aerial image (panoramas
are originally aligned with the road direction), and then extracting
the semantic scene layout, $L_g$, of the panorama using an
off-the-shelf method~\cite{badrinarayanan2015segnet} with four
semantic classes.  We then use an end-to-end training strategy to
learn to extract pixel-level features from the aerial image and
transform them to the ground-level viewpoint.  

\subsection{Network Architecture}
\label{sec:architecture}  

Our proposed network architecture is composed of four modules. A
convolutional neural network (CNN), $L_a = A(I_a;\Theta_A)$, is used
to extract semantic labels from the aerial imagery. Another CNN,
$S(I_a;\Theta_S)$, uses features extracted from aerial imagery to help 
estimate the transformation matrix, $M = F(x_r, y_r, i_c, j_c,
S(I_a;\Theta_S); \Theta_F)$, based on aerial image features and the
pixel location in the respective images. Finally, we have a
transformation module, $L_{g'}= T(L_a,M)$, that converts from the
aerial viewpoint to the ground-level using the estimated
transformation matrix, $M$. 
There are many choices for these components, and the 
remainder of this section describes the particular choices we made for this 
study. See~\figref{architecture} for a visual overview of the architecture.

\paragraph{Aerial Image Feature Extraction} For $A(I_a;\Theta_A)$, we
use the VGG16~\cite{simonyan2014very} base architecture and convert it to a 
pixel-level labeling method using the PixelNet approach~\cite{BansalChen16}.  
The core idea is to interpolate intermediate feature maps of the base
network to a uniform size, then concatenate them along the channels dimension 
to form a {\em hypercolumn}.  In our experiments, we form the hypercolumn from 
{\em conv-$\{1_2, 2_2, 3_3, 4_3\}$} of the VGG16
network.  The hypercolumn, which is now 256 $\times$ 256 $\times$ 960,
is followed by three 1 $\times$ 1 convolutional layers, with 512,
512, and 4 output channels respectively. The first two 1 $\times$ 1
convolutions have ReLU activations, the final is linear. We designate
the output of the final convolution as $L_a = A(I_a;\Theta_A)$. The
output of this stage is transformed from a aerial viewpoint to a
ground viewpoint by final stage of the network.
\label{sec:aerialnet}

\paragraph{Cross-view Semantic Transformation} We represent 
the transformation between the aerial and ground-level
viewpoints as a linear operation applied channel-wise to $L_a$.  To
transform from the $h_a \times w_a \times$ 4 aerial label, $L_a$, to
the $h_g \times w_g \times$ 4 ground label, $L_{g'}$, we need to
estimate a $h_gw_g \times h_a w_a$ matrix, $M$. Given $M$, the
transformation process is as follows: reshape the aerial label, $L_a$,
into a $h_aw_a \times $ 4 matrix, $l_a$; multiply it by $M$ to get
$l_{g'}$; then reshape $l_{g'}$ to the size of the ground label,
$L_g$, to form our estimate of the ground label, $L_{g'}$.  We
constrain $M$ such that the sum of each row is unit. To account for
the expected layout of the scene, and to handle the sky class (which
is not visible from the aerial image), we carry out the transformation
on the logits of $l_a$, $f_a$, and add a bias term, $b$ to get the
logits of $l_{g'}$, $f_{g'}$: $f_{g'} = M f_a + b$.

There are many ways of representing the transformation matrix, $M$.
The na\"{\i}ve approach is to treat $M$ as a matrix of learnable
variables. However, this approach has two downsides: (1) the
transformation does not depend on the content of the aerial image and
(2) the number of parameters scales quadratically with the number of
pixels in $L_a$ and $L_g$.  

We represent each element, $M_{rc}$, in the transformation matrix,
$M$, as the output of a neural network, $F$, which is conditioned on
the aerial image, $I_a$, and the location in the input and output
feature maps. More precisely, each element $M_{rc} =
F(x_r,y_r,i_c,j_c,S(I_a;\Theta_S))$, where $(i_c, j_c) \in [0,1]$, is
the aerial image pixel of the corresponding element, $(y_r, x_r)
\in [0,1]$ is the ground image pixel of the corresponding
element.

We now define the architecture of the transformation estimation neural
network, $F$. The value of the transformation matrix at
location $(r, c)$ is computed through a neural network, $\tilde{F}$,
followed by a {\em softmax} function to normalize the impact of all
pixels sampled from the aerial image:
\begin{align*}
\begin{split}
  M_{rc} &= F\left(r, c, S(I_a; \Theta_S)\right)
  = \frac{e^{\tilde{F}_{r,c}}}{\sum_{c'}{e^{\tilde{F}_{r,c'}}}}, \\
\textrm{where:} ~~~~~&\\
  &\tilde{F}_{r, c} = \tilde{F}\left(i, j, y, x, S(I_a; \Theta_S)\right), 
  ~~\textrm{and}\\
  & i = \lfloor c / w_a \rfloor / h_a, ~~~j = \mathrm{mod}(c,w_a) / w_a, \\
  & y = \lfloor r / w_g \rfloor / h_g, ~~~x = \mathrm{mod}(r,w_g) / w_g.
\label{equ:matmul}
\end{split}
\end{align*}
The base network, $\tilde{F}$, is a multi-layer perceptron, with ReLU
activation functions, that takes as input a 293-element vector.
The network has three layers, with 128, 64 and 1 output channels
respectively (refer to the lower part of~\figref{architecture}).  The
na\"{\i}ve approach can be considered a special case of this
representation where we ignore the aerial image and use a one-hot
encoding representation of rows and columns.  


As described above, there are two main advantages of our approach of
representing the transformation matrix: a reduction in the number of
parameters when $M$ is large and the ability to adapt to different
aerial image layouts.  An additional benefit is that if we change the
resolution of our input and output feature maps it is easy to create a
new transformation matrix, $M$, without needing to resort to
interpolation.

\subsection{Dataset}
\label{sec:dataset}

\begin{figure}
	\newlength{\aheight}
	\setlength{\aheight}{1.75cm}
	\newlength{\gwidth}
	\setlength{\gwidth}{.76\linewidth}
	\newlength{\gspace}
	\setlength{\gspace}{.4cm}
	\centering
	\includegraphics[height=\aheight,width=\aheight]{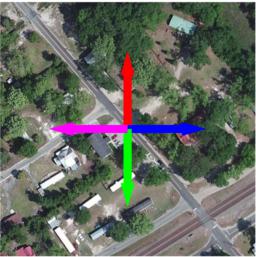}\hspace{6pt}%
	\begin{minipage}[b]{\gwidth}
	\includegraphics[height=1.75cm,width=\textwidth]{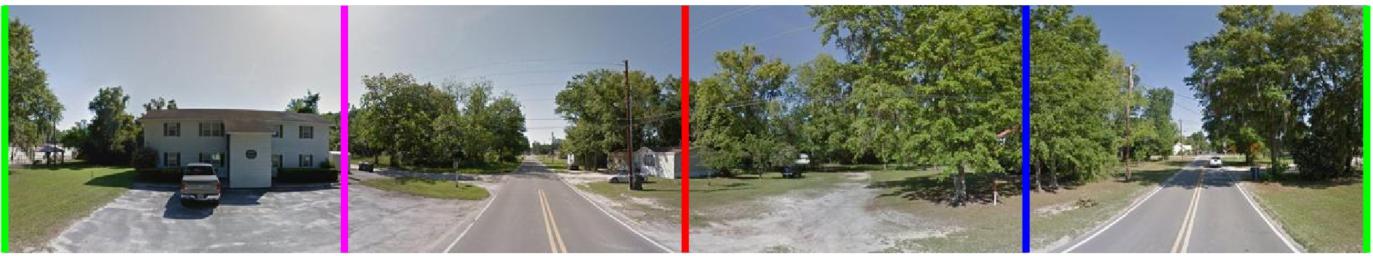}\vfill
	\end{minipage}\vspace{1pt}
	\includegraphics[height=1.75cm,	width=1.75cm]{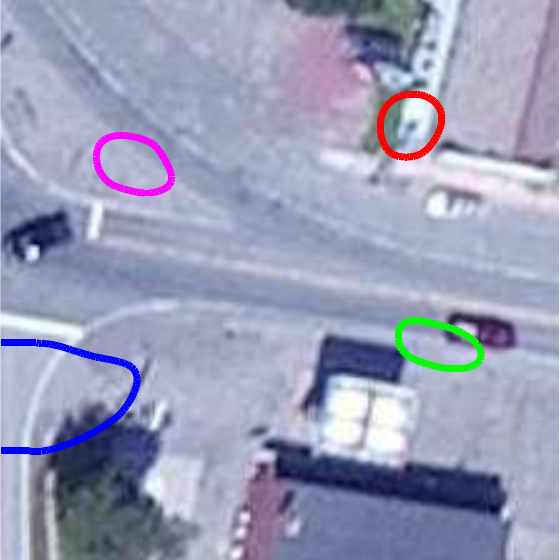}\hspace{6pt}%
	\begin{minipage}[b]{\gwidth}
	\includegraphics[height=1.75cm,width=\textwidth]{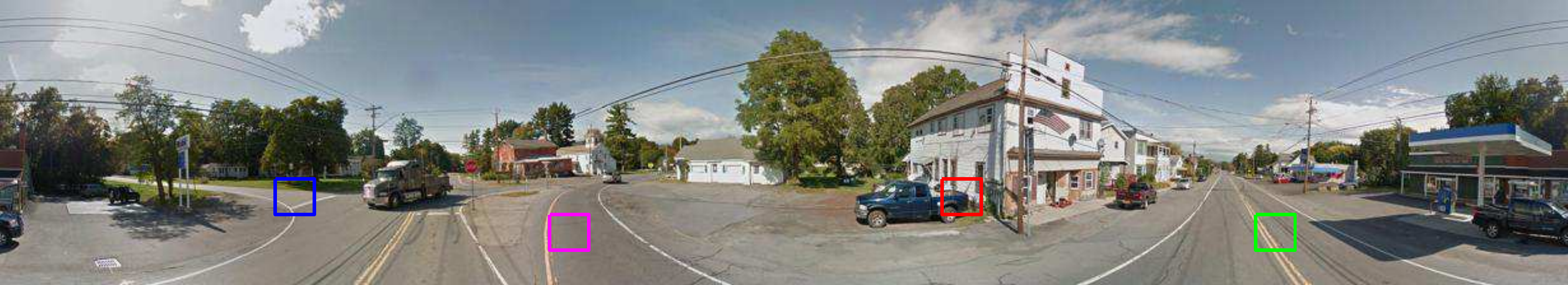}\vfill
	\end{minipage}\vspace{1pt}
	\includegraphics[height=1.75cm,	width=1.75cm]{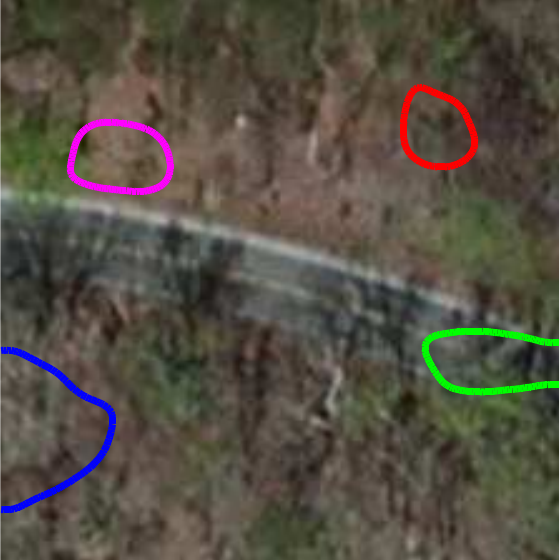}\hspace{6pt}%
	\begin{minipage}[b]{\gwidth}
	\includegraphics[height=1.75cm,width=\textwidth]{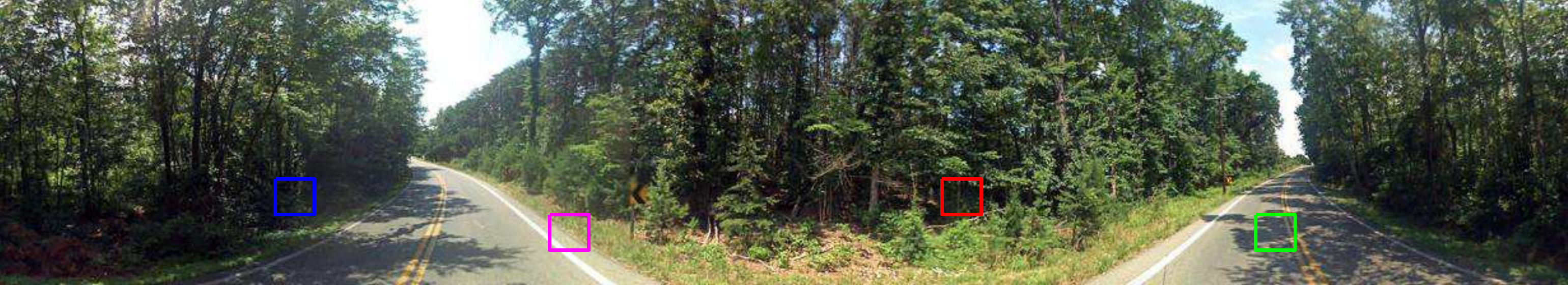}\vfill
	\end{minipage}\vspace{1pt}
	\includegraphics[height=1.75cm,	width=1.75cm]{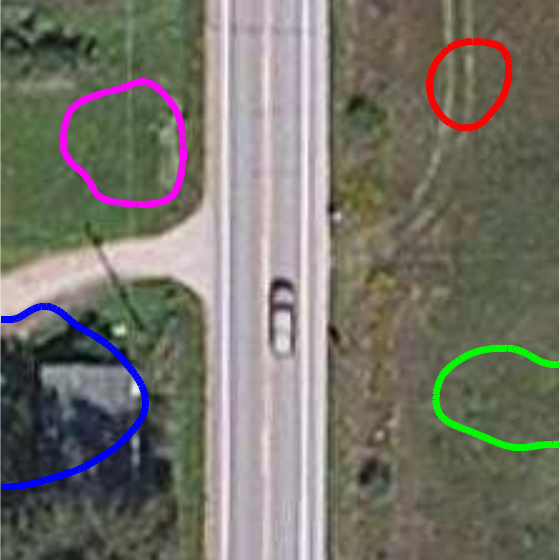}\hspace{6pt}%
	\begin{minipage}[b]{\gwidth}
	\includegraphics[height=1.75cm,width=\textwidth]{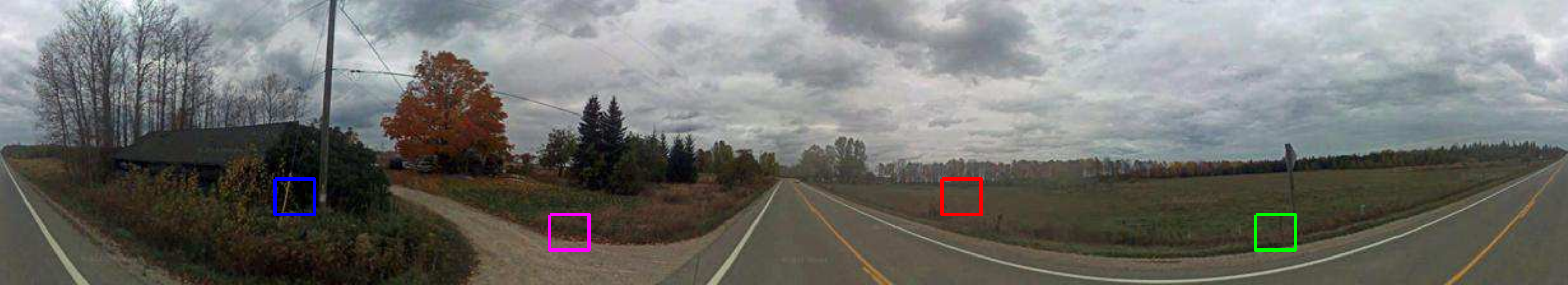}\vfill
	\end{minipage}\vspace{1pt}

  \caption{Examples of aligned aerial/ground image pairs from our
dataset. (row 1) In the aerial images, north is the up direction.  In
the ground images, north is the central column. (row 2-4) Image
dependent receptive fields estimated by our algorithm as follows: 1)
fix ground locations $(y, x)$ (locations in squares); 2) select all
$(i, j)$ (locations in contours) with high $\tilde{F}(i,j,y,x,
S(I_a; \Theta_S))$ values.  Corresponding fields between the aerial image and
the ground image are shown in the same color.}

  \label{fig:dataset}
\end{figure}

We collect our training and testing dataset from the CVUSA
dataset~\cite{workman2015wide}. CVUSA contains approximately 1.5 million
geo-tagged pairs of ground and aerial images from across the
United States. We use the Google Street View panoramas of CVUSA as our
ground images. For each panorama, we also
download an aerial image at zoom level 19 from Microsoft Bing Maps 
in the same location. We filter out panoramas with no available 
corresponding aerial imagery. Using the camera's extrinsic
information, we then warp the panoramas to align with the
aerial images.  We also crop the panoramas vertically 
to reduce the portion of the sky and
ground pixels.  In total, we collected 35,532 image pairs for
training and 8,884 image pairs for testing. Some examples aerial/ground 
image pairs in our dataset are shown \figref{dataset}. 

\subsection{Implementation Details}
\label{sec:details}

We implemented the proposed architecture using Google's TensorFlow
framework~\cite{abadi2016tensorflow}. We train our networks for 10
epochs with the Adam optimizer~\cite{kingma2014adam}. We enable batch
normalization~\cite{ioffe2015batch} with decay 0.9 in all convolutional and 
fully-connected layers (except for the output layers) to accelerate
the training process. Our implementation is available at
\url{https://github.com/viibridges/crossnet}.

The training procedure is as follows: for a given cross-view image
pair, $(I_a, I_g)$, we first compute for the ground semantic pixel
label: $I_g \rightarrow L_g$, using
SegNet~\cite{badrinarayanan2015segnet}. We then minimize the cross
entropy between $L_g$ and $T(A(I_a;\Theta_A);\Theta_T)$ with respect
to the model parameters, $\Theta_A$ and $\Theta_T$.  The resulting
architecture requires a significant amount of memory to output the
full final feature map, which would normally result in very small
batch sizes for GPU training.  Due to the PixelNet approach of using
interpolation to scale the feature maps, we are able to perform sparse
training. Instead of outputting the full-size feature map, we only
extract a dense grid of points, the resulting feature map is 17
$\times$ 17 $\times$ 4.  Despite this, at testing time,  we can provide
an aerial image and generate a full-resolution, semantically meaningful
feature map.



\section{Evaluation and Applications}
\label{sec:evaluation}
In this section, we will show that our network architecture can be
used in four different tasks: 1) weakly supervised semantic learning, 
2) aerial imagery labeling, 3) orientation regression and geocalibration, 
and 4) cross-view image synthesis. Additional qualitative results and 
the complete network structure used for cross-view image synthesis 
can be found in our supplemental materials.

\subsection{Weakly Supervised Learning}
\label{sec:transfer}

\begin{figure}
	\centering
	\includegraphics[width=.19\linewidth]{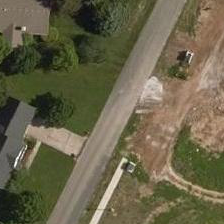} \hfill
	\includegraphics[width=.19\linewidth]{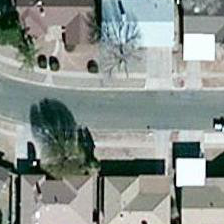} \hfill
	\includegraphics[width=.19\linewidth]{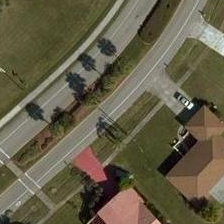} \hfill
	\includegraphics[width=.19\linewidth]{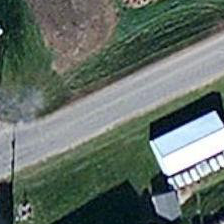} \hfill
	\includegraphics[width=.19\linewidth]{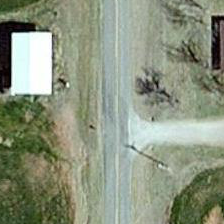}
  \newline
	\includegraphics[width=.19\linewidth]{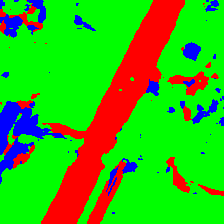} \hfill
	\includegraphics[width=.19\linewidth]{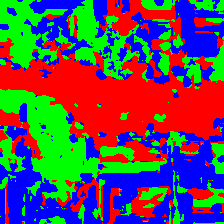} \hfill
	\includegraphics[width=.19\linewidth]{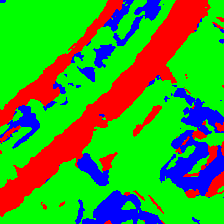} \hfill
	\includegraphics[width=.19\linewidth]{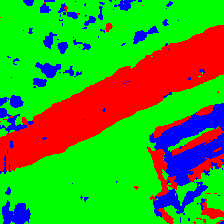} \hfill
	\includegraphics[width=.19\linewidth]{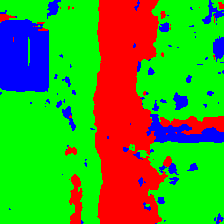}
  \caption{Example outputs from our weakly supervised learning method
    on test images.  For each aerial image (top), we show the
    pixel-level labeling inferred by our model, which uses only noisy
    ground image segmentation as labels.  We visualize three classes:
    {\em road} (red), {\em vegetation} (green), and {\em man-made} 
    (blue).}
  \label{fig:weakly}
\end{figure}

We trained our full network architecture (with randomly initialized
weights) to predict ground-level semantic labeling using the dataset
described in \secref{dataset}.  \figref{weakly} shows example output,
$L_a$, from the aerial image understanding CNN. This demonstrates that
the resulting network has learned to extract semantic features from an
aerial image, all without any manual annotated aerial imagery.

While these results are compelling, they could be better with a higher
quality ground-image segmentation method.  The method we use,
SegNet~\cite{badrinarayanan2015segnet}, was trained on mostly urban
scenes, but many of our images are from rural and suburban areas.  The
end result is that certain classes are often mislabeled in the ground
imagery, including {\em dirt} and {\em building}.  In addition,
because the panoramas and aerial images were not captured at the same
time, we are unable to accurately model transient objects, such as
vehicles and pedestrians. All these factors make the dataset very
challenging for training. Given these limitations, it is, perhaps,
surprising that the resulting aerial image segmentation method works
so well.  In the following section, we show using this network as a
starting point for strongly supervised aerial image segmentation
outperforms two standard initialization methods. 

\subsection{Cross-view for Pre-training}

We evaluate our proposed technique as a pre-training strategy for the
task of semantic-pixel labeling of aerial imagery.  Starting from
the optimal weights from the previous section, we finetune and
evaluate using the ISPRS dataset~\cite{rottensteiner2013isprs}.
This dataset contains 33 true orthophotos captured over
Vaihingen, Germany.  The ground sampling distance is 9 cm/px and there
are over 168 million pixels in total.  Ground truth is provided for 16
photos; each pixel is assigned one of six categories: {\em Impervious
surfaces, Building, Low vegetation, Tree, Car, and Clutter/background}.

\begin{figure}
  \newlength{\abc}
  \setlength{\abc}{.325\linewidth}
	\includegraphics[width=\abc]{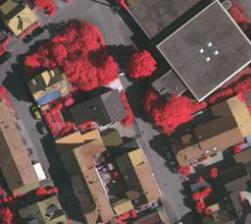}
	\includegraphics[width=\abc]{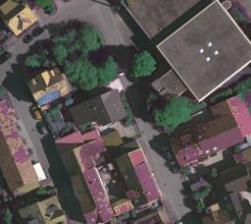}
	\includegraphics[width=\abc]{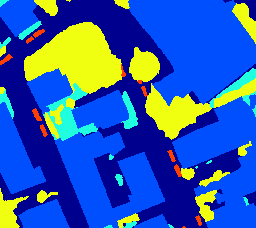}
  \caption{An example from the ISPRS dataset~\cite{rottensteiner2013isprs}. 
  (left) Near infrared image; (middle) The same image after pre-processing; 
  (right) Ground-truth annotation of the image.}
	\label{fig:isprs:dataset}
\end{figure}

\paragraph{Image Processing} Compared to the Bing Maps imagery we used
for pre-training, images in the ISPRS dataset are at a
different spatial scale and the color channels represent different
frequency bands (the R channel is actually a near infrared channel).
To ensure that the pre-trained network weights are appropriate for
the new dataset, we adjusted the scale and color channels as follows.
We first resize the ISPRS images to the equivalent of Bing Maps
zoom level 19.  We then label a pixel as vegetation if
$\frac{R}{R+G+B}$ is greater than 0.4. For each pixel labeled as
vegetation, we halve the R channel intensity and swap the R and G
channels.  The resulting images, shown in~\figref{isprs:dataset}, are
much closer in appearance to the Bing Maps imagery than the raw
imagery.

\paragraph{Evaluation Protocol} We split the 16 annotated images into
training (images 5, 7, 11, 13, 15, 17, and 21), validation sets
(images 1 and 3) and testing (images 23, 26, 28, 30, 32, 37, and 40).
From each set we extracted a set of 224 $\times$ 224 subwindows
(respectively 82, 12, and 34 from training, validation, and testing
respectively).
We then compared performance with different numbers of training
images: 1, 2, 7, 20, 54, and 82. We evaluated the performance in terms
of the average precision for all pixels.  We ignore the
\textit{Cluster/background} pixels because of the low number of
assigned pixels.

\paragraph{Training and Testing} We used the same architecture with
the aerial feature extractor, $A(I_a; \Theta_A)$, defined in
~\secref{aerialnet}, to do the semantic labeling on ISPRS.  During
training, we use the Adam optimizer to minimize the cross entropy
between the network outputs and the labels.  We use a batch size of 8,
randomly sample 1,000 pixels per image for sparse training, and train
the network till convergence.  We run the validation set every 1,000 training
iterations and save the optimal network weights for testing.  During
testing, we sample all pixels on the image to generate the dense
labeling.

We experiment using three different initializations of the VGG16
convolutional layers and finetune the remaining layers of the network: 
1) Ours: initialize with model pre-trained using our framework;
2) Random: initialize using Xavier 
initialization~\cite{xavier2010understanding}; 
3) VGG16: initialize with model pre-trained on ImageNet.

Since the VGG16 model we used in this experiment is trained without
batch normalization, it may be less competitive. To achieve a fair
comparison, we turned off batch normalization in this experiment and
re-trained the network for 15 epochs to get the pre-trained model.

\begin{figure}
	\centering
	\includegraphics[width=.8\linewidth]{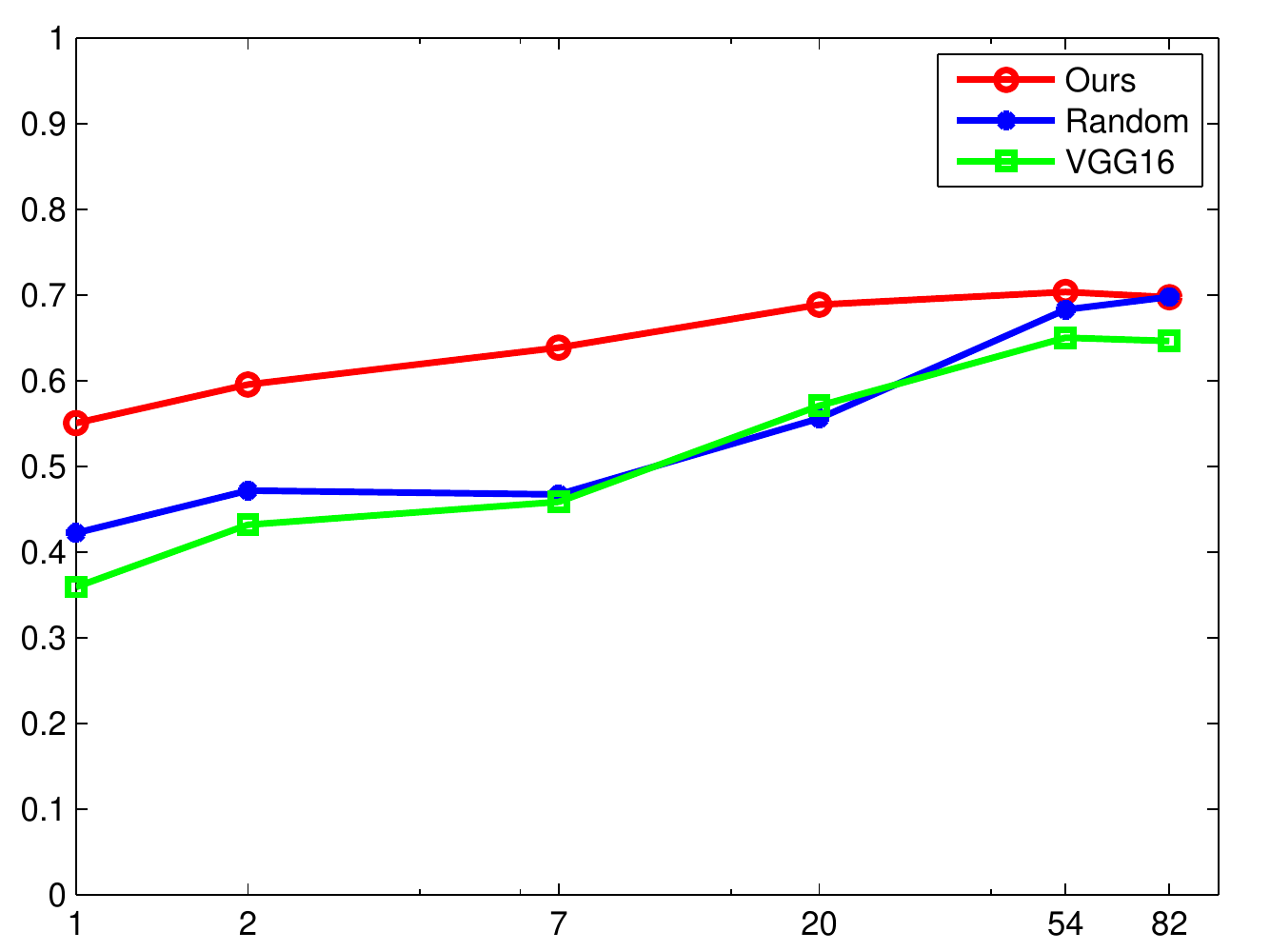}

  \caption{Performance comparison of different initialization methods
  on the ISPRS segmentation task. The $x$-axis is the number of
training images and the $y$-axis is average precision.}

	\label{fig:isprs:precisions}
\end{figure}

Our results (\figref{isprs:precisions}) show that finetuning from the 
VGG16 model performs poorly on the aerial image labeling task. We
think that the patterns it learned mostly from the ground image may
hinder pattern learning for aerial imagery. Our method outperforms
both of the other initialization strategies.  We also present the prediction
precision per class in~\tabref{isprs:precisions}. We highlight that our method
does better especially on the \textit{Building}, \textit{Low Vegetation}, and
\textit{Tree} classes, which can also be found in pre-training
annotations.

\begin{table}
  \centering
  \caption{Per-Class Precision on the ISPRS Segmentation Task}
  \begin{tabular}{@{}llcccccc@{}}
    \toprule
    \multirow{2}{*}{Class} & \multirow{2}{*}{Init.} & \multicolumn{5}{c}{Number 
    of training samples} \\
    \cline{3-8}
                   &       & 1 & 2 & 7 & 20 & 54 & 82\\
    \hline
    \multirow{3}{*}{\textbf{\specialcell{Imp.}}}
    & Ours & 0.67 & \textbf{0.74} & \textbf{0.63} & \textbf{0.64} & 0.66 & 
    0.64  \\ 
    & Random & \textbf{0.70} & 0.70 & 0.54 & 0.62 & 0.61 & \textbf{0.73}  \\ 
    & VGG16 & 0.60 & 0.55 & 0.55 & 0.61 & \textbf{0.70} & 0.59  \\ 
    \hline
    \multirow{3}{*}{\textbf{Bldg}} 
    & Ours & 0.72 & \textbf{0.76} & \textbf{0.76} & \textbf{0.80} & 0.75 & 
    \textbf{0.78}  \\ 
    & Random & 0.56 & 0.62 & 0.63 & 0.64 & \textbf{0.82} & 0.71  \\ 
    & VGG16 & \textbf{0.78} & 0.72 & 0.71 & 0.69 & 0.70 & 0.75  \\ 
    \hline
    \multirow{3}{*}{\textbf{\specialcell{Low.}}}
    & Ours & \textbf{0.37} & \textbf{0.43} & \textbf{0.51} & \textbf{0.65} & 
    \textbf{0.67} & \textbf{0.67}  \\ 
    & Random & 0.29 & 0.29 & 0.29 & 0.37 & 0.67 & 0.64  \\ 
    & VGG16 & 0.25 & 0.25 & 0.29 & 0.44 & 0.53 & 0.57  \\ 
    \hline
    \multirow{3}{*}{\textbf{Tree}} 
    & Ours & \textbf{0.68} & \textbf{0.54} & \textbf{0.71} & \textbf{0.71} & 
    \textbf{0.74} & 0.74  \\ 
    & Random & 0.42 & 0.46 & 0.49 & 0.56 & 0.71 & 0.69  \\ 
    & VGG16 & 0.36 & 0.44 & 0.50 & 0.55 & 0.65 & \textbf{0.74}  \\ 
    \hline
    \multirow{3}{*}{\textbf{Car}} 
    & Ours & \textbf{0.13} & \textbf{0.46} & \textbf{0.67} & \textbf{0.48} & 
    \textbf{0.48} & 0.49  \\ 
    & Random & 0.05 & 0.08 & 0.10 & 0.25 & 0.45 & \textbf{0.57}  \\ 
    & VGG16 & 0.05 & 0.11 & 0.20 & 0.20 & 0.25 & 0.23  \\ 
    \bottomrule
    \end{tabular}
  \label{tab:isprs:precisions}
\end{table}

\subsection{Cross-view for Geocalibration}


We show how the ground-level feature maps we estimate from aerial
imagery can be used to estimate the orientation and location of a
ground image.  We show quantitative results for the orientation
estimation task and qualitative results for simultaneous orientation
and location estimation. We use the following datasets for all
experiments:

\begin{figure*}
  \setlength{\aheight}{63pt}
  \setlength{\gwidth}{52.44pt}
  \centering

  \begin{minipage}[b]{\gwidth}
  \includegraphics[width=\textwidth]{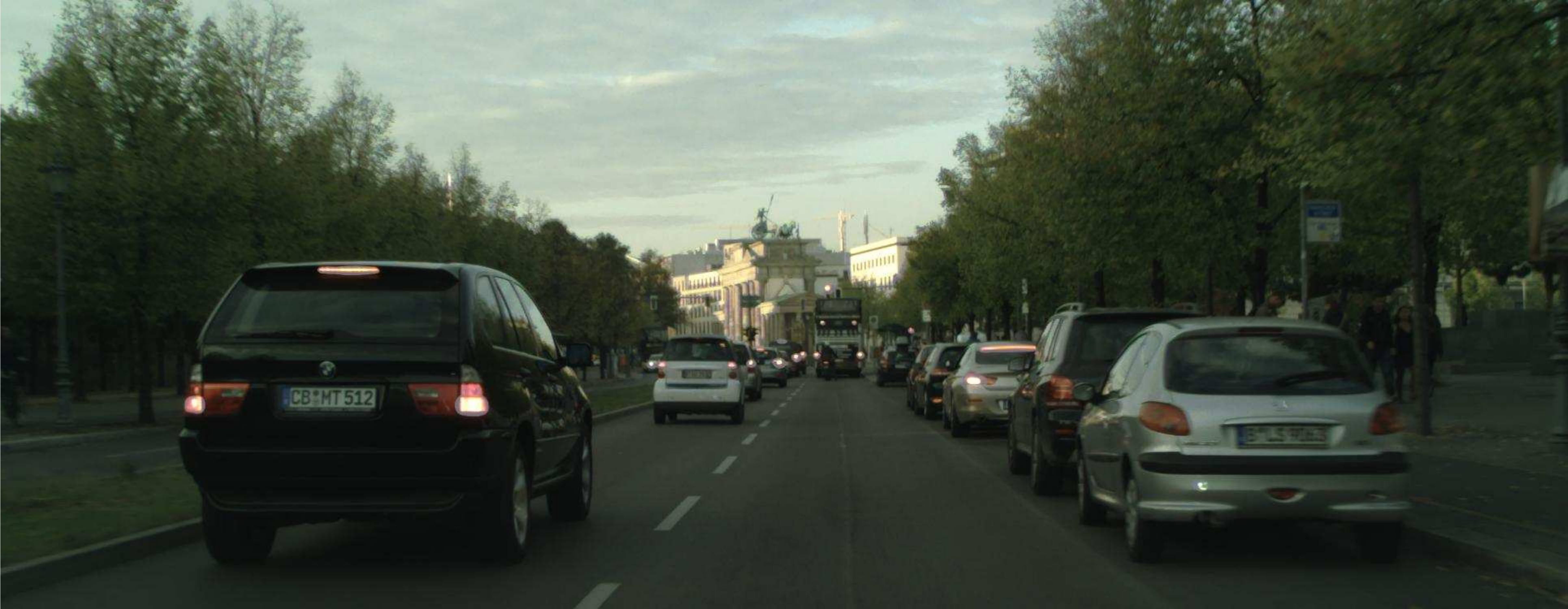}
  \includegraphics[width=\textwidth]{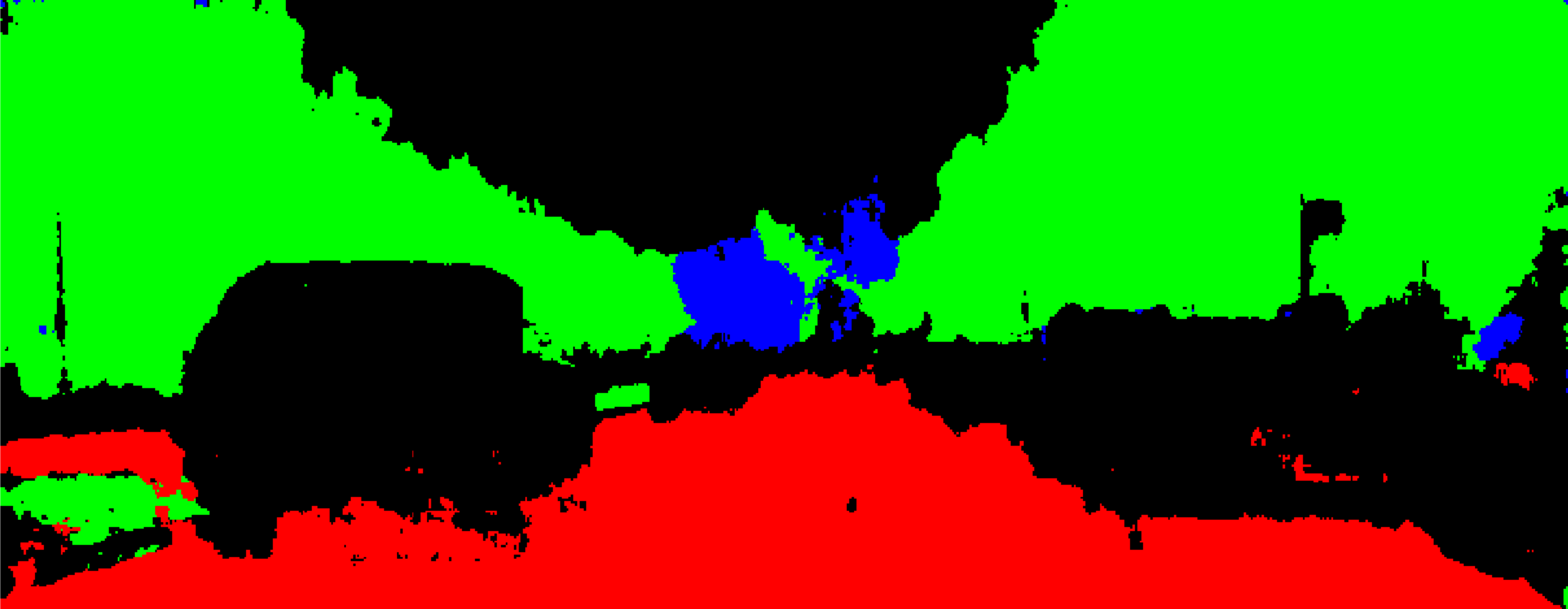}
  \includegraphics[width=\textwidth]{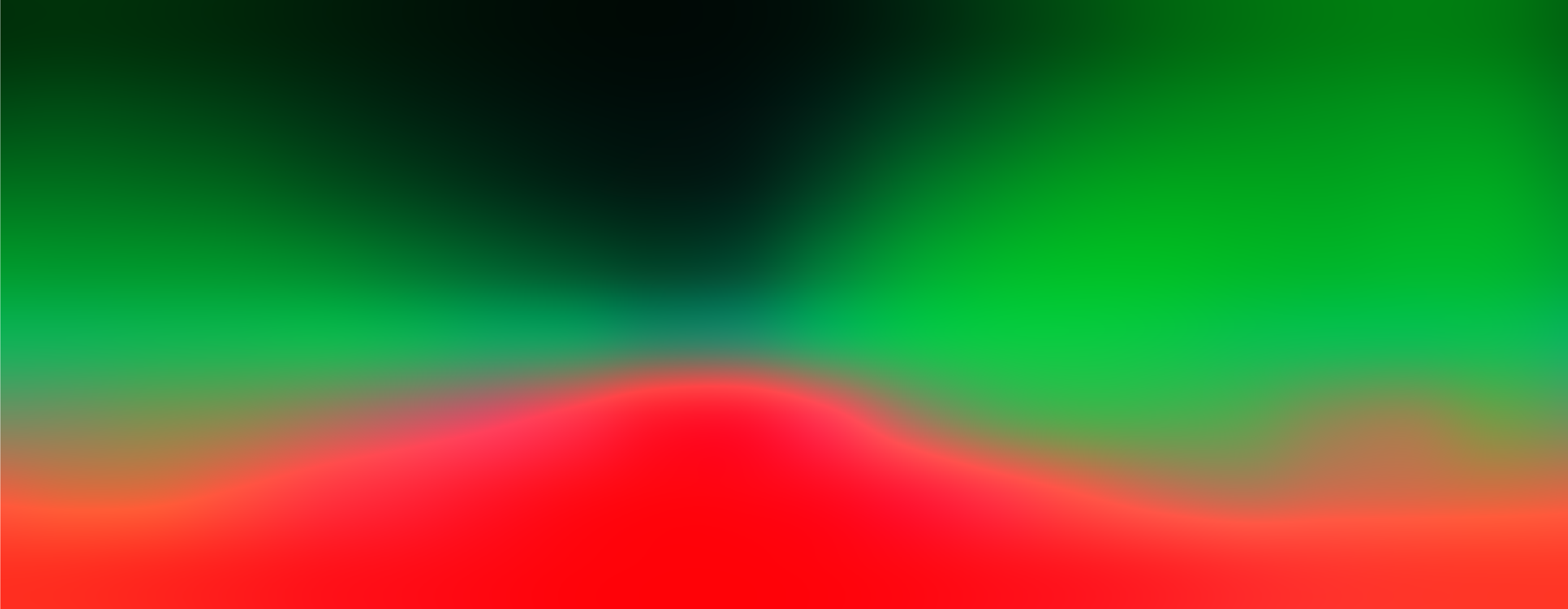}
  \end{minipage}
  \includegraphics[height=\aheight]{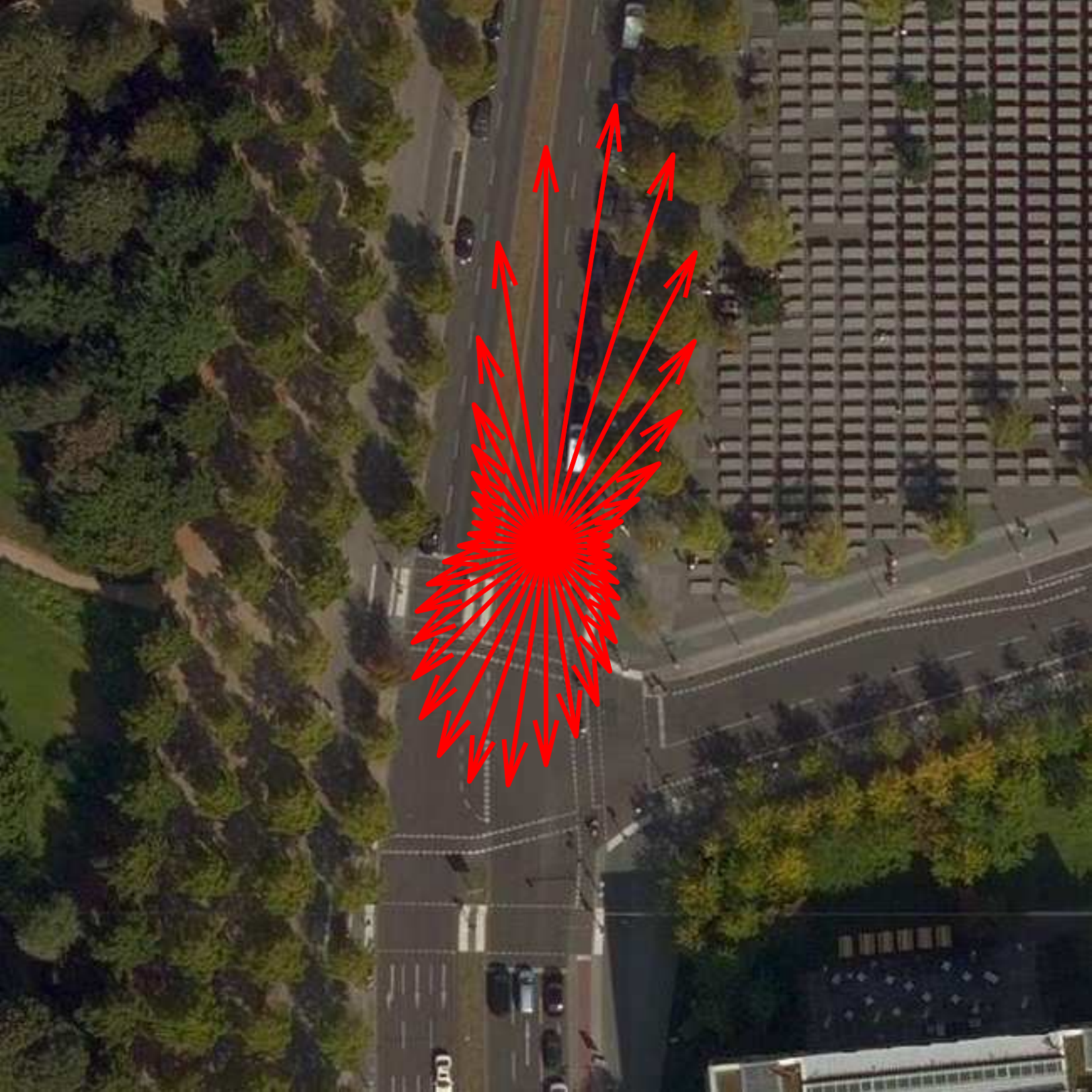}
  \hfill
  \begin{minipage}[b]{\gwidth}
  \includegraphics[width=\textwidth]{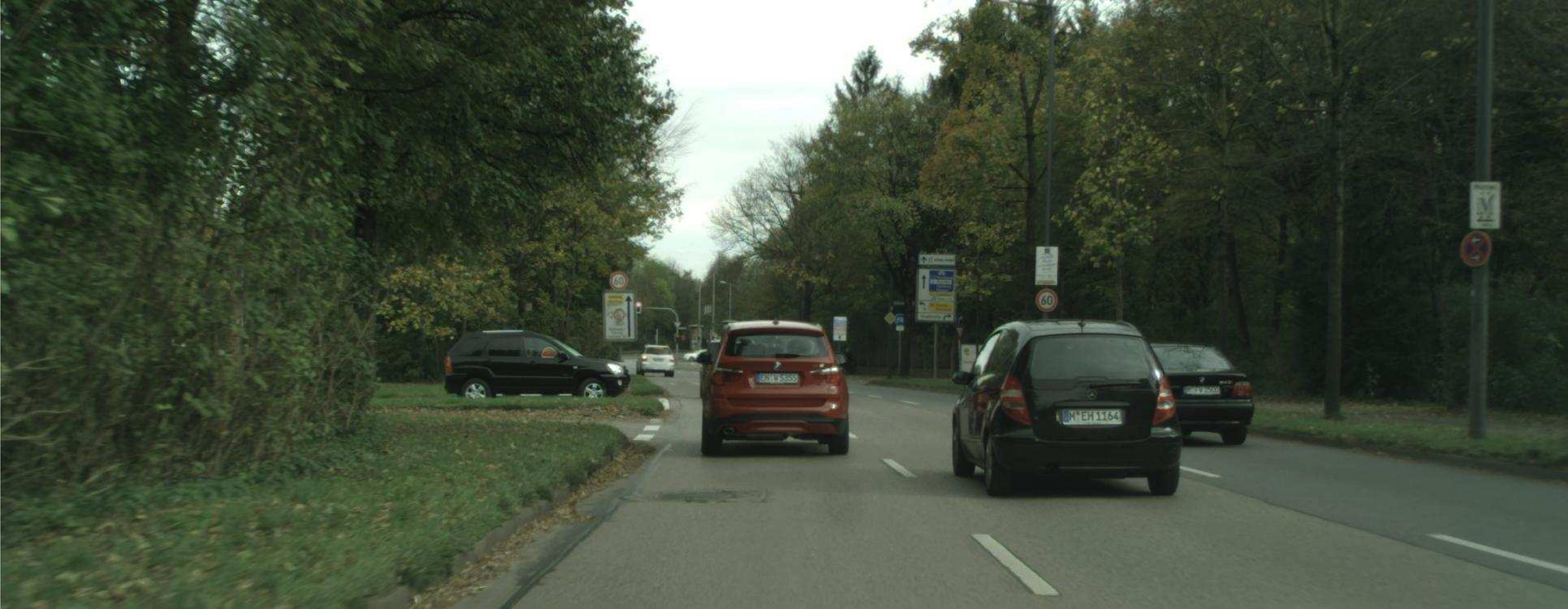}
  \includegraphics[width=\textwidth]{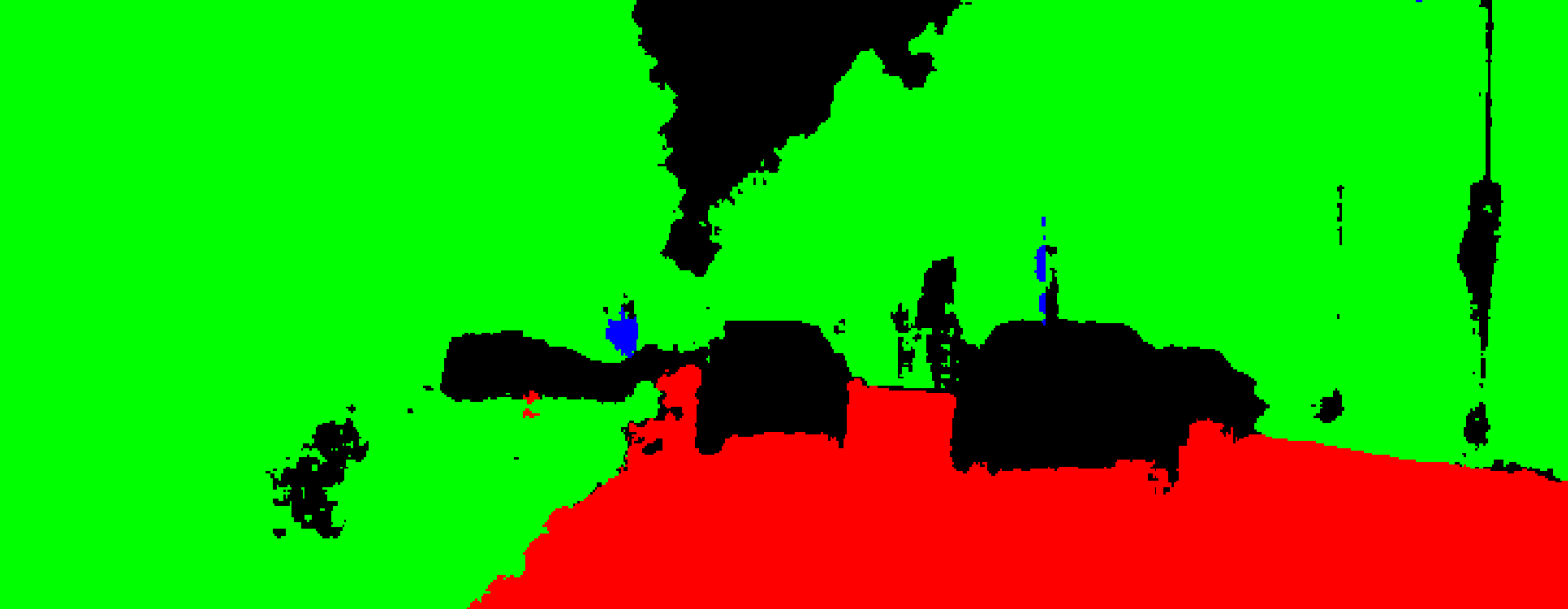}
  \includegraphics[width=\textwidth]{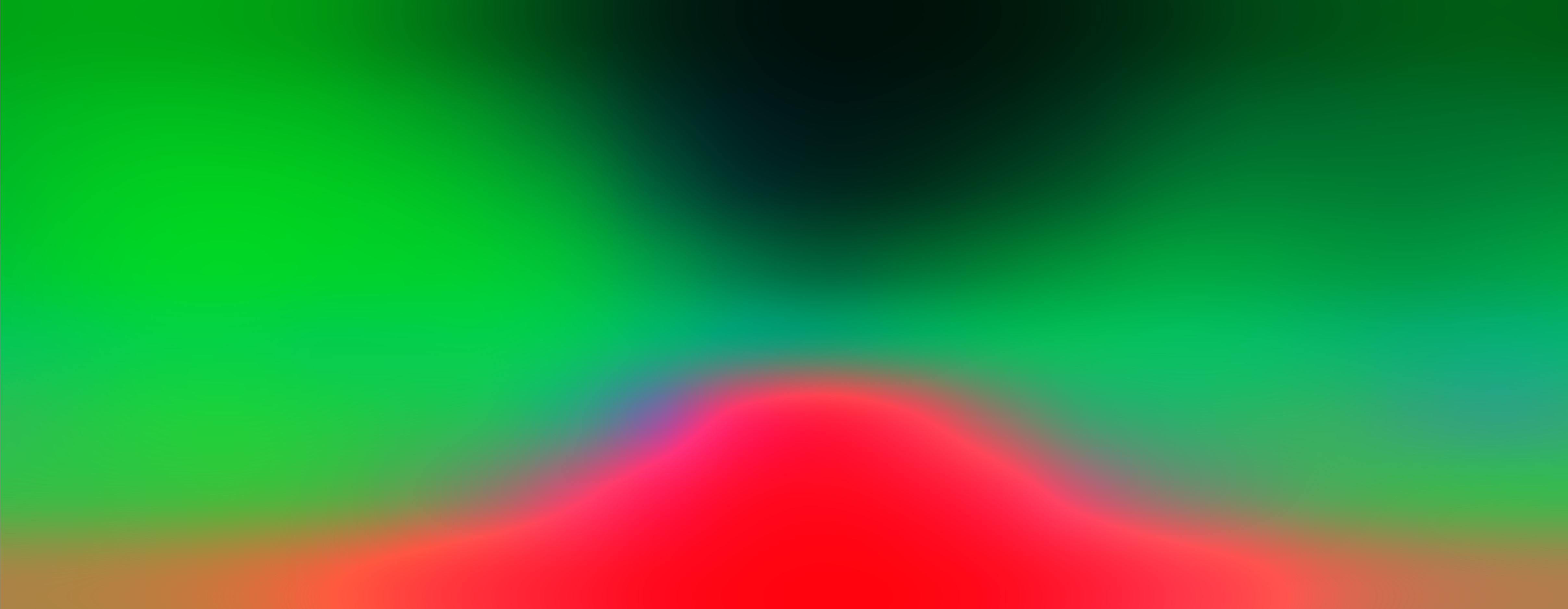}
  \end{minipage}
  \includegraphics[height=\aheight]{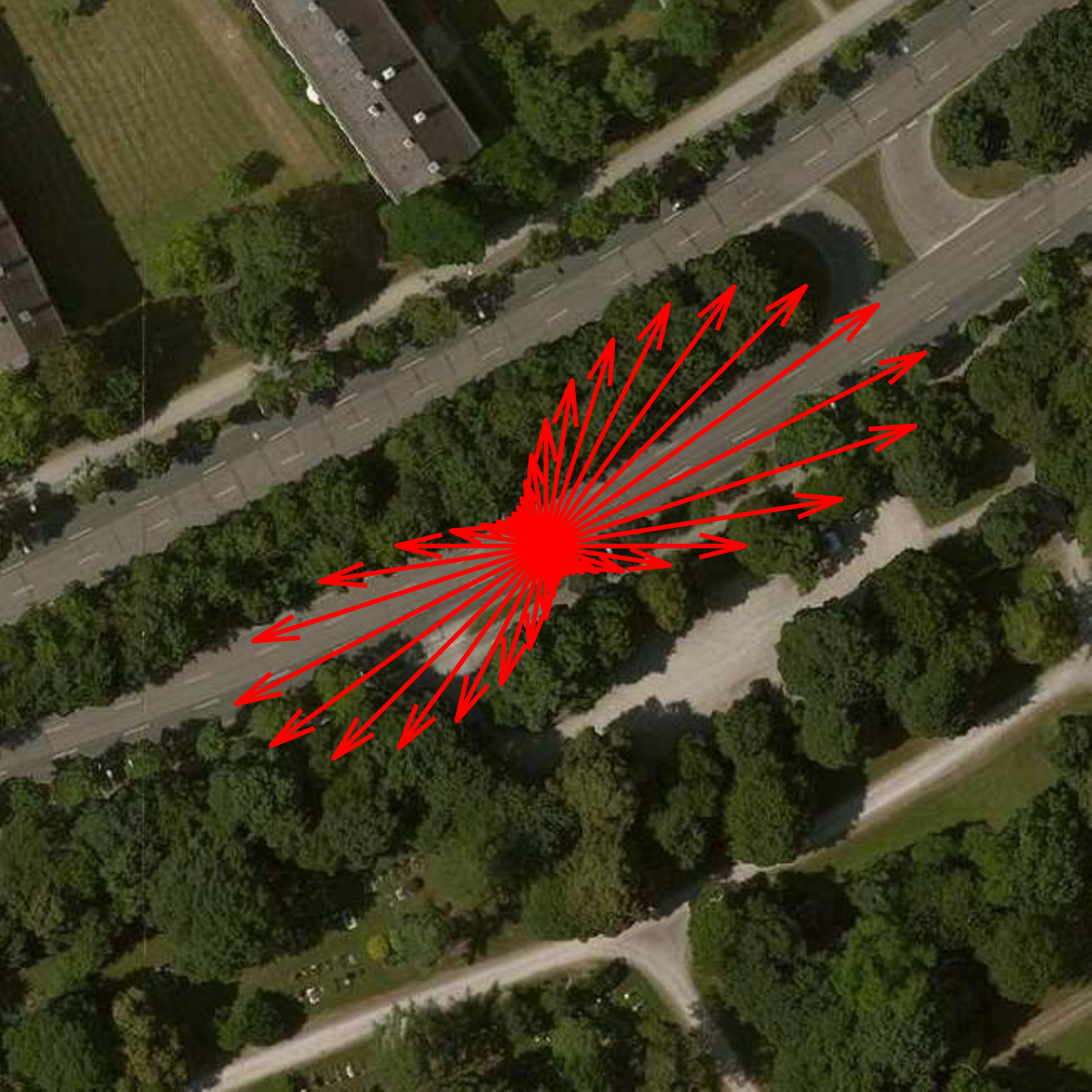}
  \hfill
  \begin{minipage}[b]{\gwidth}
  \includegraphics[width=\textwidth]{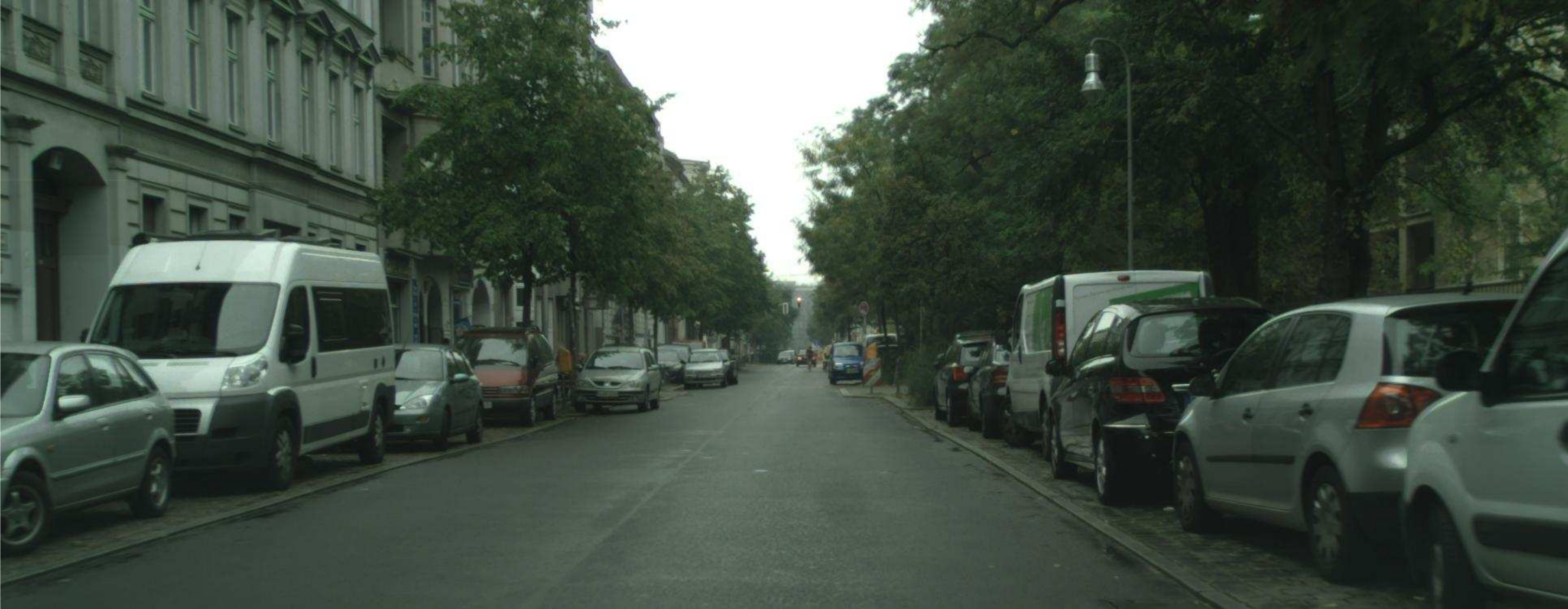}
  \includegraphics[width=\textwidth]{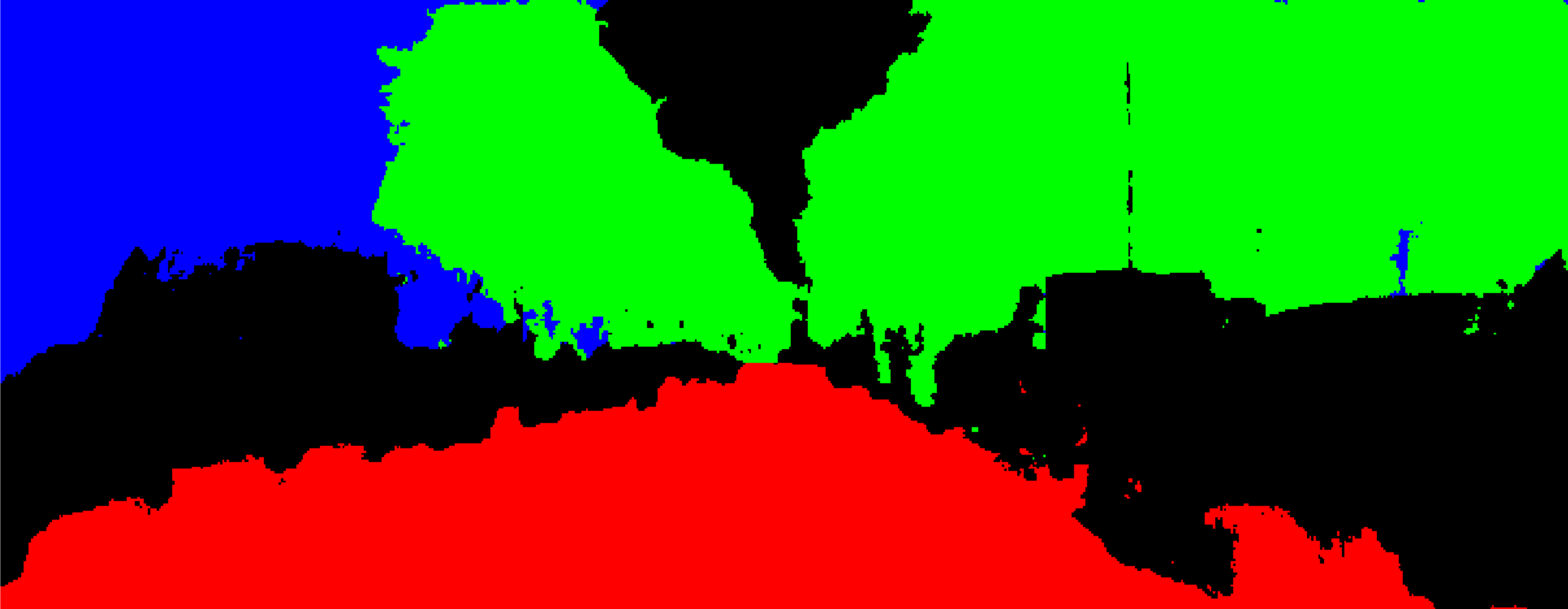}
  \includegraphics[width=\textwidth]{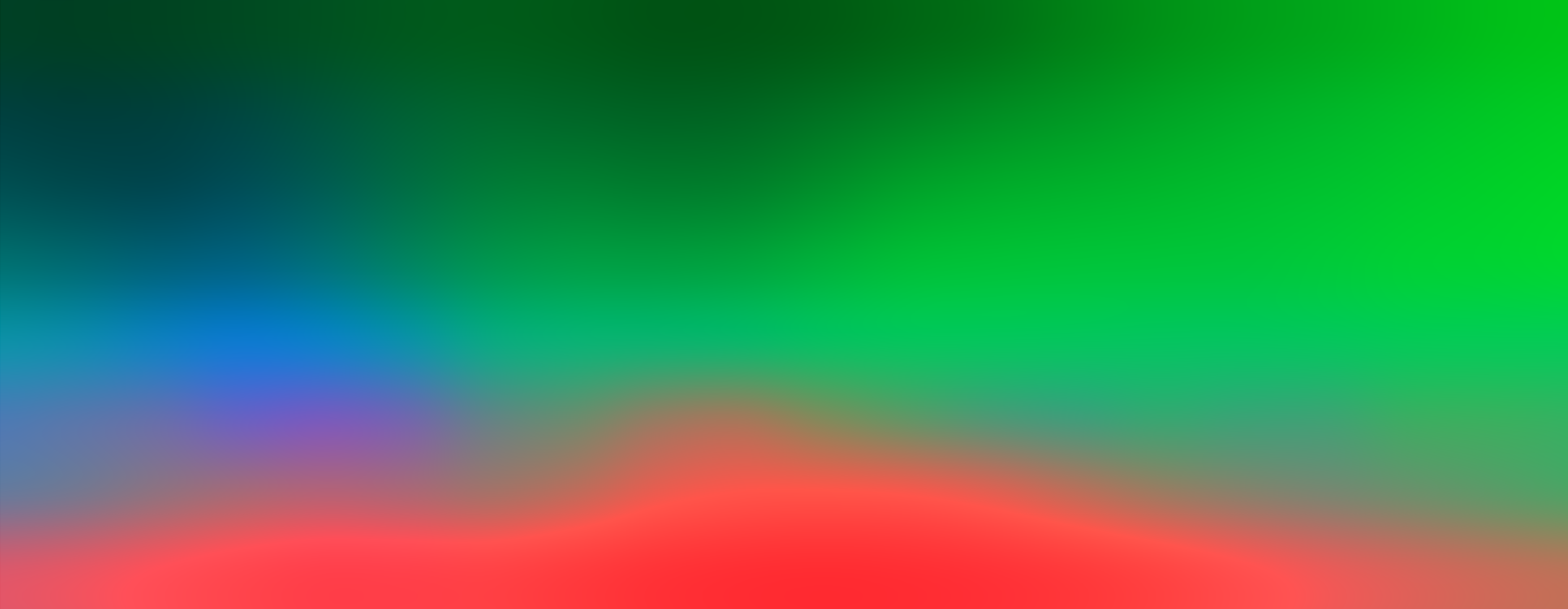}
  \end{minipage}
  \includegraphics[height=\aheight]{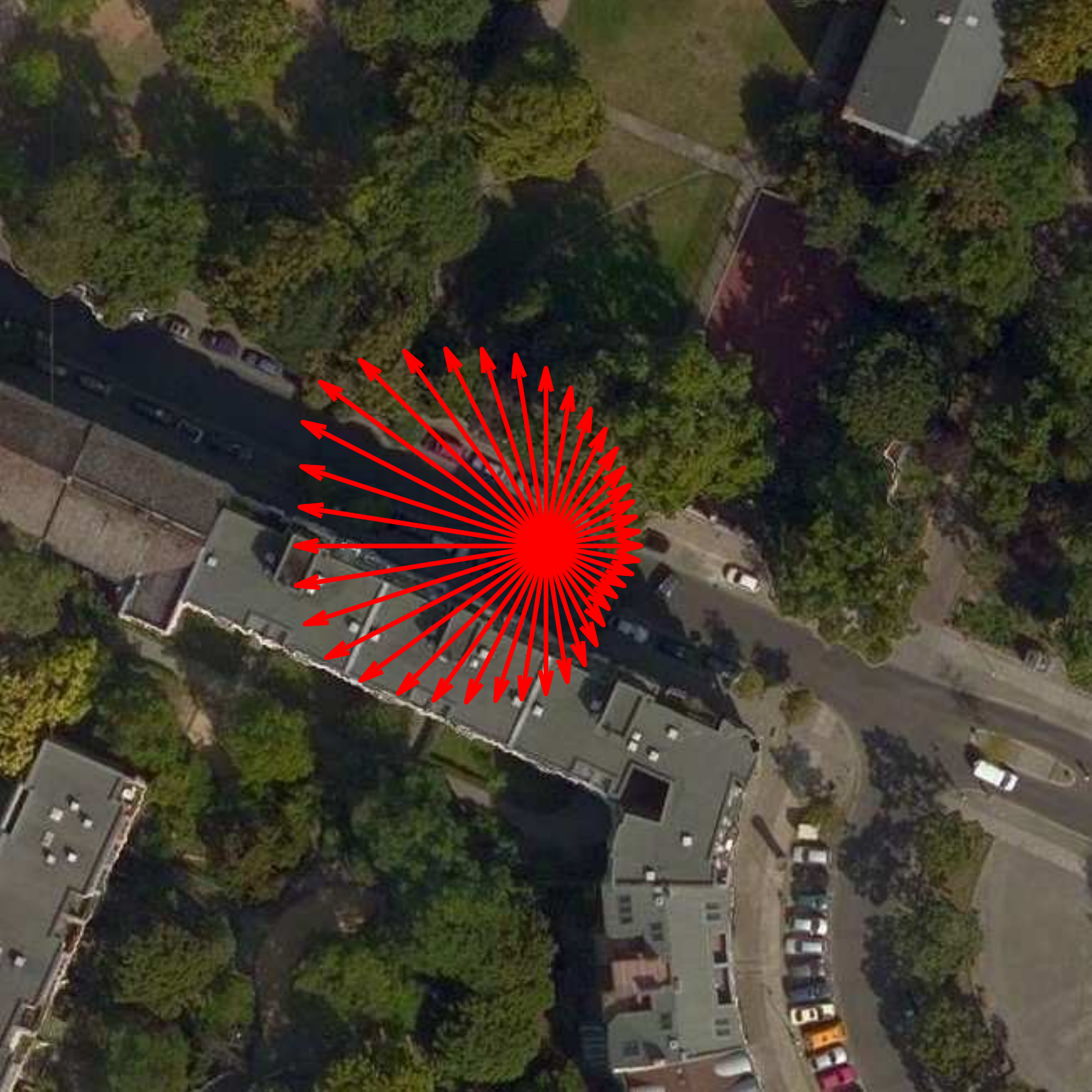}
  \hfill
  \begin{minipage}[b]{\gwidth}
  \includegraphics[width=\textwidth]{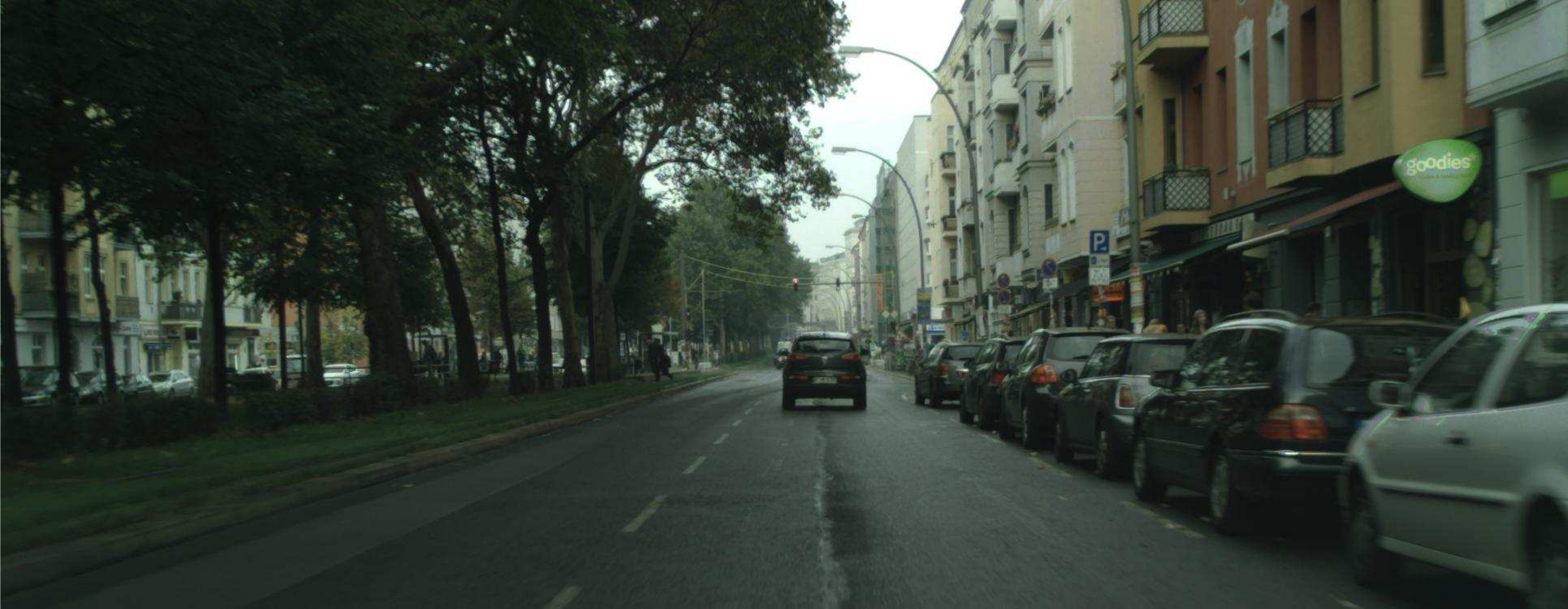}
  \includegraphics[width=\textwidth]{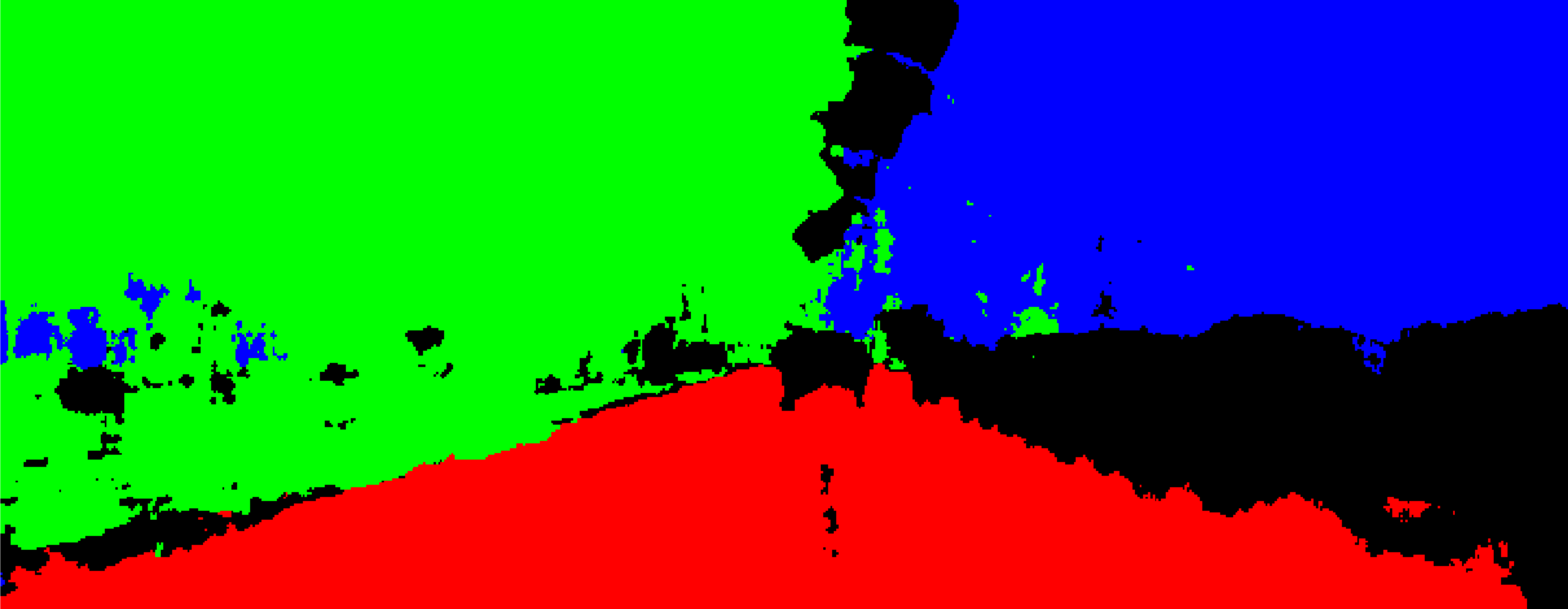}
  \includegraphics[width=\textwidth]{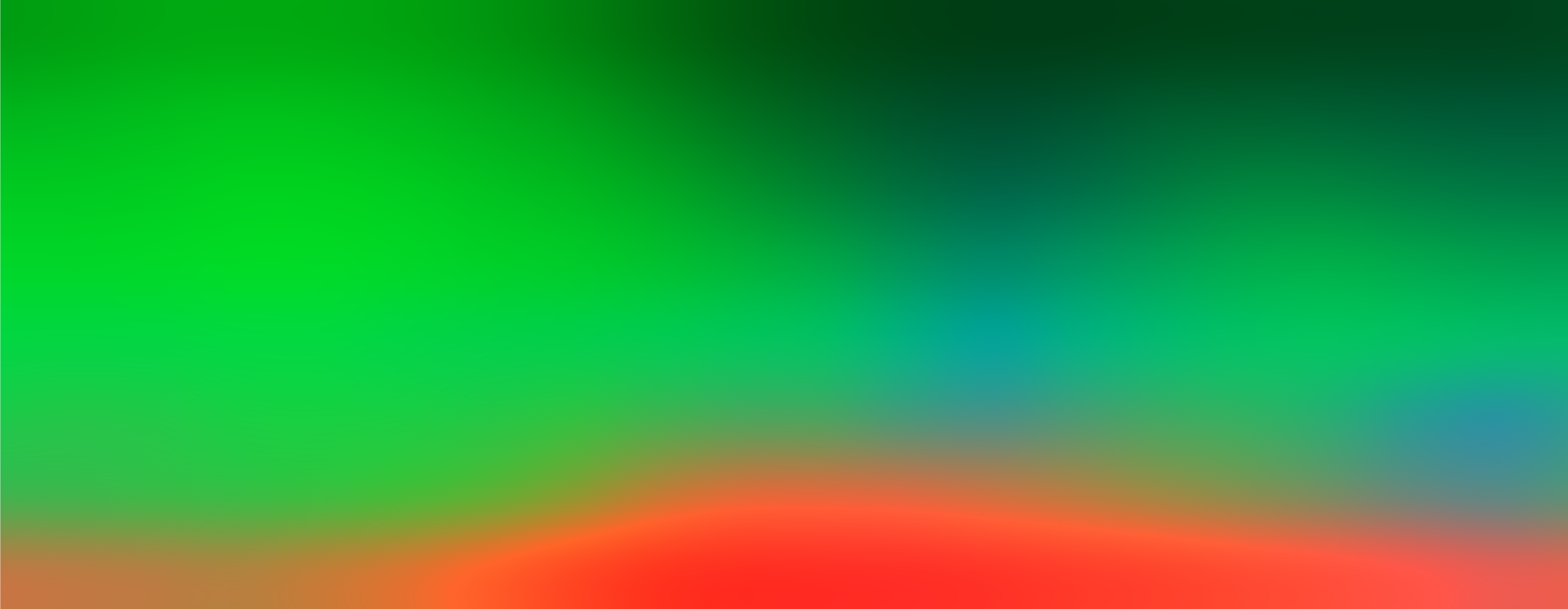}
  \end{minipage}
  \includegraphics[height=\aheight]{cityscapes/261_pred}

\vspace{5pt}
  \setlength{\aheight}{63pt}
  \setlength{\gwidth}{53.25pt}
  \begin{minipage}[b]{\gwidth}
  \includegraphics[width=\textwidth]{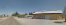}
  \includegraphics[width=\textwidth]{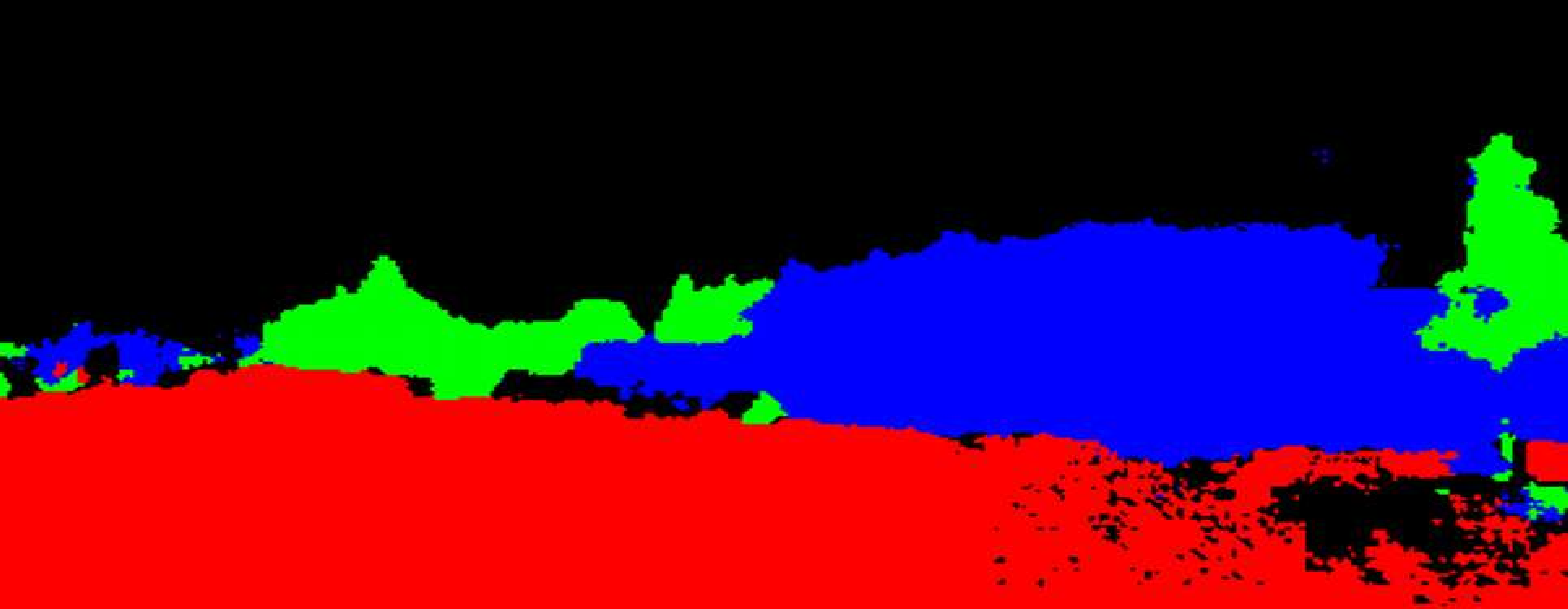}
  \includegraphics[width=\textwidth]{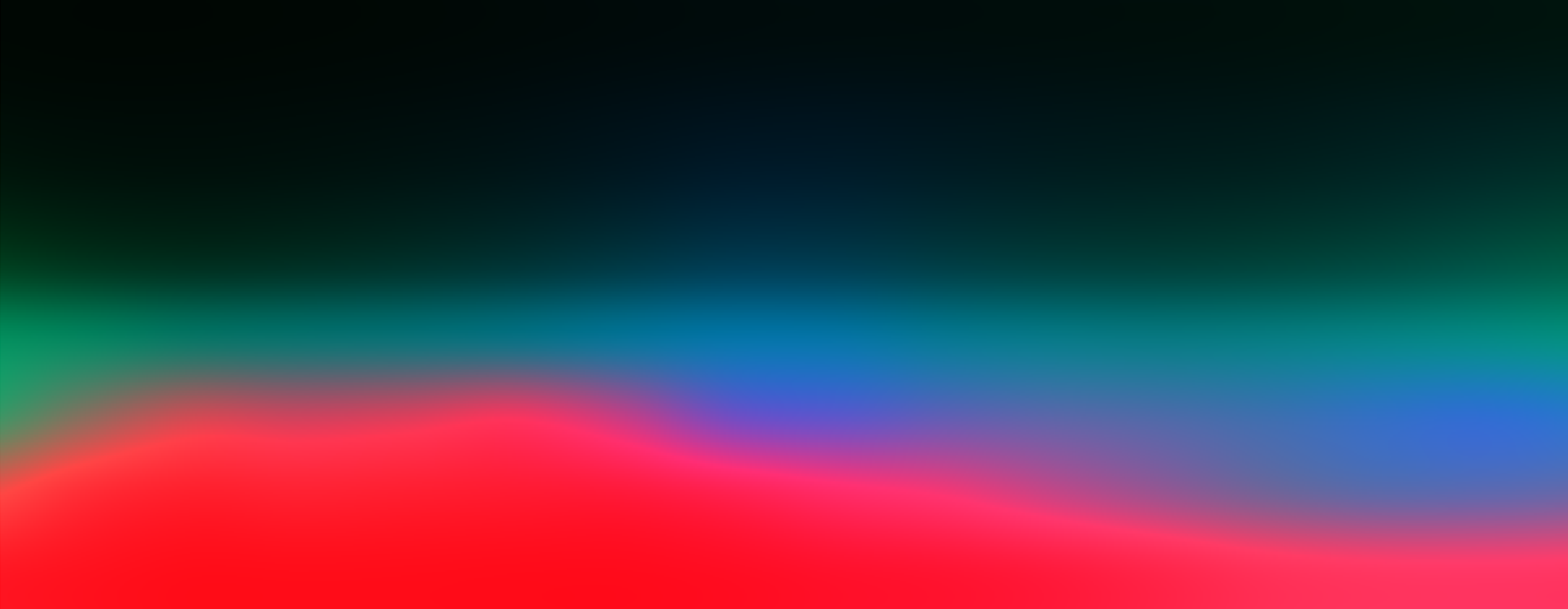}
  \end{minipage}
  \includegraphics[height=\aheight]{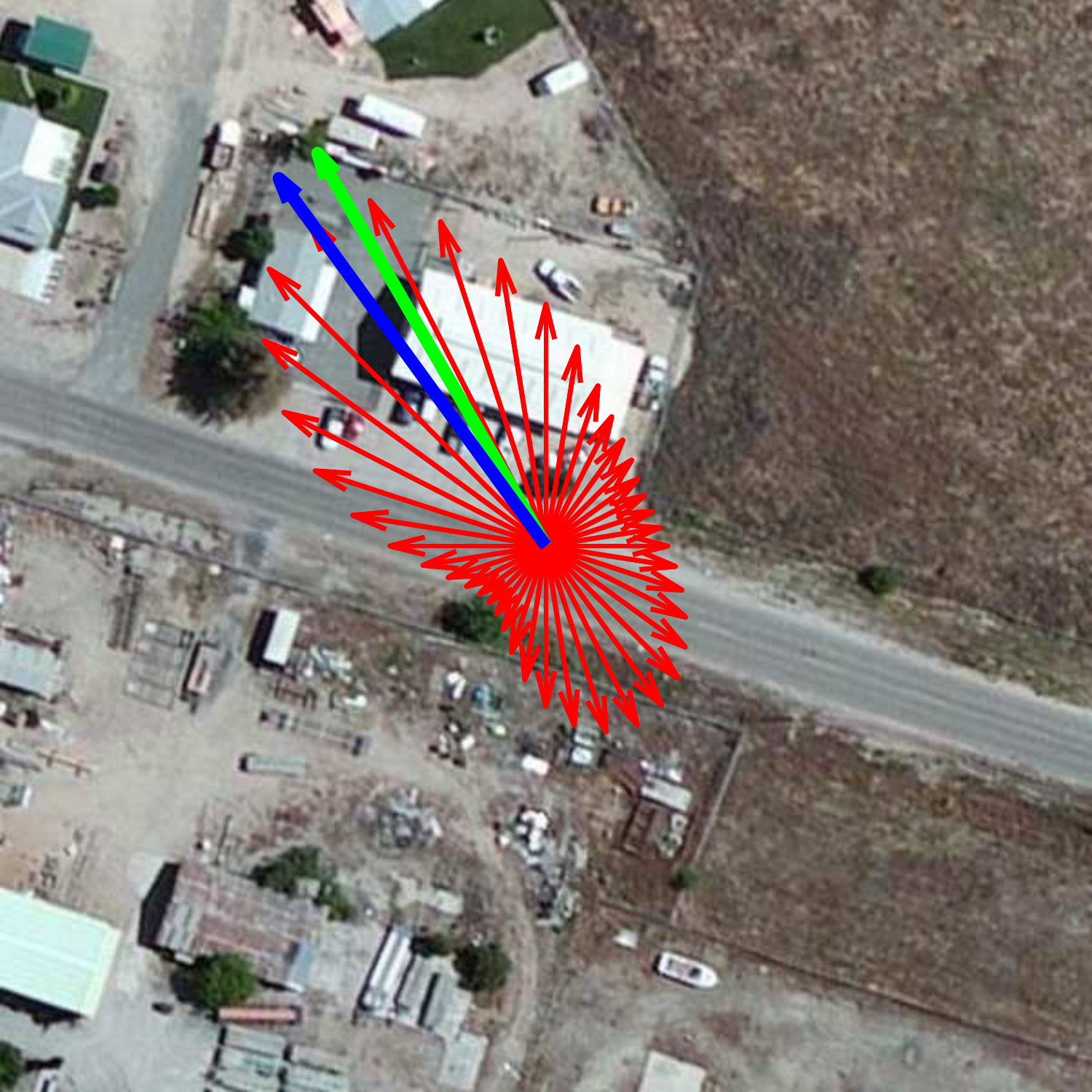}
  \hfill
  \begin{minipage}[b]{\gwidth}
  \includegraphics[width=\textwidth]{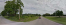}
  \includegraphics[width=\textwidth]{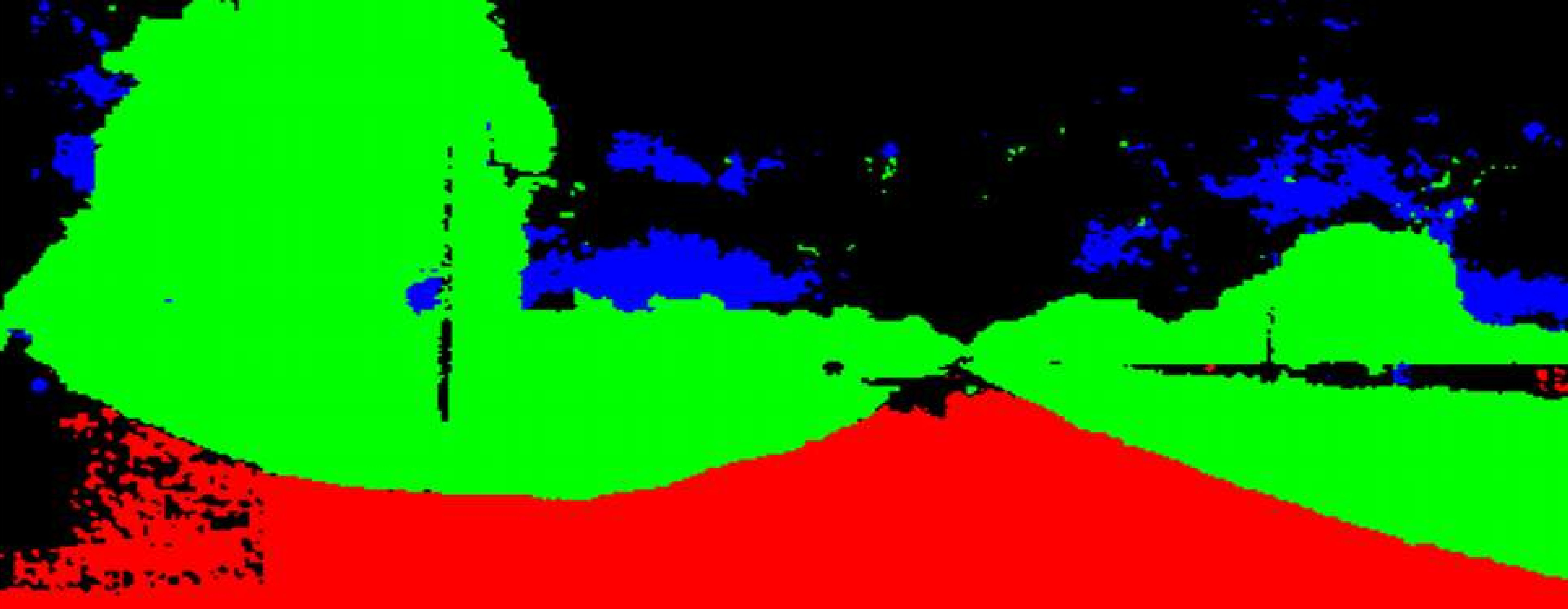}
  \includegraphics[width=\textwidth]{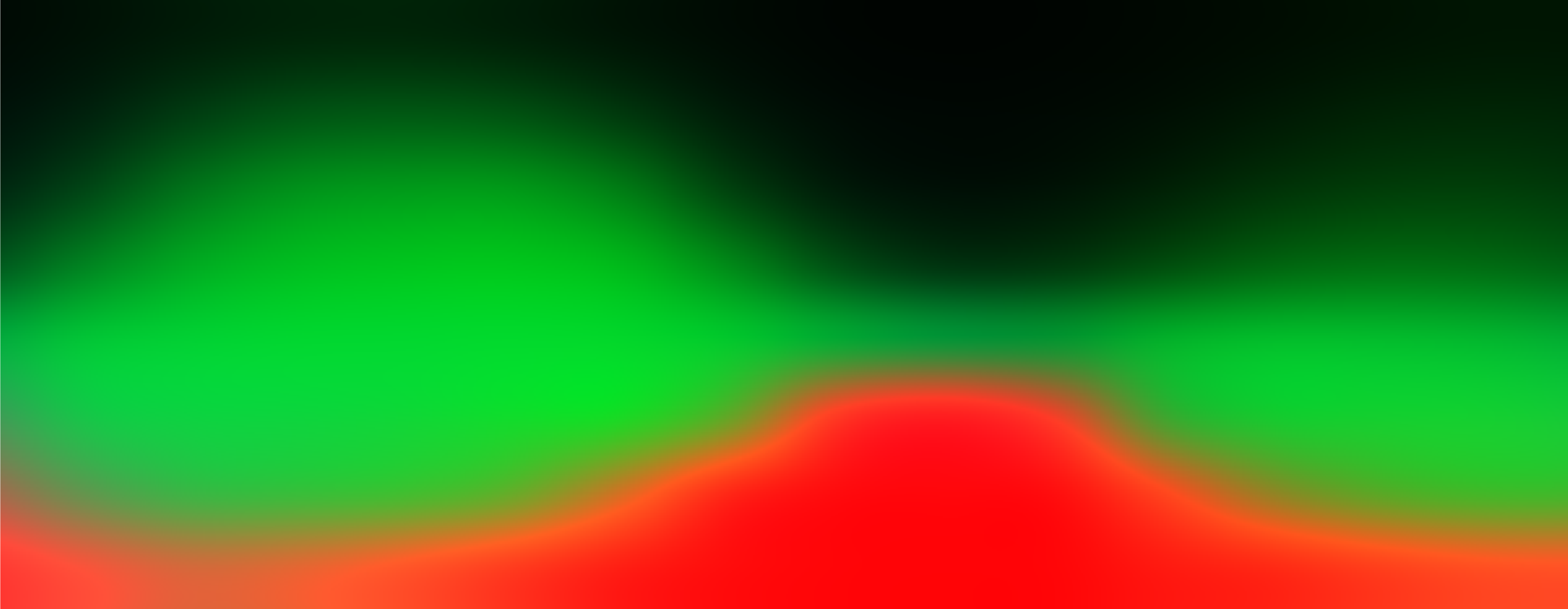}
  \end{minipage}
  \includegraphics[height=\aheight]{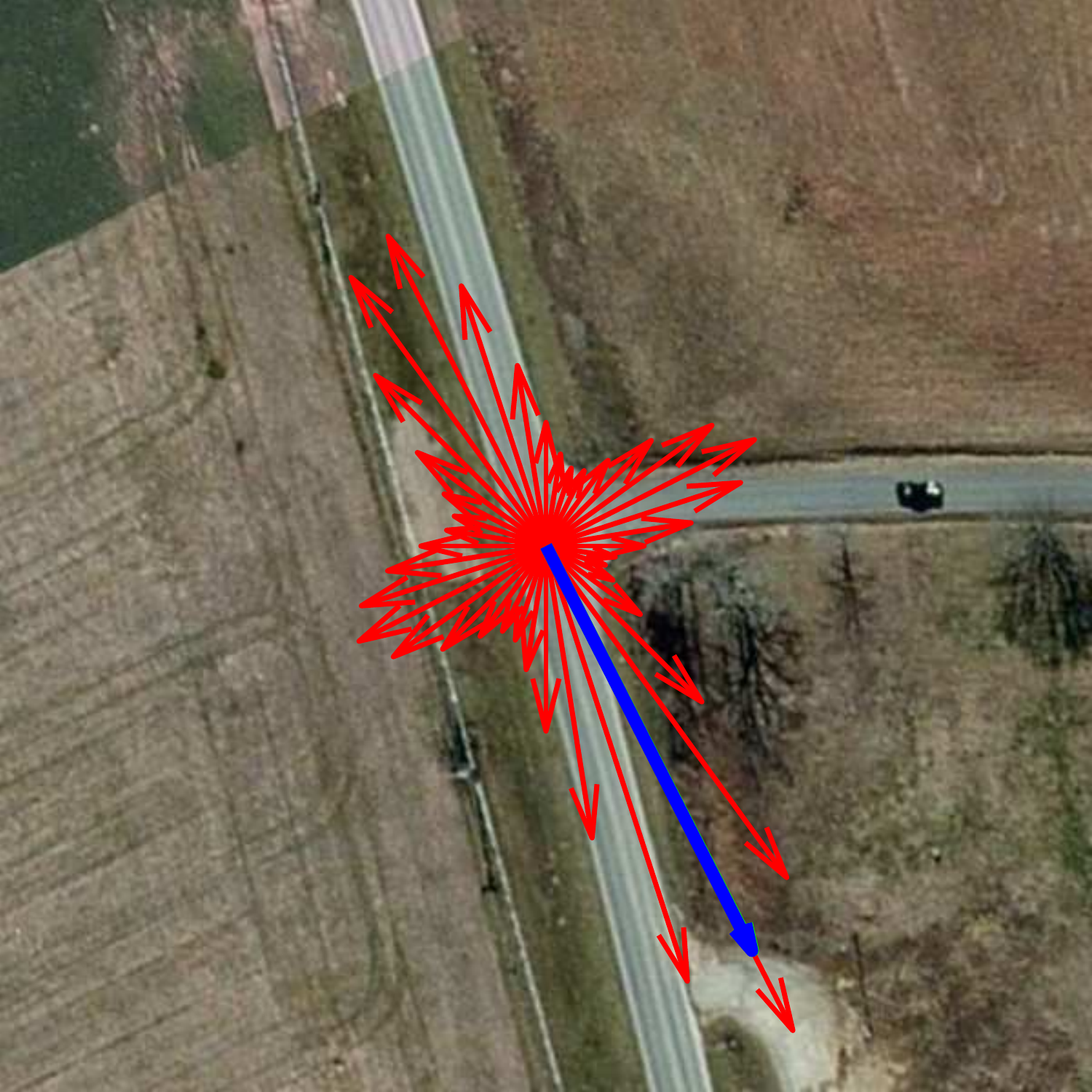}
  \hfill
  \begin{minipage}[b]{\gwidth}
  \includegraphics[width=\textwidth]{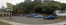}
  \includegraphics[width=\textwidth]{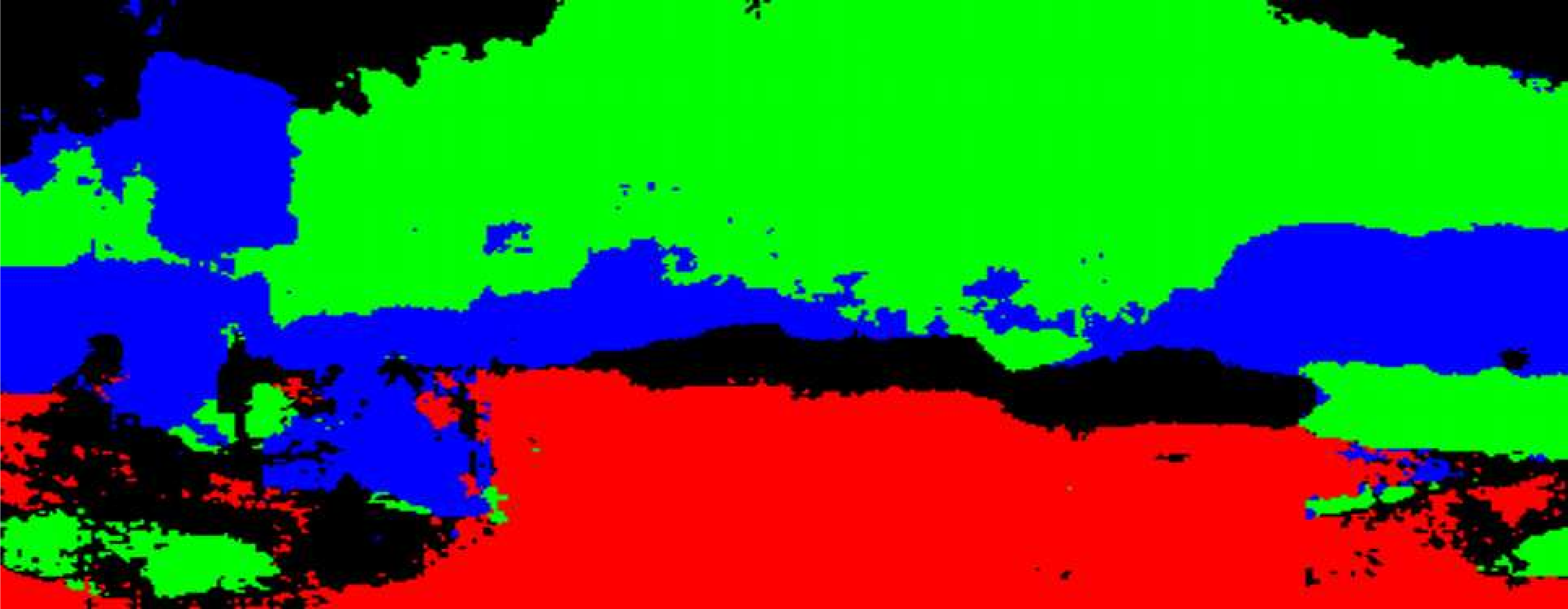}
  \includegraphics[width=\textwidth]{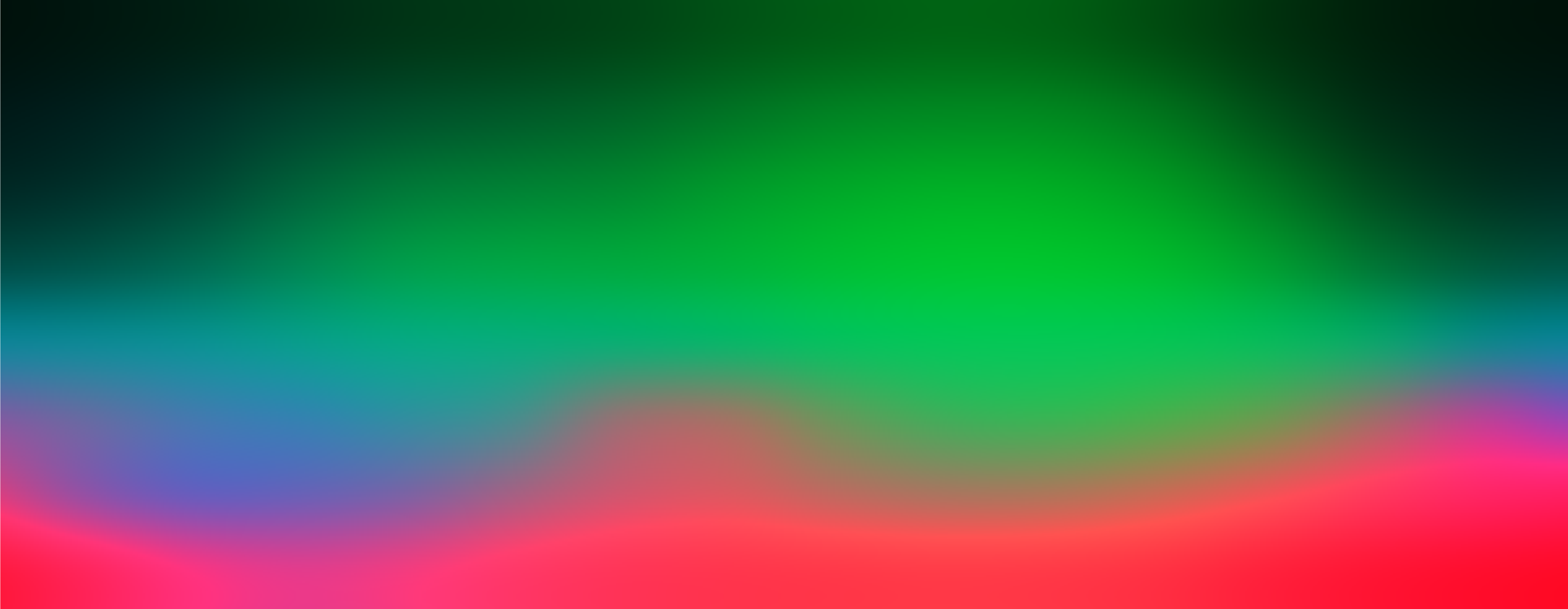}
  \end{minipage}
  \includegraphics[height=\aheight]{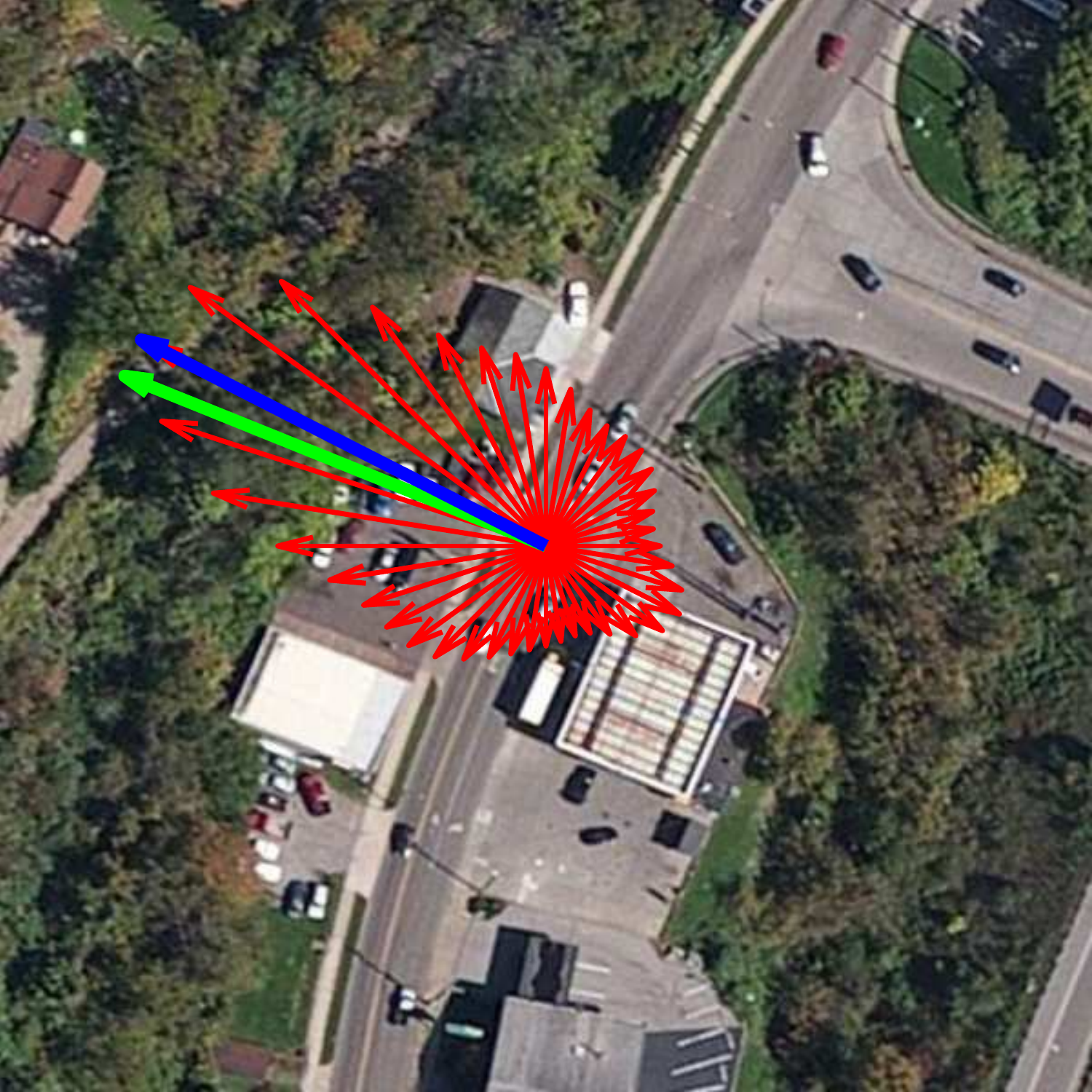}
  \hfill
  \begin{minipage}[b]{\gwidth}
  \includegraphics[width=\textwidth]{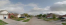}
  \includegraphics[width=\textwidth]{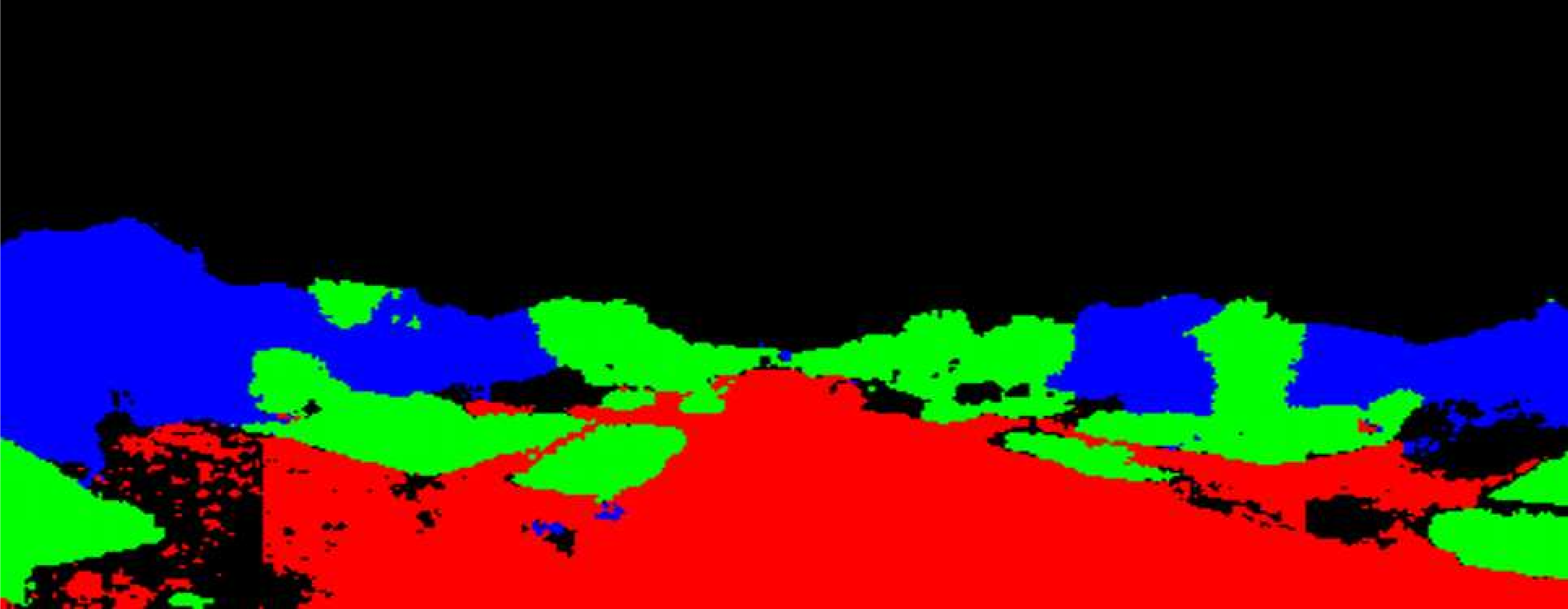}
  \includegraphics[width=\textwidth]{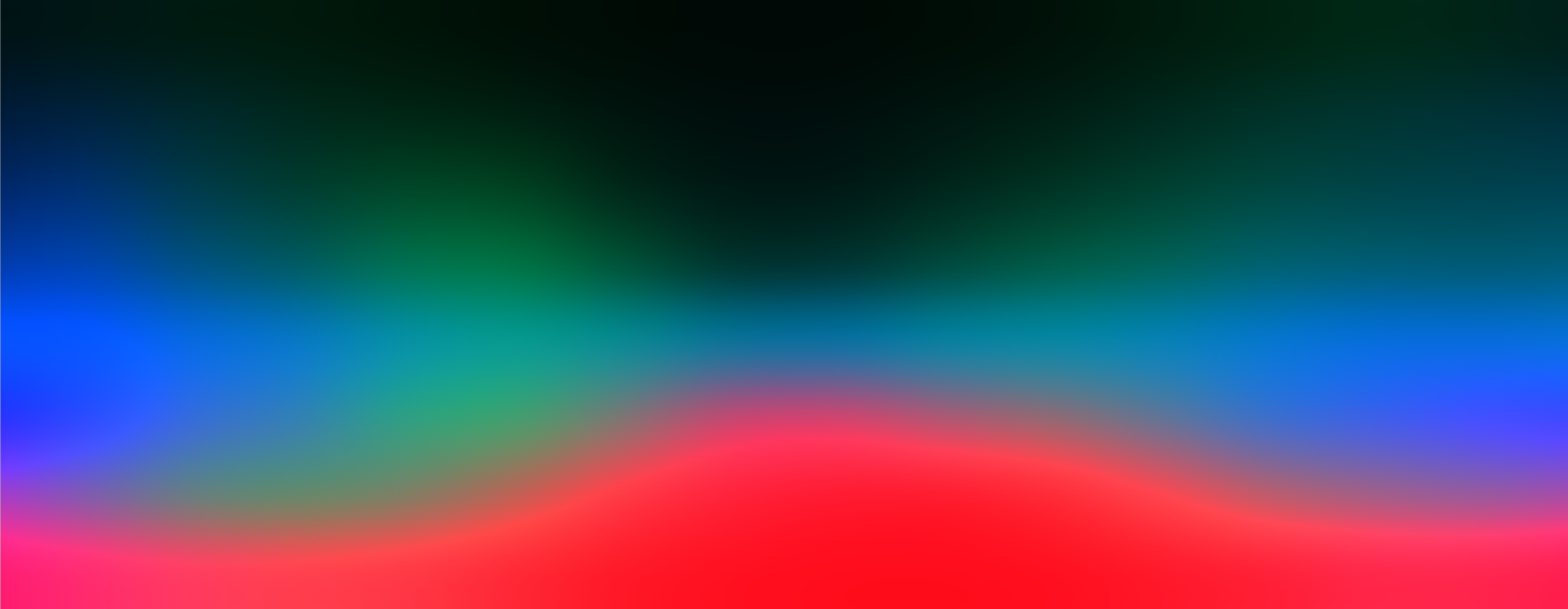}
  \end{minipage}
  \includegraphics[height=\aheight]{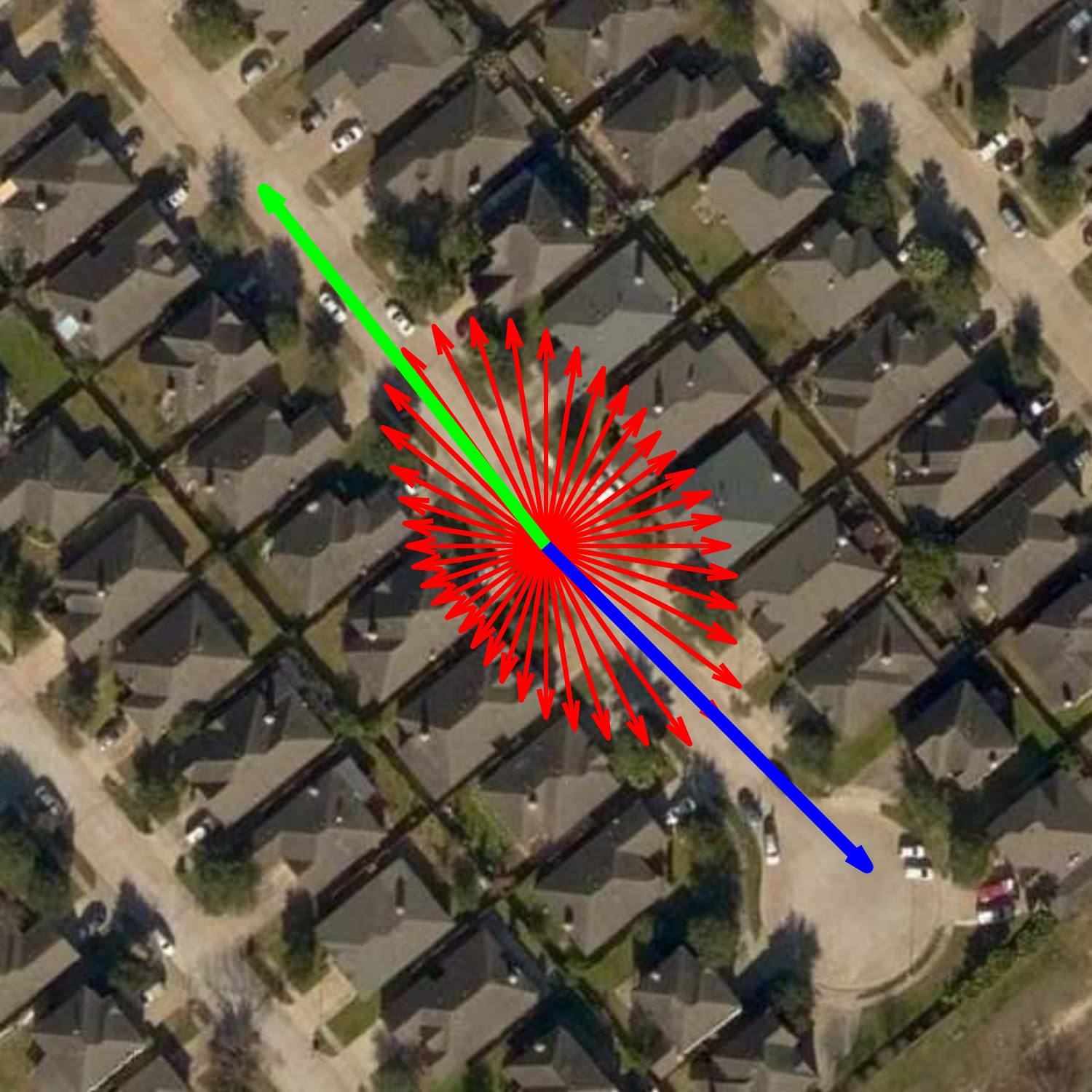}

  \caption{Qualitative results of orientation predictions on Cityscapes
dataset (top) and CVUSA (bottom). The $I_g$, $L_g$ and
$L_{g'}$ are stacked vertically on the left side of the aerial image.
We visualize three classes on the labels:
    {\em road} (red), {\em vegetation} (green), and {\em man-made} 
    (blue).
The discrete PDFs of the ground camera orientation are visualized with
red arrows, whose lengths indicate the magnitudes. In the CVUSA
results, the ground truth (green) and the optimal prediction (blue)
are also shown with the orientation PDF. The last prediction result is
a typical failure case of our method, where the scene is symmetric
from the aerial point of view.}
  \label{fig:orient:qualitative}
\end{figure*}

\begin{figure}
	\centering
	\includegraphics[width=.7\linewidth]{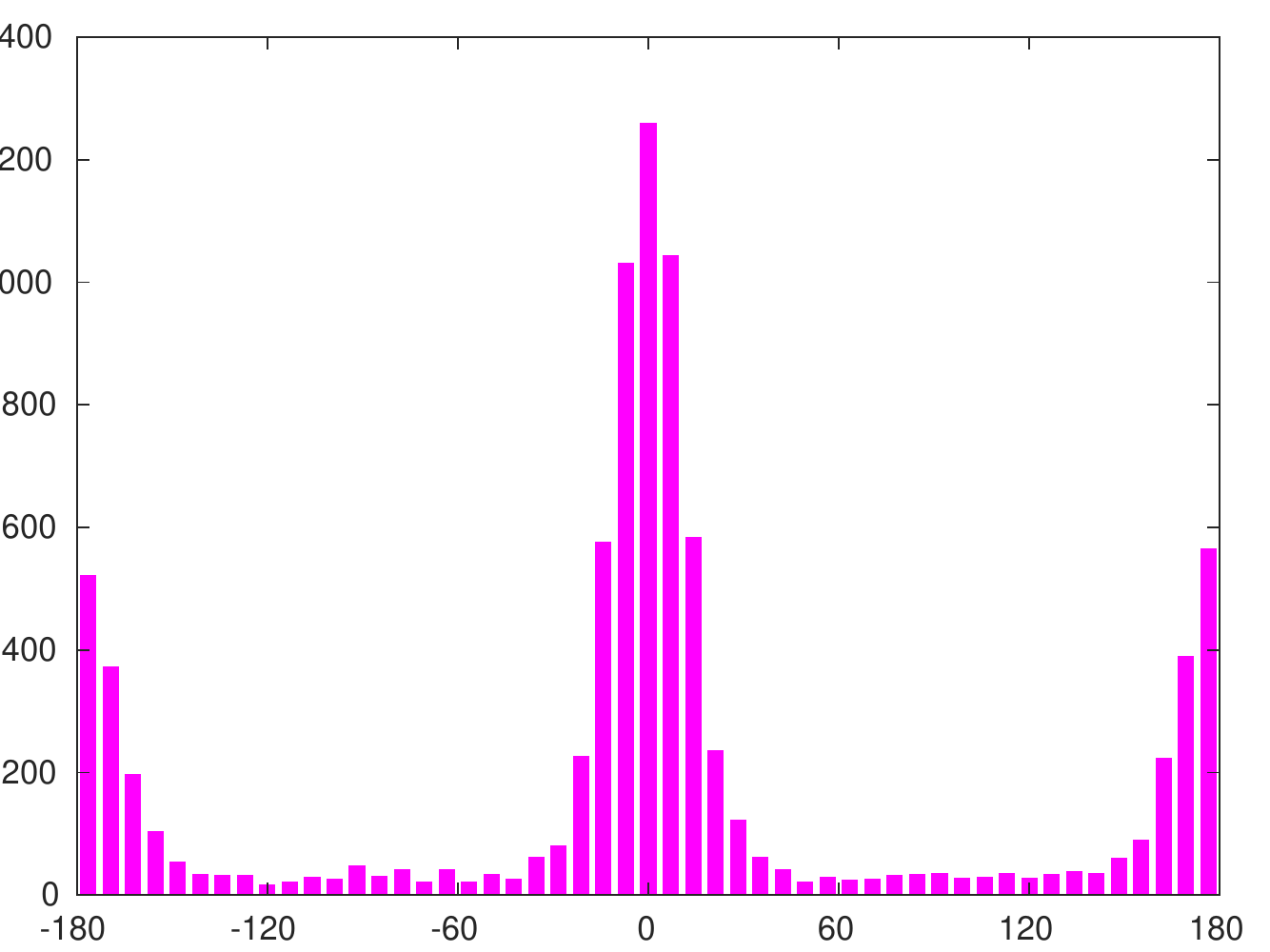}
  \caption{Histogram of the orientation errors on the CVUSA dataset.}
	\label{fig:orient:err}
\end{figure}

\begin{itemize}

\item{\textbf{CVUSA:}} We use the test set introduced
in~\secref{dataset} to create this dataset, it has two parts for
orientation estimation and geocalibration, respectively.  For the
orientation regression task, we rotate the aerial image to a random
angle. For the fine-grained geocalibration experiment, we center-crop
the aerial image around a random $x,y$ offset, then rotate the image to a
random angle.  In both experiments, the heading direction of the
ground images are the same. We center crop a 224 $\times$ 448
cutout from each ground image as the query image.

\item{\textbf{Cityscapes:}}  The Cityscapes dataset~\cite{Cordts2016Cityscapes} 
is a recently released benchmark dataset designed to support the task
of urban scene understanding through semantic pixel labeling. It
consists of stereo video from 50 different cities and fine pixel-level
annotations for 5,000 frames and coarse pixel-level annotations for
20,000 frames. 

\end{itemize}

\paragraph{Orientation Estimation}

For this task, we assume the location and focal length of the ground
image, $I_g$, is known but the orientation is not. The intuition
behind our method is that the semantic labeling of the ground image
will be most similar to the feature map of the aerial image at the
actual orientation.  For a query ground image, $I_g$, the first step
is to download the corresponding aerial image, $I_a$.  We then infer
the semantic labeling of the query image, $I_g \rightarrow L_g$, and
predict the ground image label from the aerial image using our learned
network, $I_a \rightarrow L_{g'}$.  We assign an energy to each
possible orientation by computing the cross entropy between $L_g$ and
$L_{g'}$ in a sliding window fashion across all possible orientations.
We select the orientation with the lowest energy.  We present sample results in 
\figref{orient:qualitative} and a histogram of the orientation errors on the 
CVUSA dataset in \figref{orient:err}.


\paragraph{Fine-grained Geocalibration}

\begin{figure*}
  \setlength{\aheight}{93pt}
  \setlength{\gwidth}{60.65pt}
  \centering
  \begin{minipage}[b]{\gwidth}
  \includegraphics[width=\textwidth]{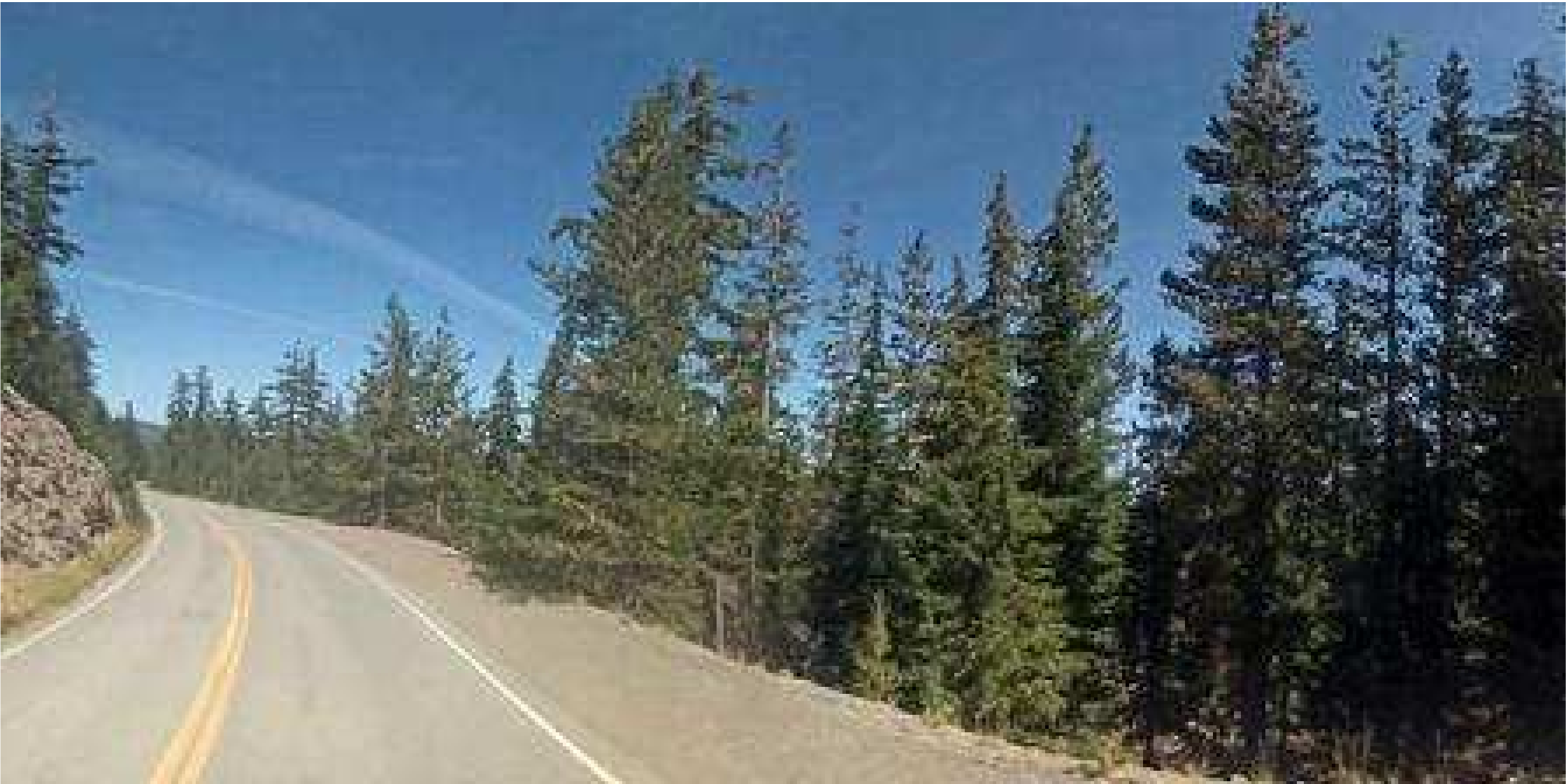}
  \includegraphics[width=\textwidth]{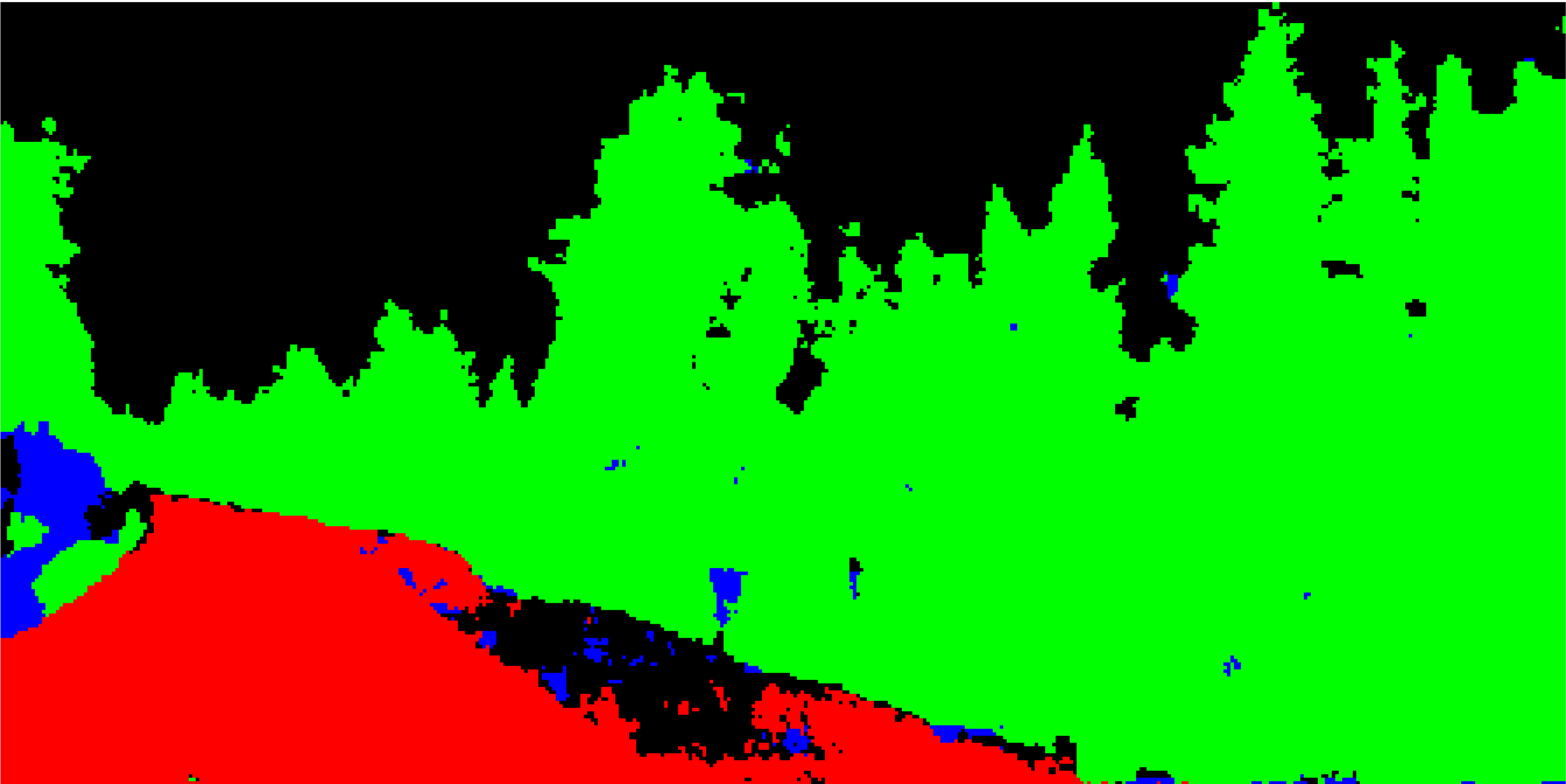}
  \includegraphics[width=\textwidth]{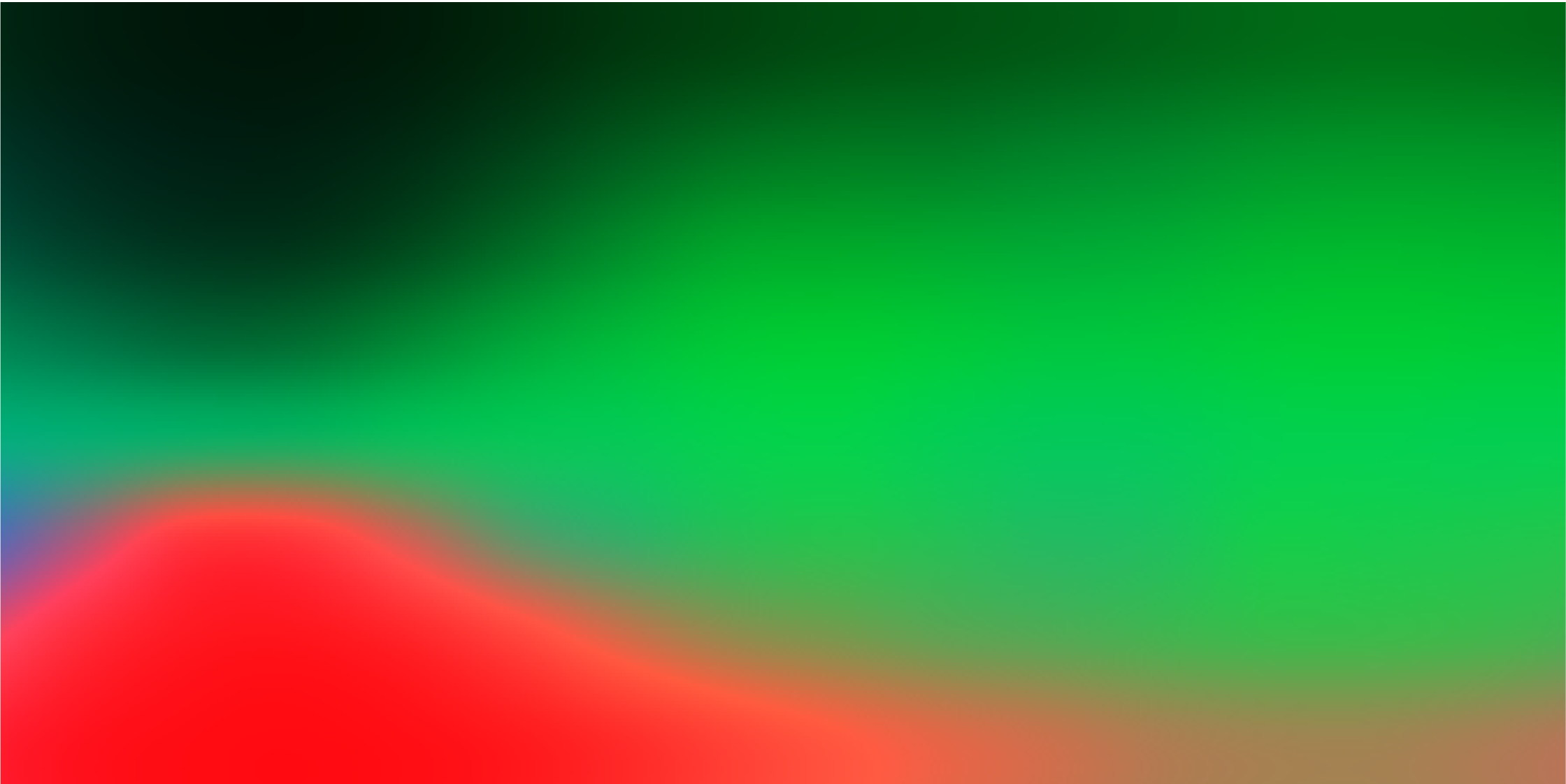}
  \end{minipage}
  \includegraphics[height=\aheight]{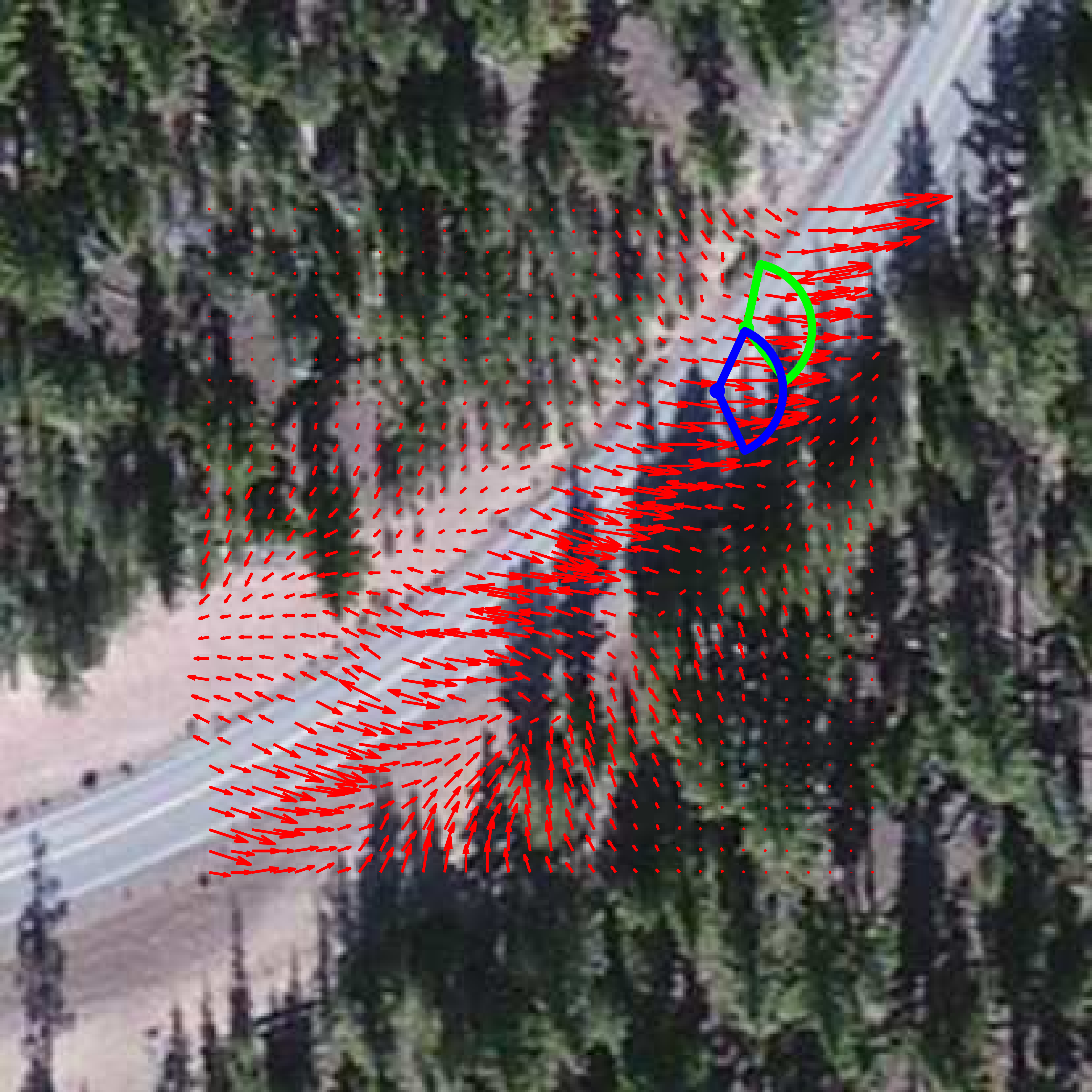}
  \hfill
  \begin{minipage}[b]{\gwidth}
  \includegraphics[width=\textwidth]{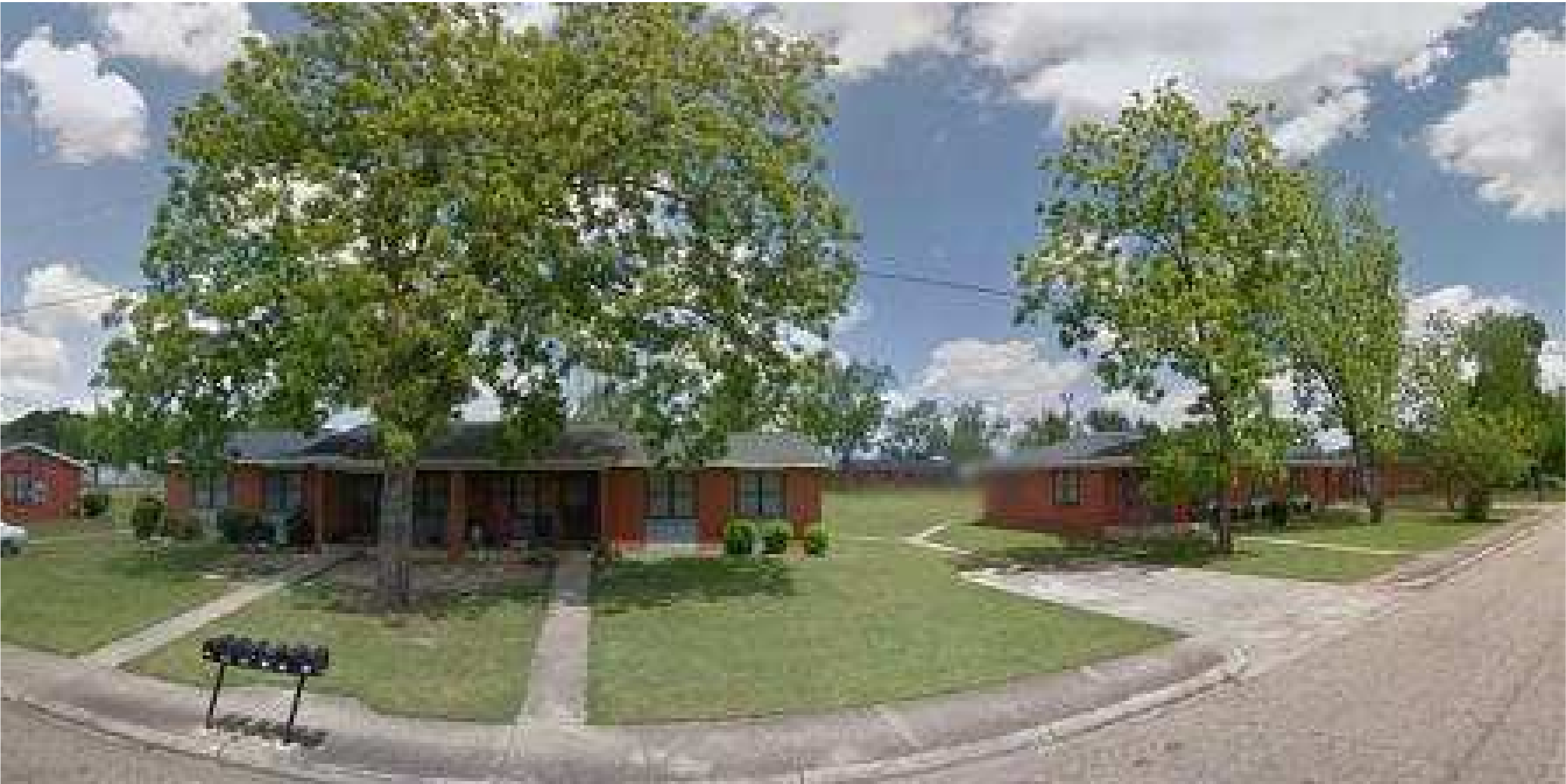}
  \includegraphics[width=\textwidth]{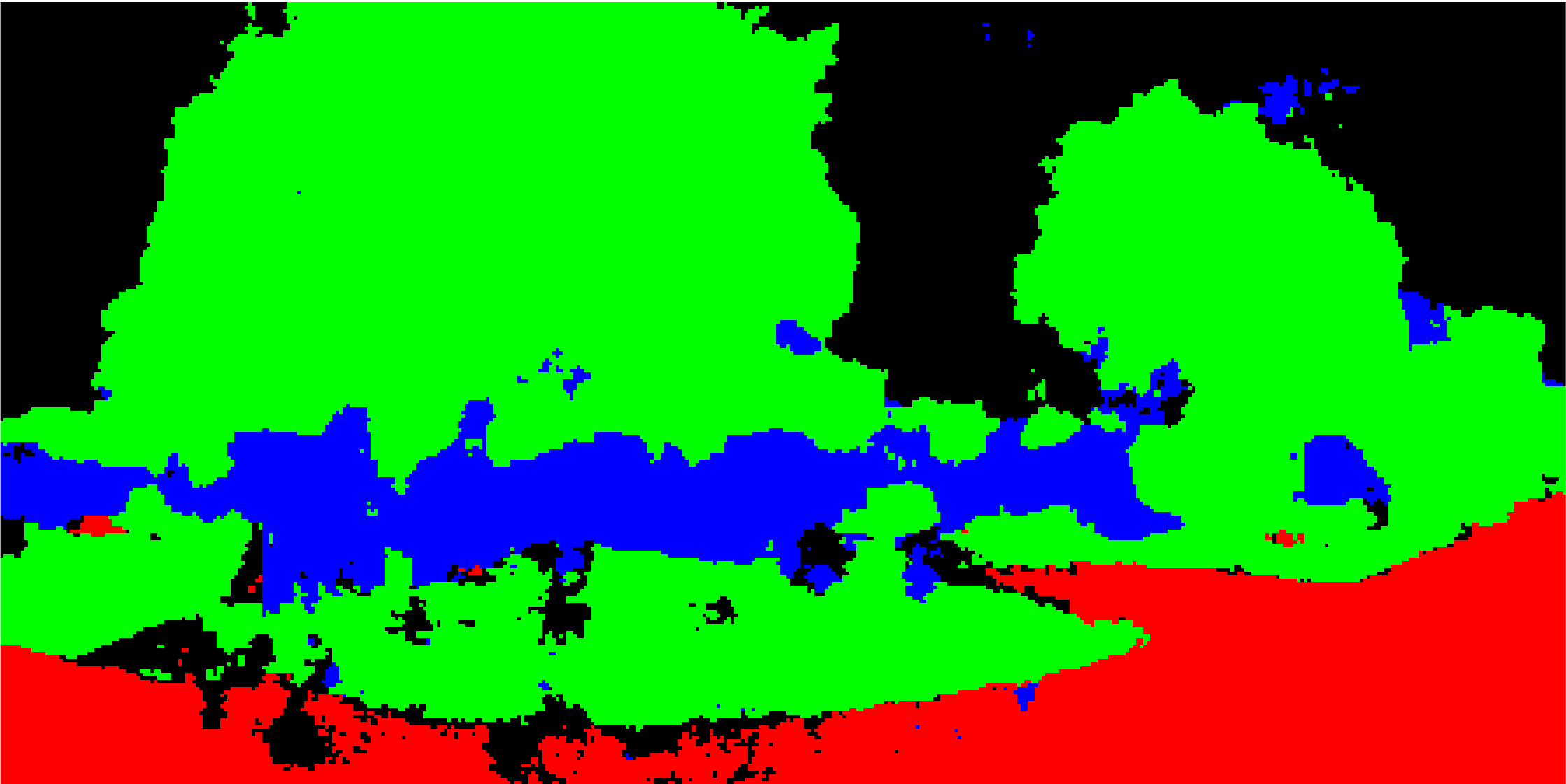}
  \includegraphics[width=\textwidth]{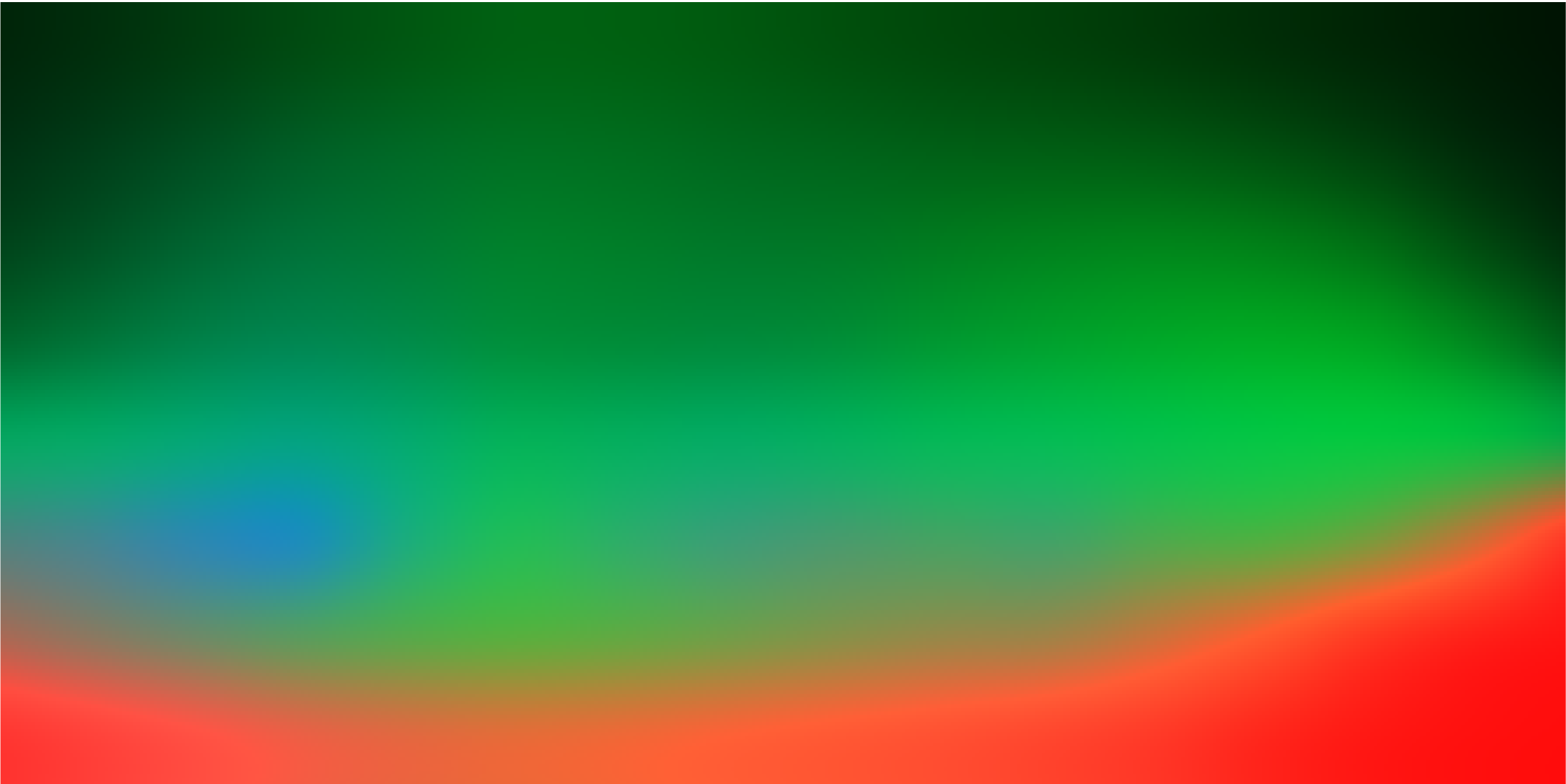}
  \end{minipage}
  \includegraphics[height=\aheight]{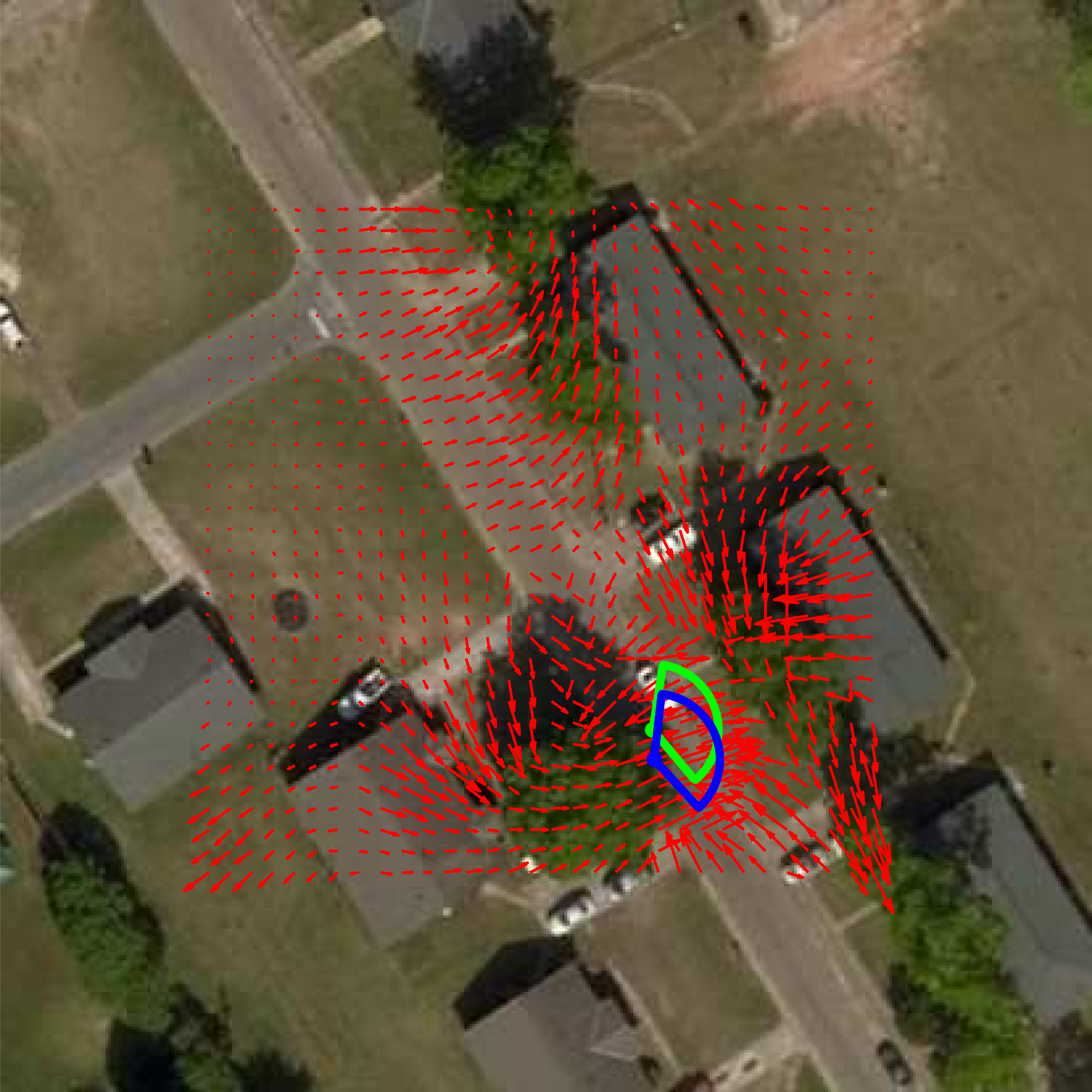}
  \hfill
  \begin{minipage}[b]{\gwidth}
  \includegraphics[width=\textwidth]{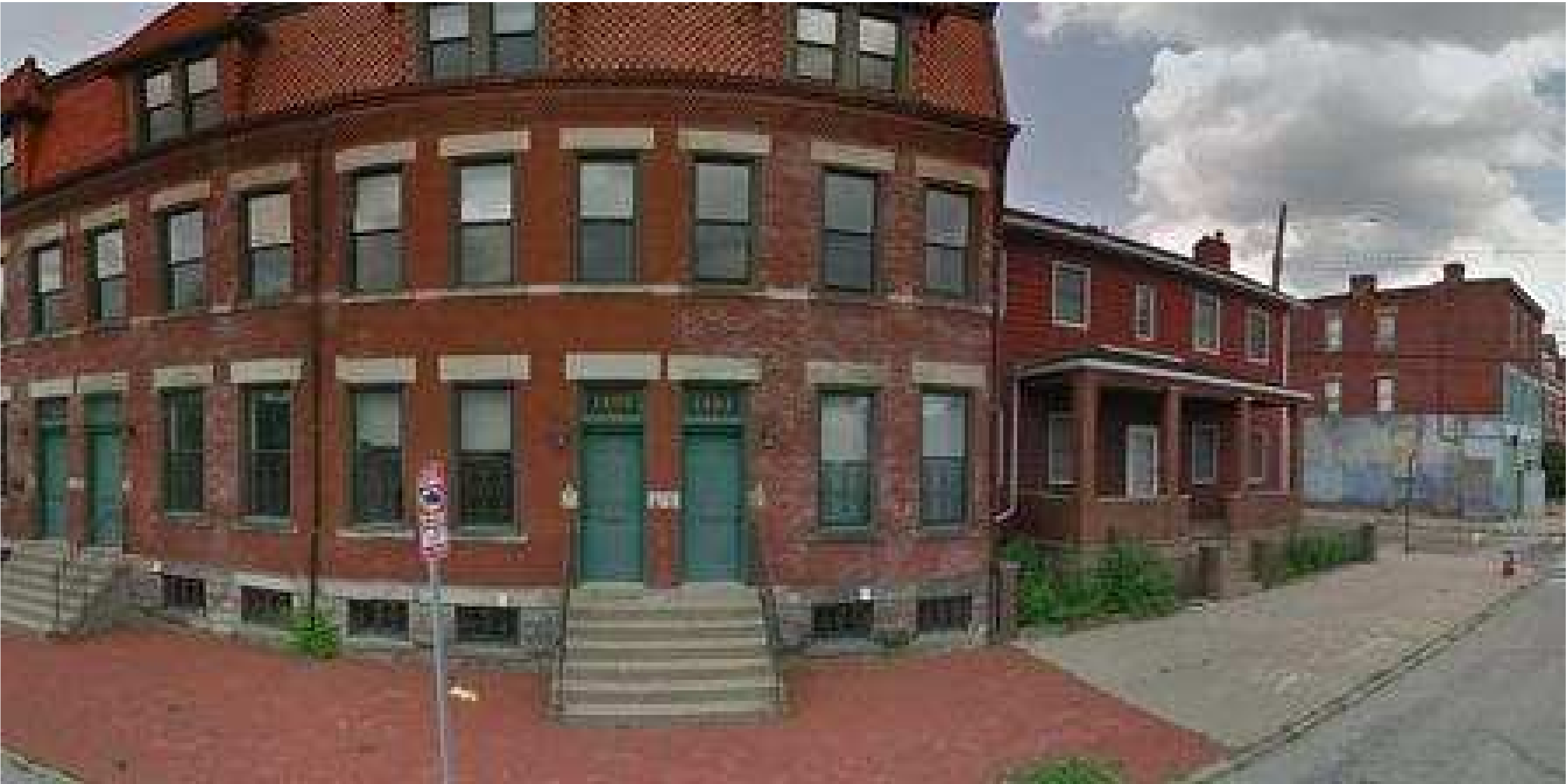}
  \includegraphics[width=\textwidth]{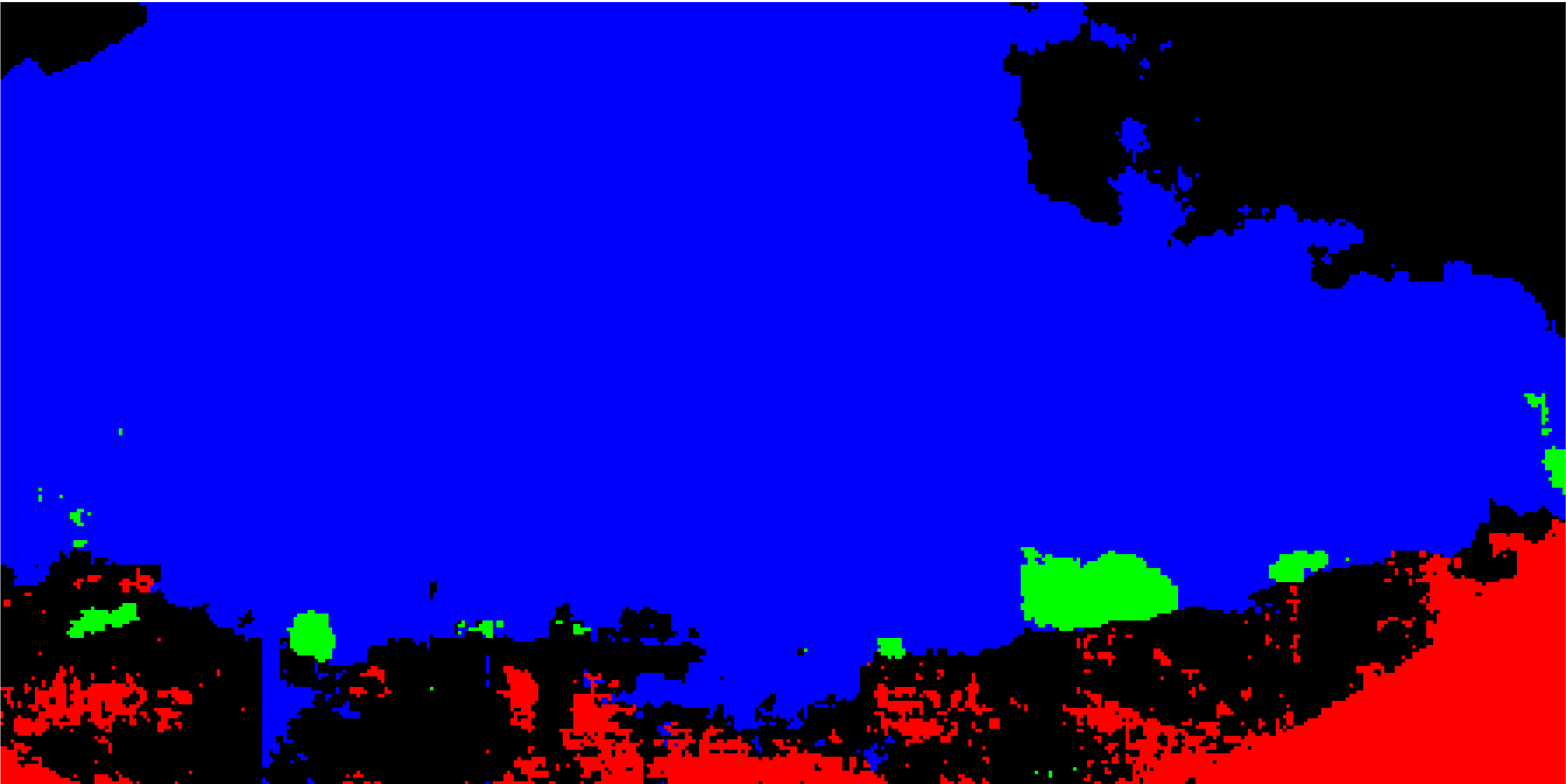}
  \includegraphics[width=\textwidth]{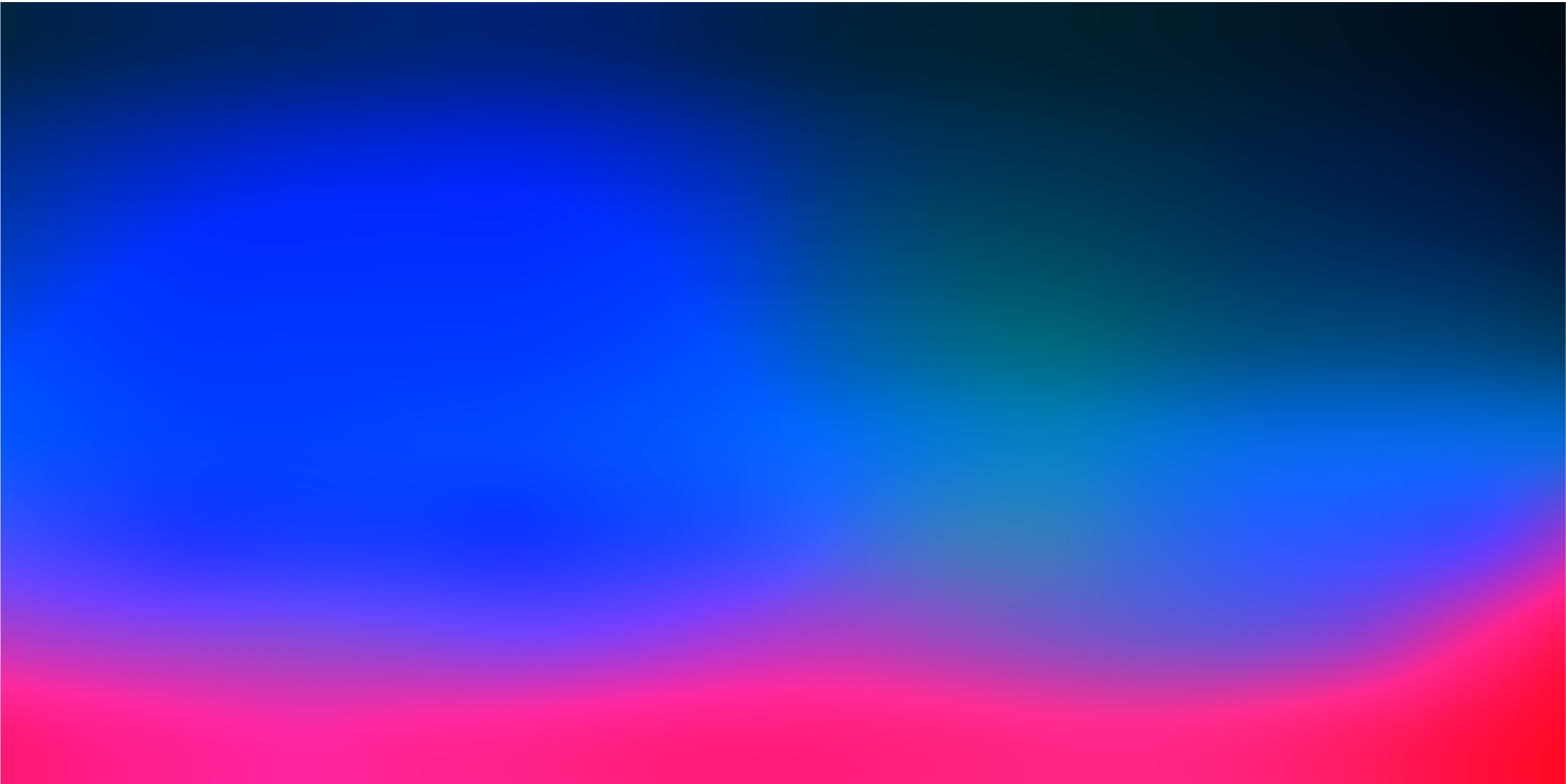}
  \end{minipage}
  \includegraphics[height=\aheight]{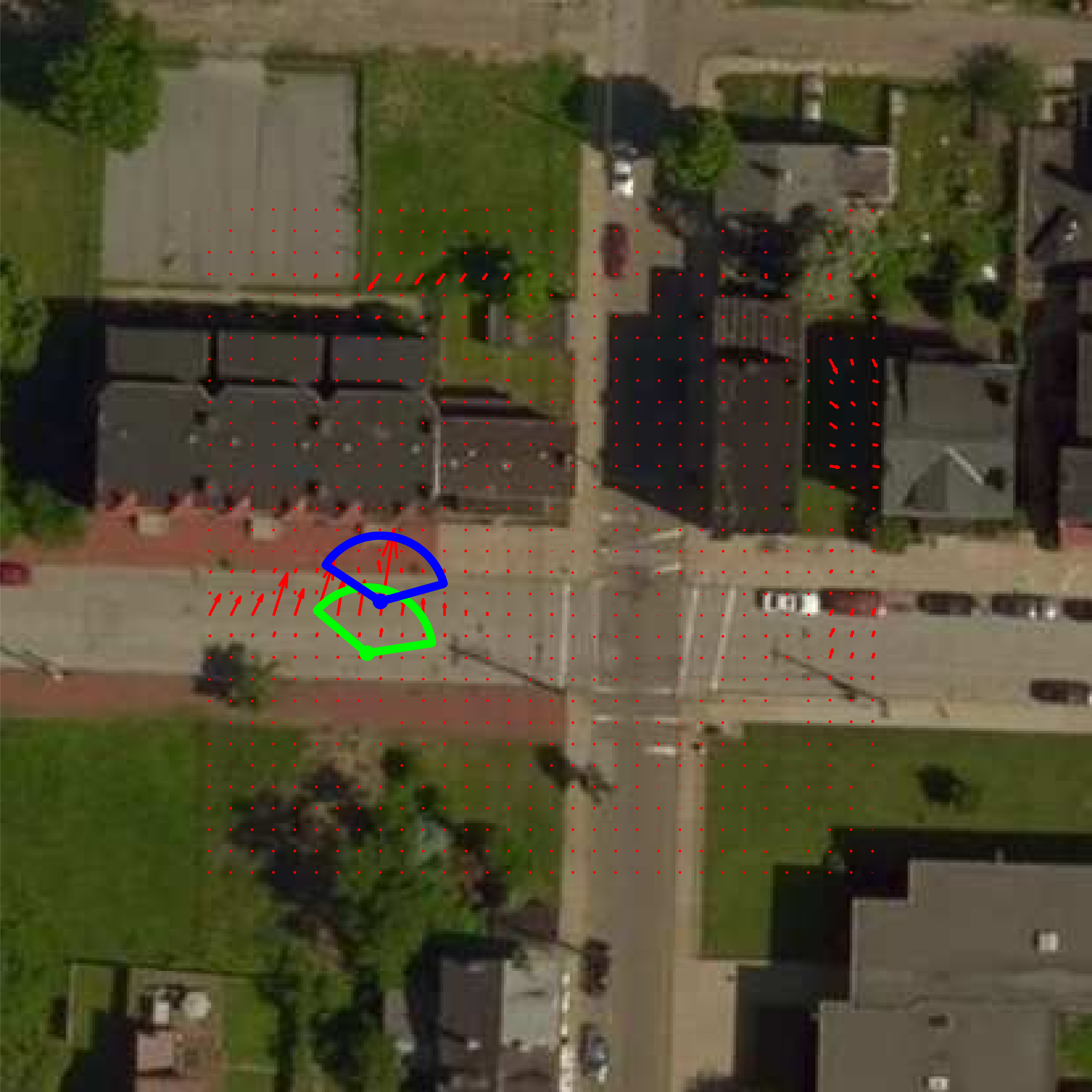}
  \hfill

  \caption{Fine-grained geocalibration results on CVUSA. (left) From
top to bottom are the $I_g$, $L_g$, and $L_{g'}$ respectively.  We
visualize three classes on the labels: {\em road} (red), {\em
vegetation} (green), and {\em man-made} (blue).  (right) Orientation
flow map (red), where the arrow direction indicates the optimal
direction at that location and length indicates the magnitude. We also
show the optimal prediction and the ground-truth frustums in blue and
green respectively.}
  \label{fig:geocali:cali}
\end{figure*}

For this task, we assume that we know the focal length of the camera
and have a rough estimate of the camera location (i.e., with 100
meters).  We extract 256 $\times$ 256 aerial images from the area
around our rough estimate and extract the corresponding ground-level
feature maps.  We apply our orientation estimation procedure to each
feature map.  The result is a distribution over orientations for each
location.  \figref{geocali:cali} shows several example results,
including the most likely direction for each location, as well as the
most likely location and orientation pair.

\subsection{Synthesizing Ground Images from Aerial Images}

\begin{figure}
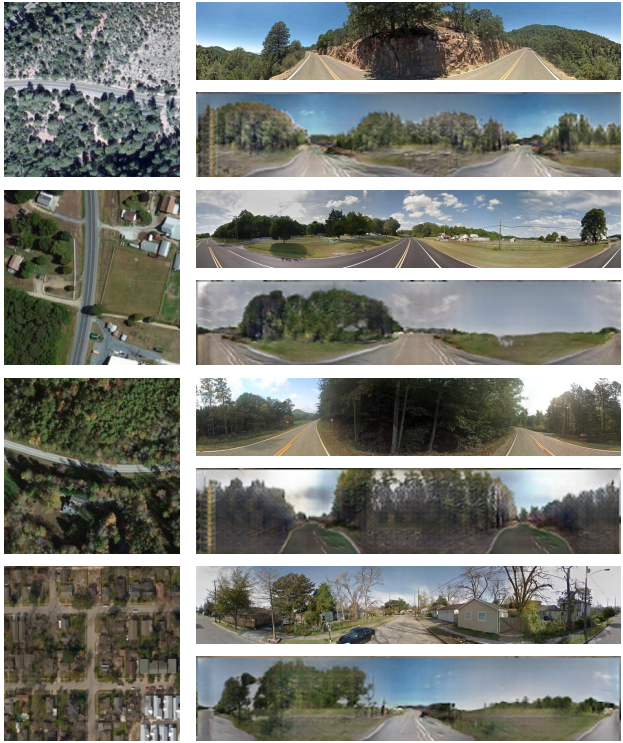

	\newlength{\awidth}
	\setlength{\awidth}{.279\linewidth}
	\setlength{\gwidth}{.67725\linewidth}
	\setlength{\gspace}{4pt}
	\centering
	\includegraphics[width=\awidth]{gan/aerial1}\hspace*{\gspace}
	\begin{minipage}[b]{\gwidth}
	\includegraphics[width=\textwidth]{gan/ground1}\vfill
	\vspace*{\gspace}
	\includegraphics[width=\textwidth]{gan/ground1_pred}
	\end{minipage}\vspace*{\gspace}
	
	\includegraphics[width=\awidth]{gan/aerial2}\hspace*{\gspace}
	\begin{minipage}[b]{\gwidth}
	\includegraphics[width=\textwidth]{gan/ground2}\vfill
	\vspace*{\gspace}
	\includegraphics[width=\textwidth]{gan/ground2_pred}
	\end{minipage}\vspace*{\gspace}
	
	\includegraphics[width=\awidth]{gan/aerial3}\hspace*{\gspace}
	\begin{minipage}[b]{\gwidth}
	\includegraphics[width=\textwidth]{gan/ground3}\vfill
	\vspace*{\gspace}
	\includegraphics[width=\textwidth]{gan/ground3_pred}
	\end{minipage}\vspace*{\gspace}
	
	\includegraphics[width=\awidth]{gan/aerial4}\hspace*{\gspace}
	\begin{minipage}[b]{\gwidth}
	\includegraphics[width=\textwidth]{gan/ground4}\vfill
	\vspace*{\gspace}
	\includegraphics[width=\textwidth]{gan/ground4_pred}
	\end{minipage}

	\caption{Synthesized ground-level views. Each row shows an aerial image 
	(left), its corresponding ground-level panorama (top-right), and predicted 
	ground-level panorama (bottom-right).}
	\label{fig:gan}
\end{figure}

We propose a novel application to infer a ground image by using features 
extracted from our network. We begin by describing our network structure and 
then show qualitative results for different generated ground-level scenes.

Our network architecture is based on the deep, directed generative model 
proposed by Kim \etal~\cite{kim2016deep}. Their model consists of two parts: a 
deep generator, $G$, which generates images that try to minimize a deep energy 
model, $E$. A low energy implies the image is real and high energy implies the 
image is fake. The architecture and training methods are inspired by generative 
adversarial networks~\cite{goodfellow2014generative}, however it provides a 
energy-based formulation of the discriminator to address common 
instabilities of adversarial training. 

A complete description of the architecture used to design the deep generator 
and deep energy model is provided in our supplemental materials. We begin by 
extracting an 8 
$\times$ 40 $\times$ 512 cross-view feature map, $f$, that has been learned to 
relate an aerial and ground image pair.  The generator is given $f$ 
along with random noise, $z$, as input. The generator outputs a 64 $\times$ 320 
panorama, $I_{\hat{g}}$, that represents the predicted ground image. The 
cross-view feature, predicted panorama, and the ground truth panorama, $I_g$, 
are then passed into the energy model which returns an energy function,

\setlength{\belowdisplayskip}{0pt} \setlength{\belowdisplayshortskip}{0pt}
\setlength{\abovedisplayskip}{0pt} \setlength{\abovedisplayshortskip}{0pt}
\begin{equation*}
E_\Theta (\mathbf{f}, \mathbf{I_{g^*}})) = \frac{1}{\sigma^2} \mathbf{f}^T 
\mathbf{f} - 
\mathbf{b}^T \mathbf{f} \\ - \sum_{i}^{} \mathrm{log}(1 + e^{W_i^T 
\mathbf{I_{g^*}} + b_i}),
\end{equation*}

\noindent similar to the free energy of a Gaussian Restricted Boltzmann Machine 
(RBM). Batch normalization~\cite{ioffe2015batch} is applied in every layer of 
both models, except for the final layers. ReLU activations are used throughout 
the generator and Leaky ReLU, with leak parameter $\alpha = 0.2$, are used in 
the energy model. The models are updated in an alternating fashion, where the 
generator is updated twice for every update of the energy 
model. Both the generator and energy model are optimized using the Adam 
optimizer, with moment parameters $\beta_1 = 0.5$ and 
$\beta_2 = 0.999$. We train using batch sizes of 32 for 30 epochs. 

Example outputs generated by our network are shown in \figref{gan}. Each row 
contains an aerial image (left), its respective ground panorama (top-right), 
and our prediction of the ground scene layout (bottom-right), which would 
ideally be the same as its above image. The network has learned the most common 
features, such as roads and their orientations, as well as trees and grass. 
However, it has difficulty hallucinating buildings and the sky, which is likely 
caused by highly variable appearance factors.

We note that the resolution of the synthesized ground-level panoramas is much 
lower than the original panorama, however adversarial generation of 
high-resolution images is an active area of research. We expect that in the 
near future we will be able to use our learned features in a similar manner to 
generate full-resolution panoramas. Additionally, algorithmic improvements to 
our ground image segmentation method would provide more photo-realistic 
predictions.

\section{Conclusion}

We introduced a novel strategy for using automatically labeled ground
images as a form of weak supervision for learning to understand aerial
images. The key is to simultaneously learn to extract features from
the aerial image and learn to map from the aerial to the ground image.
We demonstrated that by using this process we are able to
automatically extract semantically meaningful features from aerial
imagery, refine these to obtain more accurate pixel-level labeling of
aerial imagery, estimate the location and orientation of a ground
image, and synthesize novel ground-level views.  The proposed
technique is equally applicable to other forms of imagery, including
NIR, multispectral, and hyperspectral. For future work, we plan to
explore richer ground image annotation methods to explore the limits
of what is predictable about a ground-level view from an aerial view.

%
%
%
%
%

\ifcvprfinal
\section*{Acknowledgements}
We gratefully acknowledge the support of NSF CAREER grant
(IIS-1553116), a Google Faculty Research Award, and an AWS Research
Education grant.
\fi

{\small
\bibliographystyle{ieee}
\bibliography{../references}
}

\clearpage
\graphicspath{{appendix/figs/}}
\section*{Appendix}
\begin{figure} [b]
\makebox[0pt][l]{%
 \begin{minipage}{\textwidth}
  \centering 
  \includegraphics[width=\linewidth]{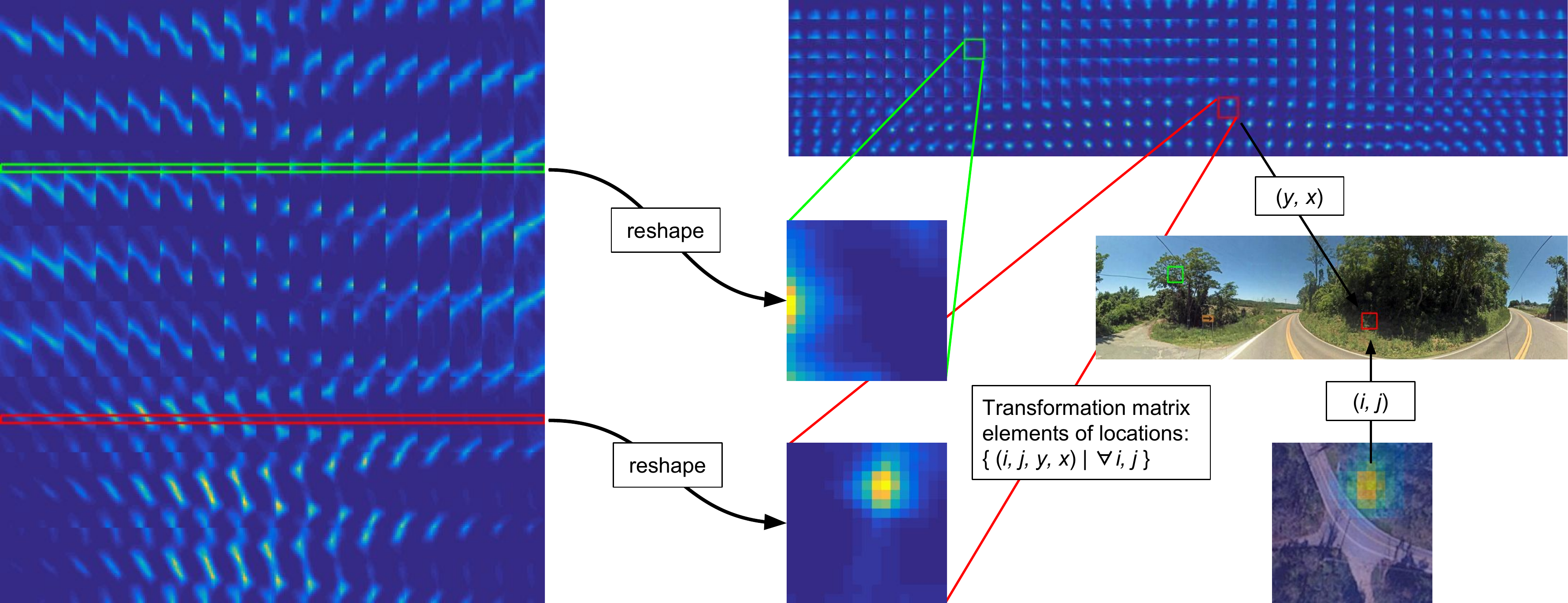} 
  \caption{Visualization of the transformation matrix. (left) 
transformation matrix, $M$; (right-top) An alternative visualization, 
$M'$, of the transformation matrix. $M'$ contains $h_g \times w_g$ 
cells (square heat maps). Each cell, 
$m_{yx}'$, is reshaped to size $h_a \times w_a$, from one row of $M$ 
that corresponds to locations, $\{(i, j, y, x) \mid 
\forall{i, j} \}$. We also present the aerial image (overlapped 
with $m_{yx}'$) and the ground 
image to illustrate how the hot spot of $m_{yx}'$ corresponds to 
the location, $(y, x)$, on the 
ground image.} 
\label{fig:app:transfmat} 

 \end{minipage}}
\end{figure}

\begin{figure*}[t]
  \centering
  \newlength{\aw}
  \newlength{\gw}
  \newlength{\vgap}
  \setlength{\aw}{45pt}
  \setlength{\gw}{108pt}
  \setlength{\vgap}{11pt}
   \begin{minipage}[b]{\aw}
   \includegraphics[width=\textwidth]{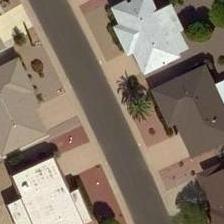}
   \includegraphics[width=\textwidth]{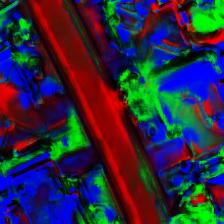}
   \end{minipage} %
   \begin{minipage}[b]{\gw}
   \includegraphics[width=\textwidth, height=\aw]{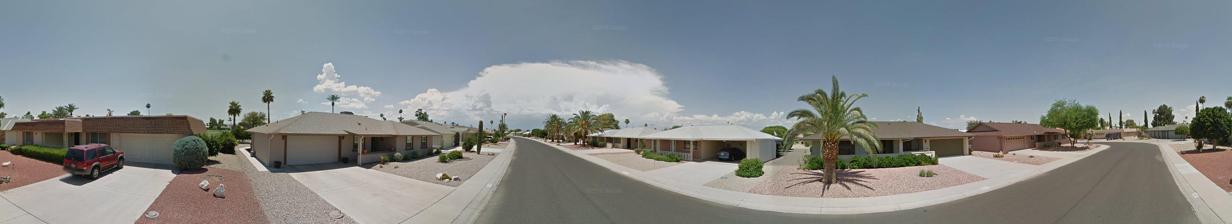}
   \includegraphics[width=\textwidth, height=\aw]{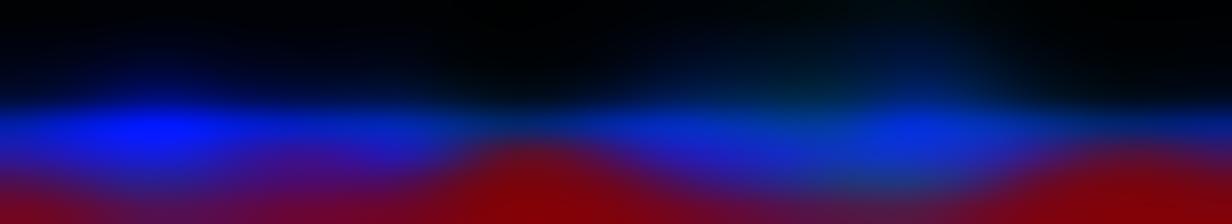}
   \end{minipage} \hfill %
   \begin{minipage}[b]{\aw}
   \includegraphics[width=\textwidth]{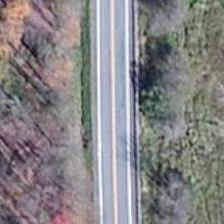}
   \includegraphics[width=\textwidth]{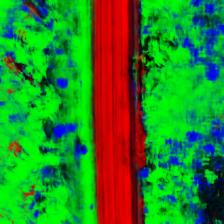}
   \end{minipage} %
   \begin{minipage}[b]{\gw}
   \includegraphics[width=\textwidth, height=\aw]{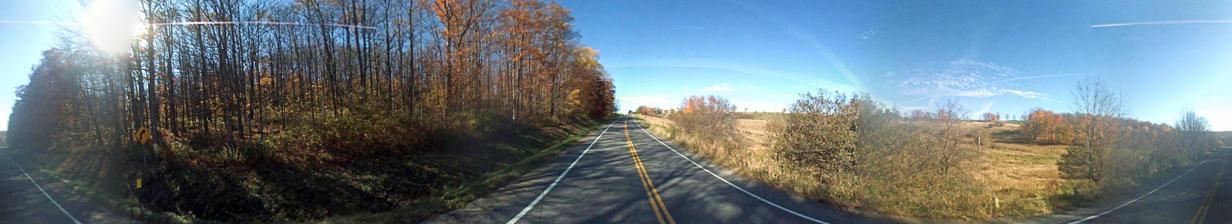}
   \includegraphics[width=\textwidth, height=\aw]{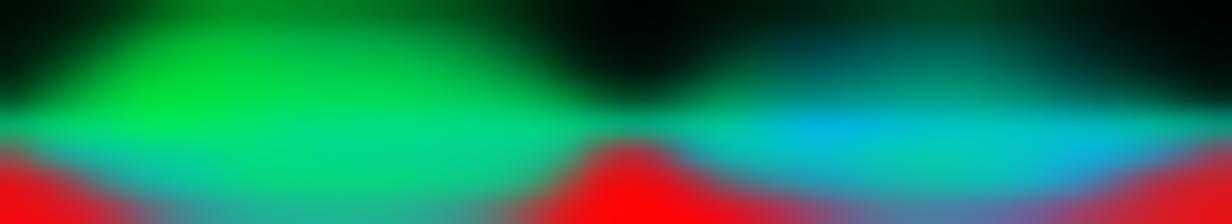}
   \end{minipage} \hfill %
   \begin{minipage}[b]{\aw}
   \includegraphics[width=\textwidth]{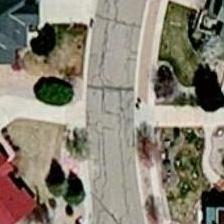}
   \includegraphics[width=\textwidth]{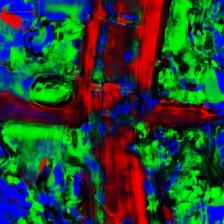}
   \end{minipage} %
   \begin{minipage}[b]{\gw}
   \includegraphics[width=\textwidth, height=\aw]{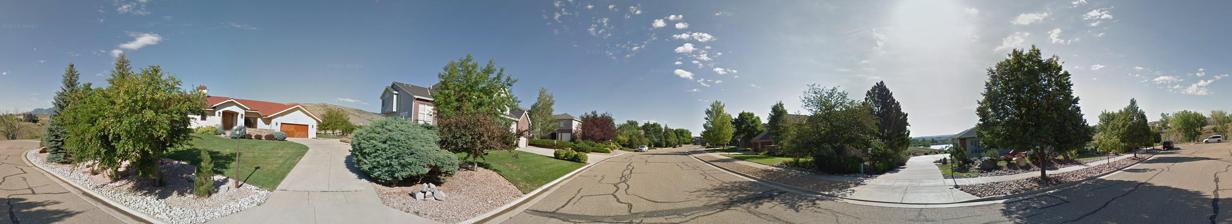}
   \includegraphics[width=\textwidth, height=\aw]{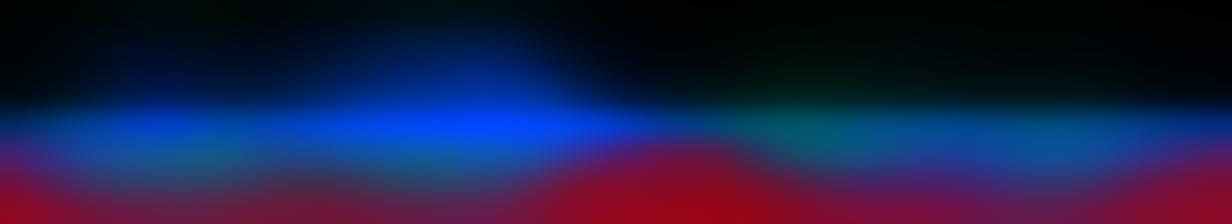}
   \end{minipage} \hfill %
   \vspace{\vgap}
   \begin{minipage}[b]{\aw}
   \includegraphics[width=\textwidth]{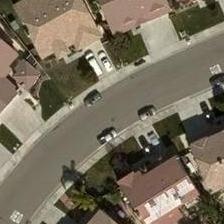}
   \includegraphics[width=\textwidth]{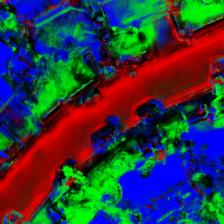}
   \end{minipage} %
   \begin{minipage}[b]{\gw}
   \includegraphics[width=\textwidth, height=\aw]{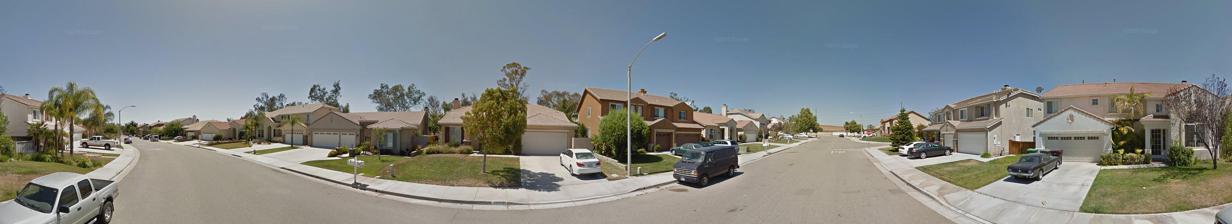}
   \includegraphics[width=\textwidth, height=\aw]{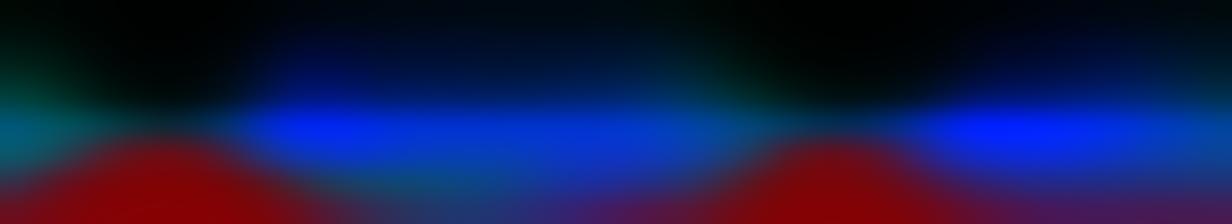}
   \end{minipage} \hfill %
   \begin{minipage}[b]{\aw}
   \includegraphics[width=\textwidth]{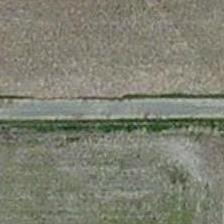}
   \includegraphics[width=\textwidth]{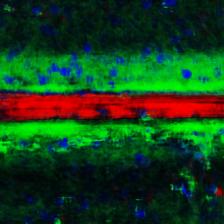}
   \end{minipage} %
   \begin{minipage}[b]{\gw}
   \includegraphics[width=\textwidth, height=\aw]{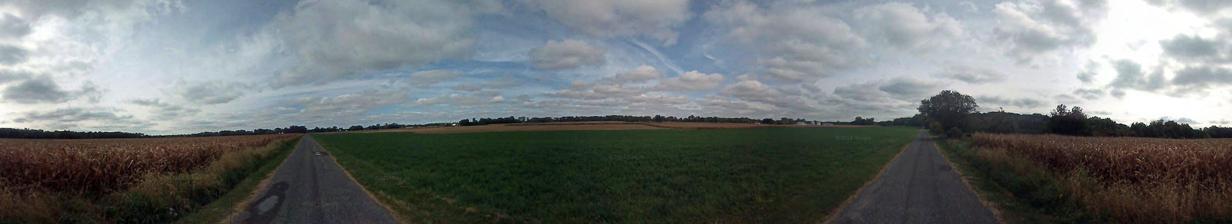}
   \includegraphics[width=\textwidth, height=\aw]{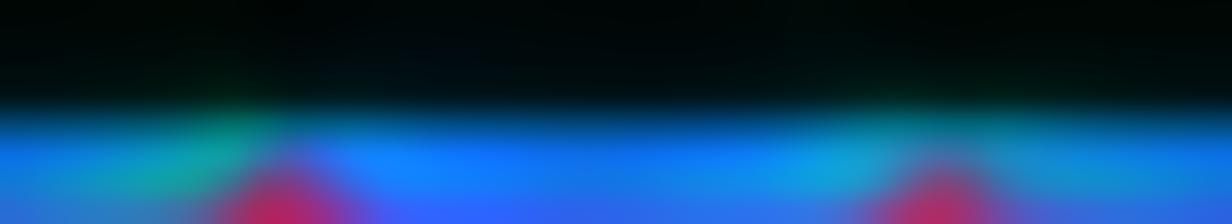}
   \end{minipage} \hfill %
   \begin{minipage}[b]{\aw}
   \includegraphics[width=\textwidth]{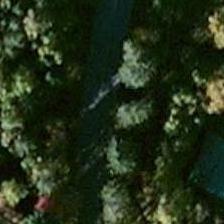}
   \includegraphics[width=\textwidth]{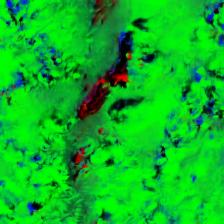}
   \end{minipage} %
   \begin{minipage}[b]{\gw}
   \includegraphics[width=\textwidth, height=\aw]{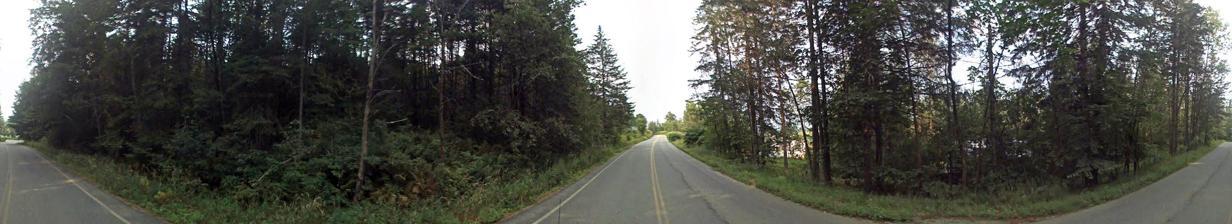}
   \includegraphics[width=\textwidth, height=\aw]{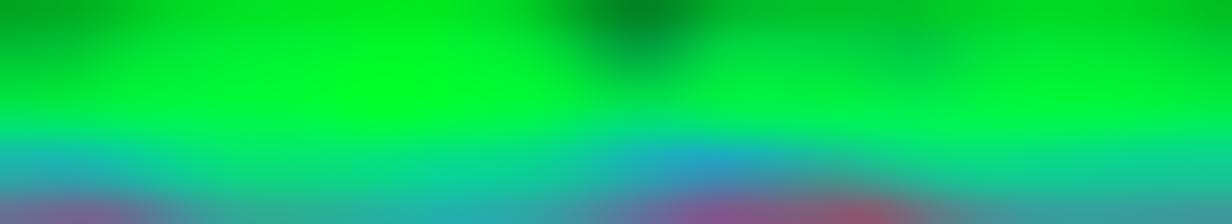}
   \end{minipage} \hfill %
   \vspace{\vgap}
   \begin{minipage}[b]{\aw}
   \includegraphics[width=\textwidth]{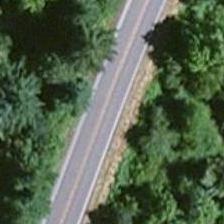}
   \includegraphics[width=\textwidth]{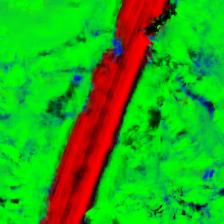}
   \end{minipage} %
   \begin{minipage}[b]{\gw}
   \includegraphics[width=\textwidth, height=\aw]{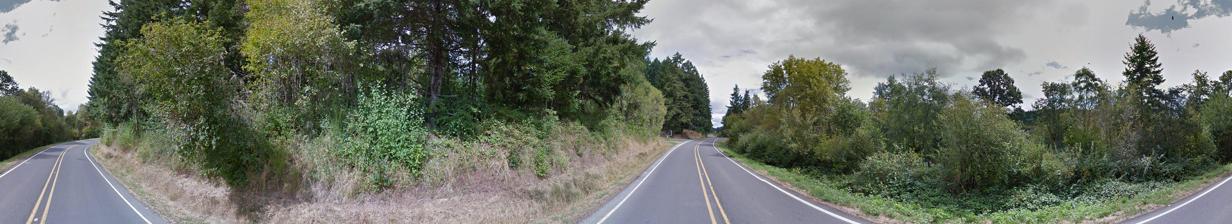}
   \includegraphics[width=\textwidth, height=\aw]{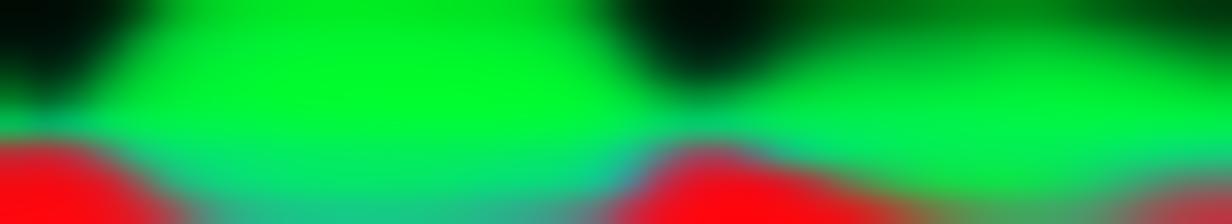}
   \end{minipage} \hfill %
   \begin{minipage}[b]{\aw}
   \includegraphics[width=\textwidth]{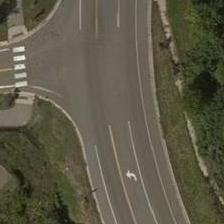}
   \includegraphics[width=\textwidth]{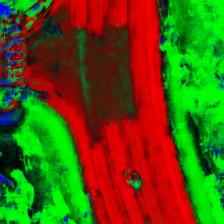}
   \end{minipage} %
   \begin{minipage}[b]{\gw}
   \includegraphics[width=\textwidth, height=\aw]{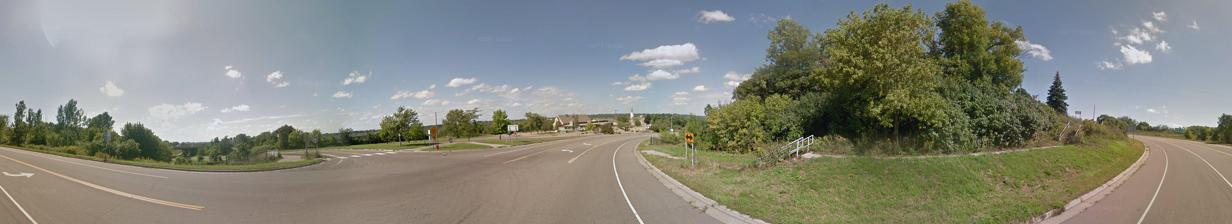}
   \includegraphics[width=\textwidth, height=\aw]{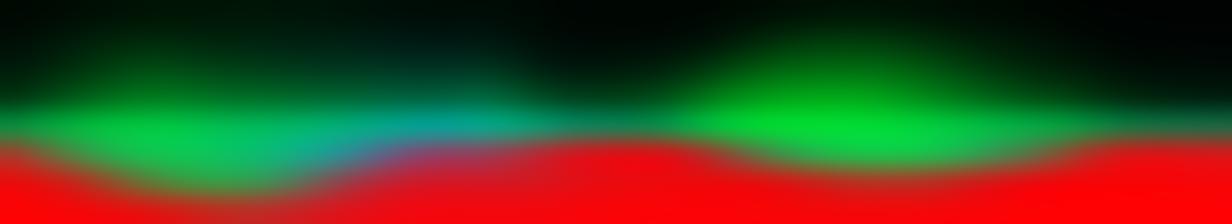}
   \end{minipage} \hfill %
   \begin{minipage}[b]{\aw}
   \includegraphics[width=\textwidth]{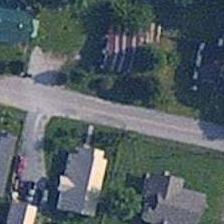}
   \includegraphics[width=\textwidth]{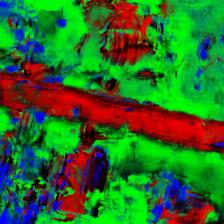}
   \end{minipage} %
   \begin{minipage}[b]{\gw}
   \includegraphics[width=\textwidth, height=\aw]{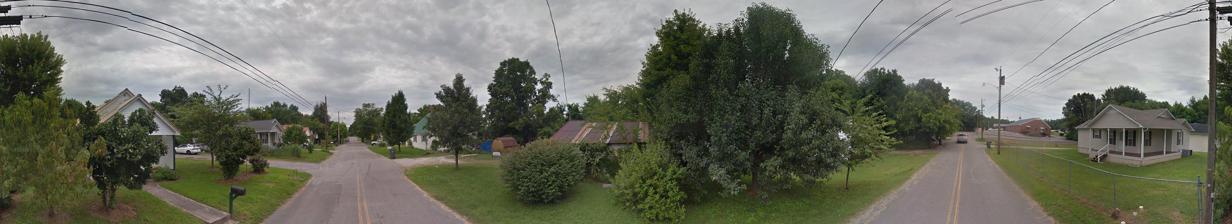}
   \includegraphics[width=\textwidth, height=\aw]{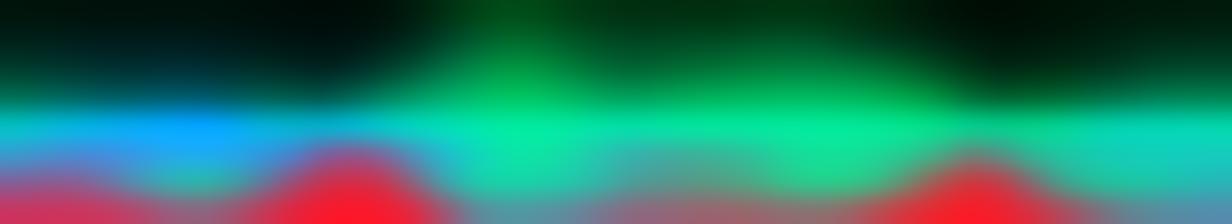}
   \end{minipage} \hfill %
   \vspace{\vgap}
   \begin{minipage}[b]{\aw}
   \includegraphics[width=\textwidth]{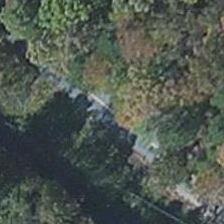}
   \includegraphics[width=\textwidth]{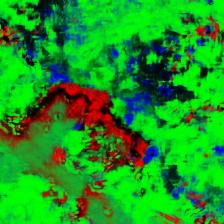}
   \end{minipage} %
   \begin{minipage}[b]{\gw}
   \includegraphics[width=\textwidth, height=\aw]{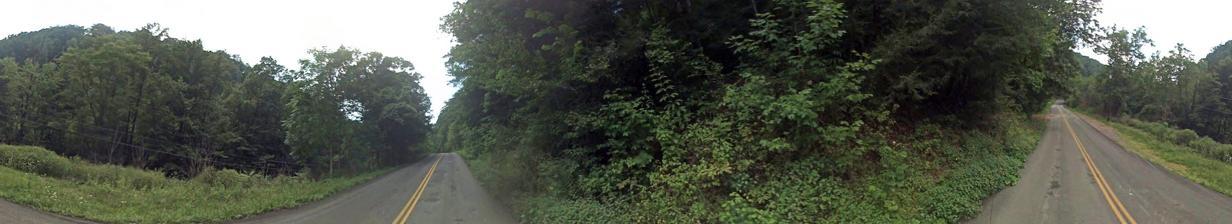}
   \includegraphics[width=\textwidth, height=\aw]{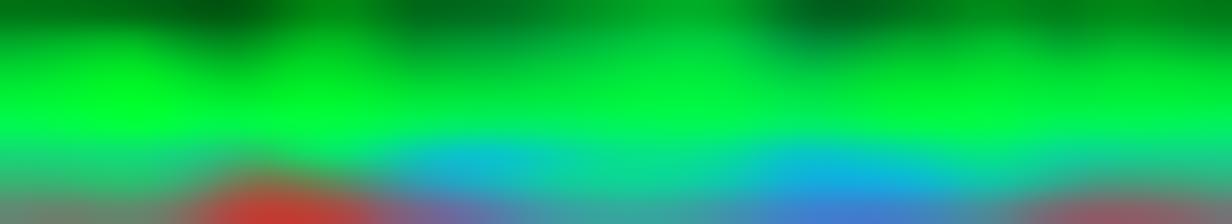}
   \end{minipage} \hfill %
   \begin{minipage}[b]{\aw}
   \includegraphics[width=\textwidth]{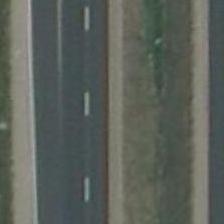}
   \includegraphics[width=\textwidth]{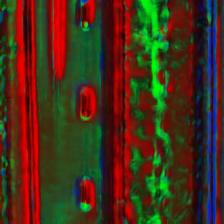}
   \end{minipage} %
   \begin{minipage}[b]{\gw}
   \includegraphics[width=\textwidth, height=\aw]{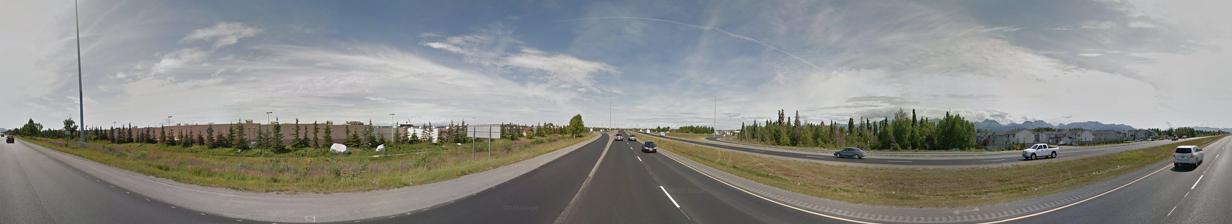}
   \includegraphics[width=\textwidth, height=\aw]{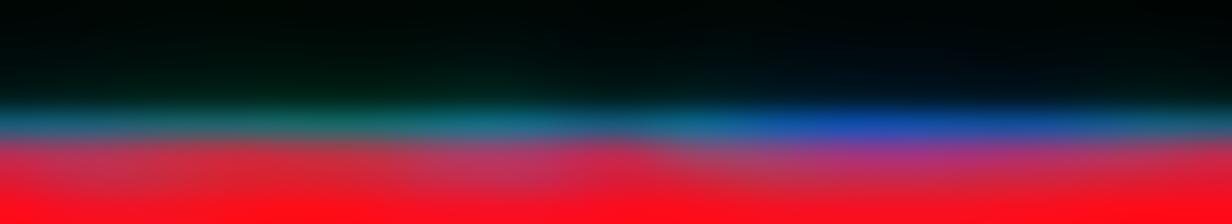}
   \end{minipage} \hfill %
   \begin{minipage}[b]{\aw}
   \includegraphics[width=\textwidth]{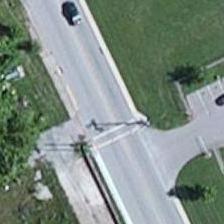}
   \includegraphics[width=\textwidth]{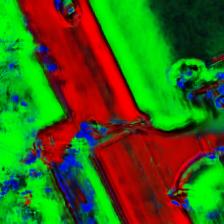}
   \end{minipage} %
   \begin{minipage}[b]{\gw}
   \includegraphics[width=\textwidth, height=\aw]{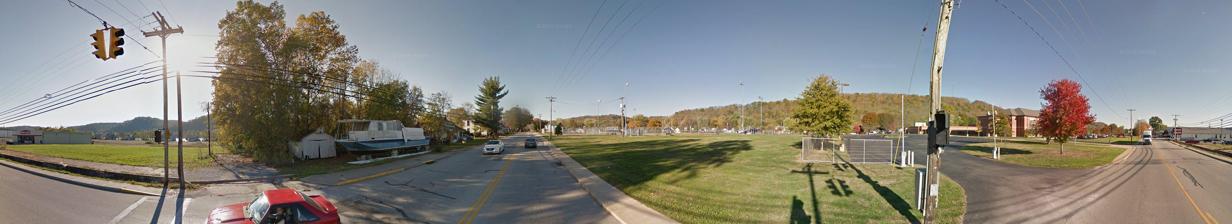}
   \includegraphics[width=\textwidth, height=\aw]{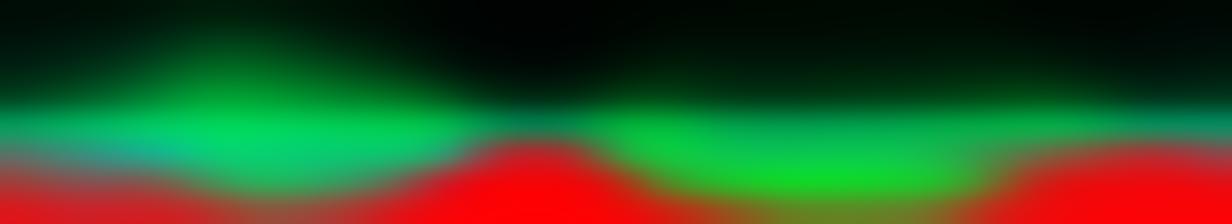}
   \end{minipage} \hfill %
   \vspace{\vgap}
   \begin{minipage}[b]{\aw}
   \includegraphics[width=\textwidth]{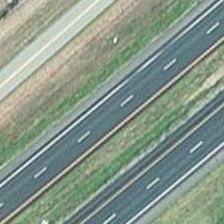}
   \includegraphics[width=\textwidth]{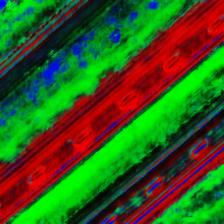}
   \end{minipage} %
   \begin{minipage}[b]{\gw}
   \includegraphics[width=\textwidth, height=\aw]{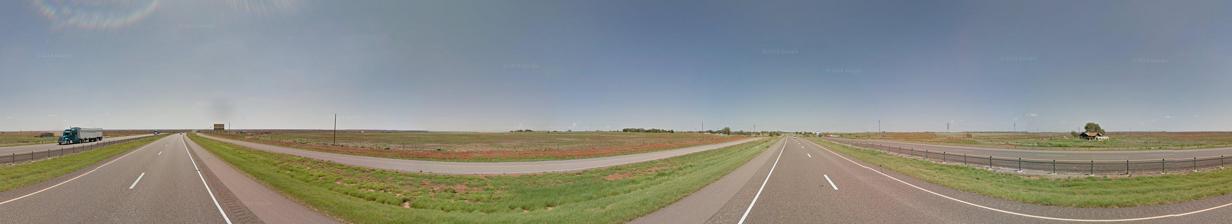}
   \includegraphics[width=\textwidth, height=\aw]{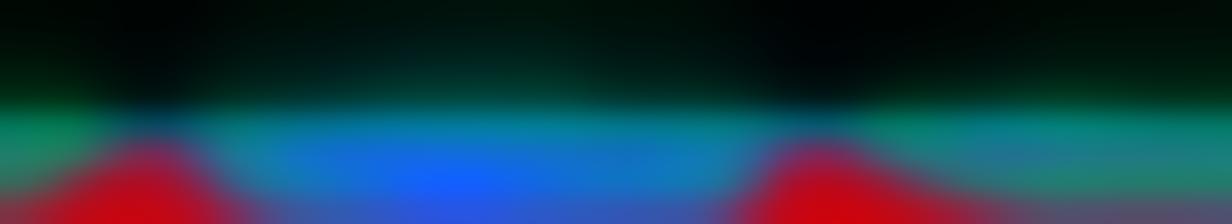}
   \end{minipage} \hfill %
   \begin{minipage}[b]{\aw}
   \includegraphics[width=\textwidth]{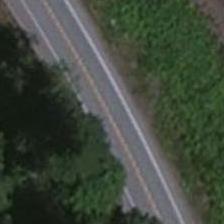}
   \includegraphics[width=\textwidth]{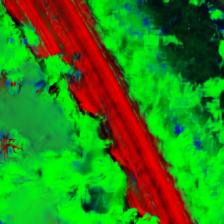}
   \end{minipage} %
   \begin{minipage}[b]{\gw}
   \includegraphics[width=\textwidth, height=\aw]{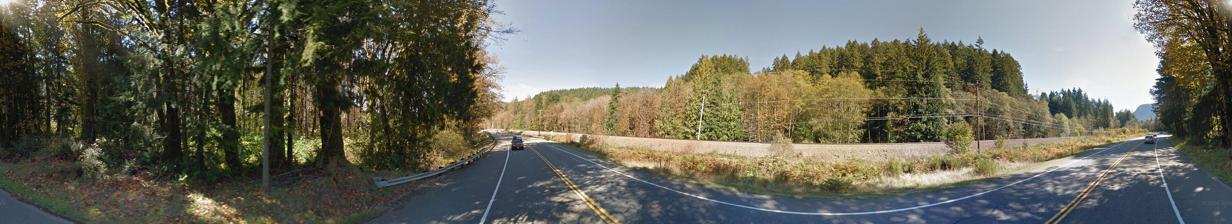}
   \includegraphics[width=\textwidth, height=\aw]{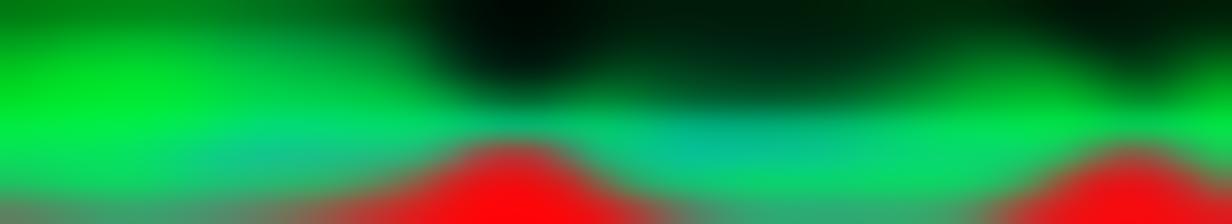}
   \end{minipage} \hfill %
   \begin{minipage}[b]{\aw}
   \includegraphics[width=\textwidth]{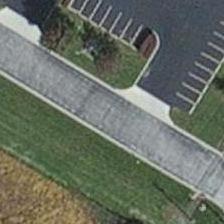}
   \includegraphics[width=\textwidth]{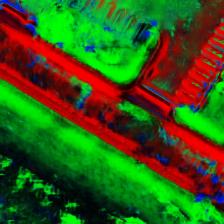}
   \end{minipage} %
   \begin{minipage}[b]{\gw}
   \includegraphics[width=\textwidth, height=\aw]{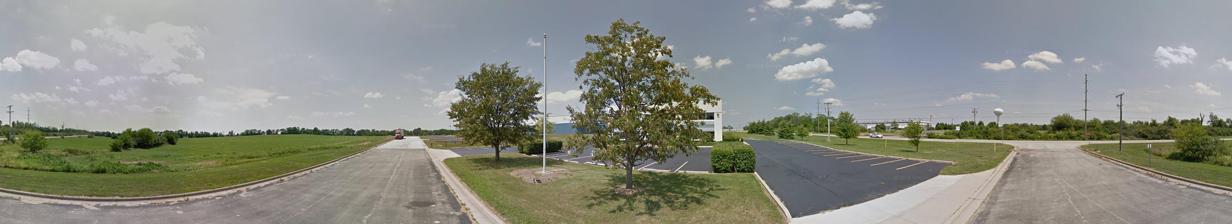}
   \includegraphics[width=\textwidth, height=\aw]{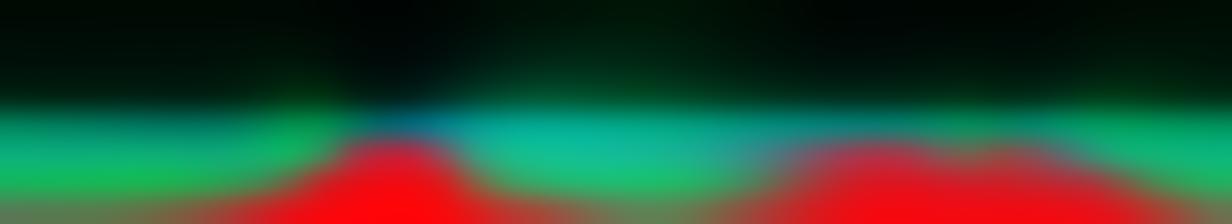}
   \end{minipage} \hfill %
   \vspace{\vgap}

  \caption{Randomly selected test image results from our weakly
  supervised learning method. The left column shows the aerial image
  (top) and the corresponding pixel-level labeling (bottom); The right
  column shows the ground image (top) of the same location and its
  pixel-level labeling (bottom) inferred by our model from the aerial
  image pixel-level labeling. We visualize three classes: {\em road}
  (red), {\em vegetation} (green), and {\em man-made} (blue).}
  \label{fig:app:weakly}
\end{figure*}

\begin{figure*}[h]
  \newlength{\aggap}
  \setlength{\aggap}{8pt}
  \setlength{\aheight}{93pt}
  \setlength{\gwidth}{60.65pt}
  \centering
  \begin{minipage}[b]{\gwidth}
  \includegraphics[width=\textwidth]{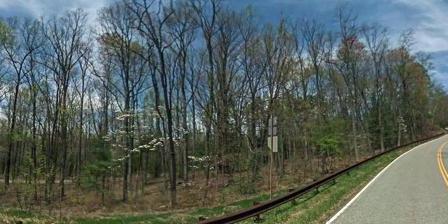}
  \includegraphics[width=\textwidth]{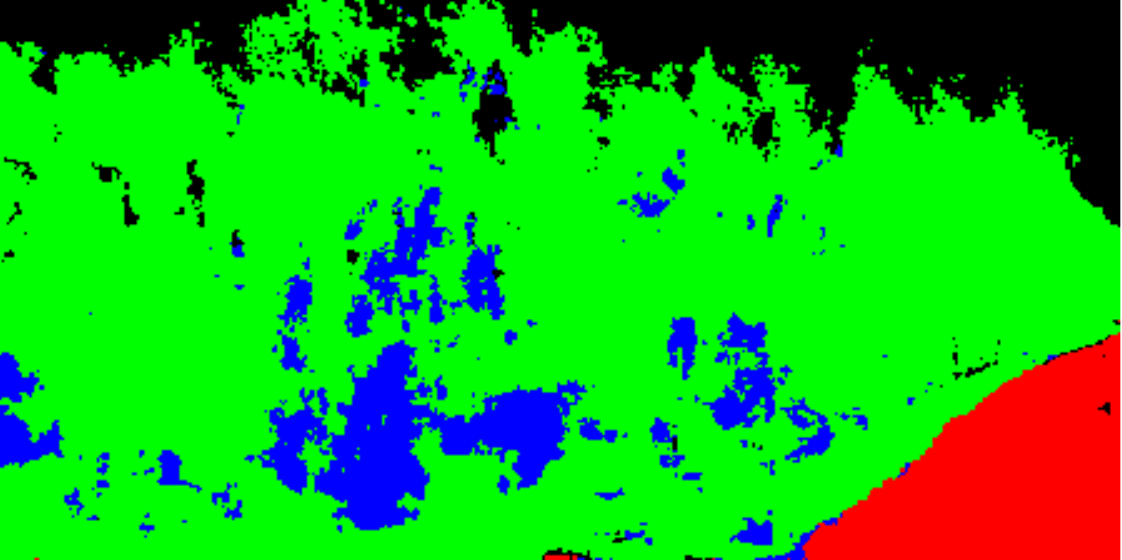}
  \includegraphics[width=\textwidth]{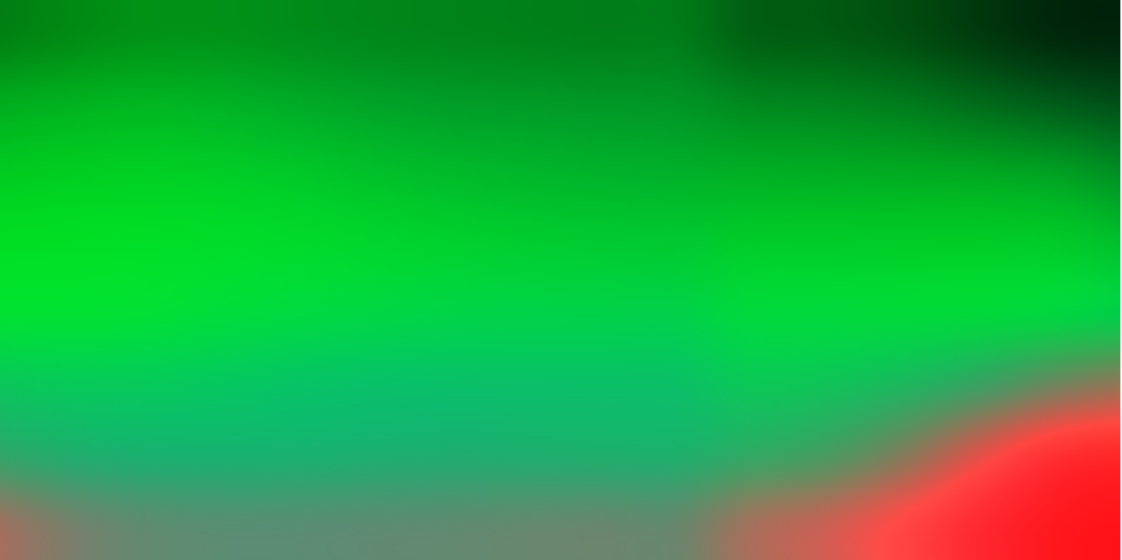}
  \end{minipage}
  \includegraphics[height=\aheight]{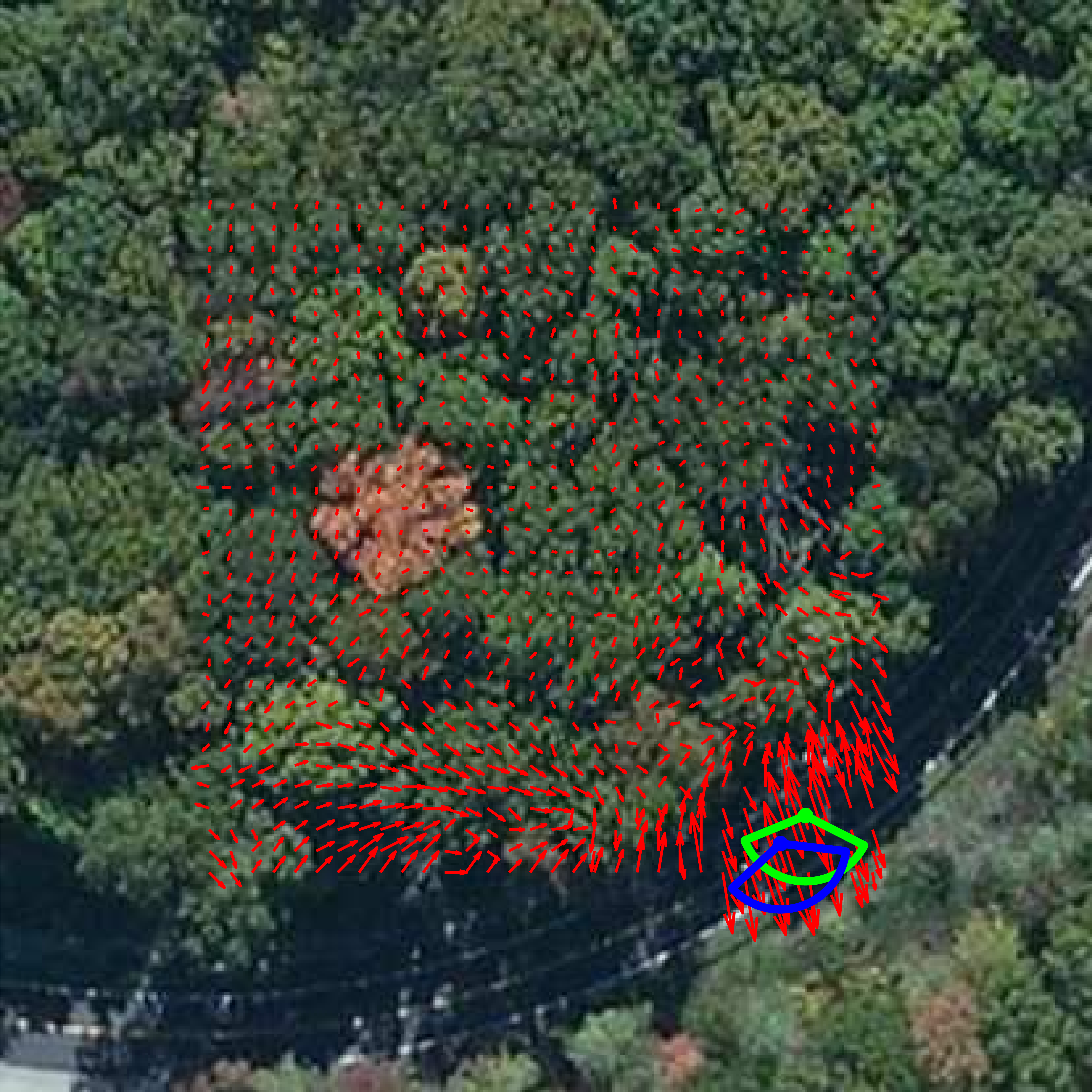} \hfill %
  \begin{minipage}[b]{\gwidth}
  \includegraphics[width=\textwidth]{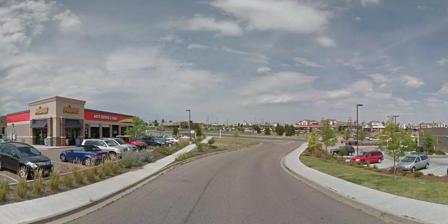}
  \includegraphics[width=\textwidth]{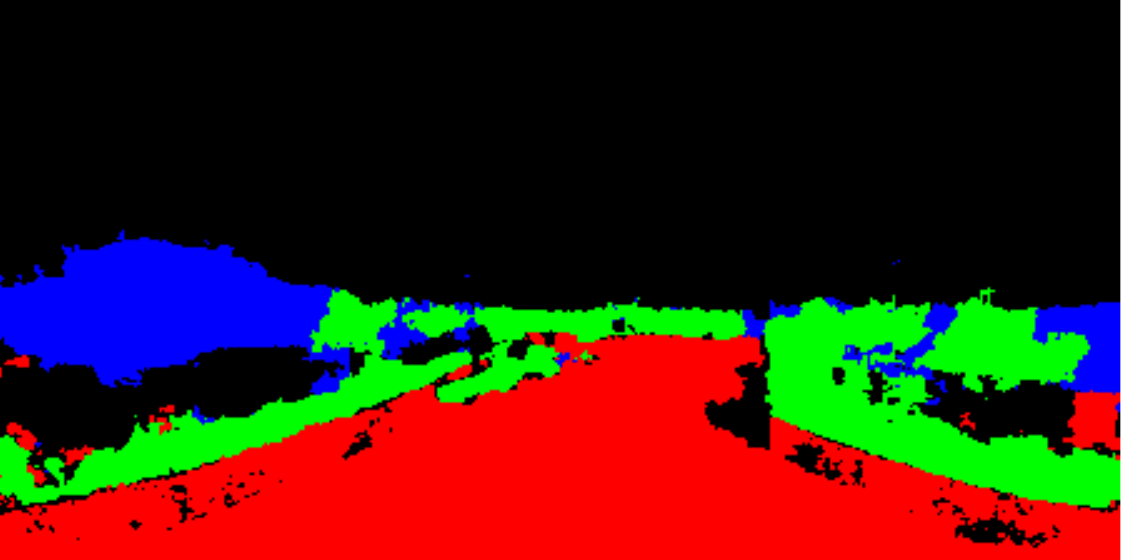}
  \includegraphics[width=\textwidth]{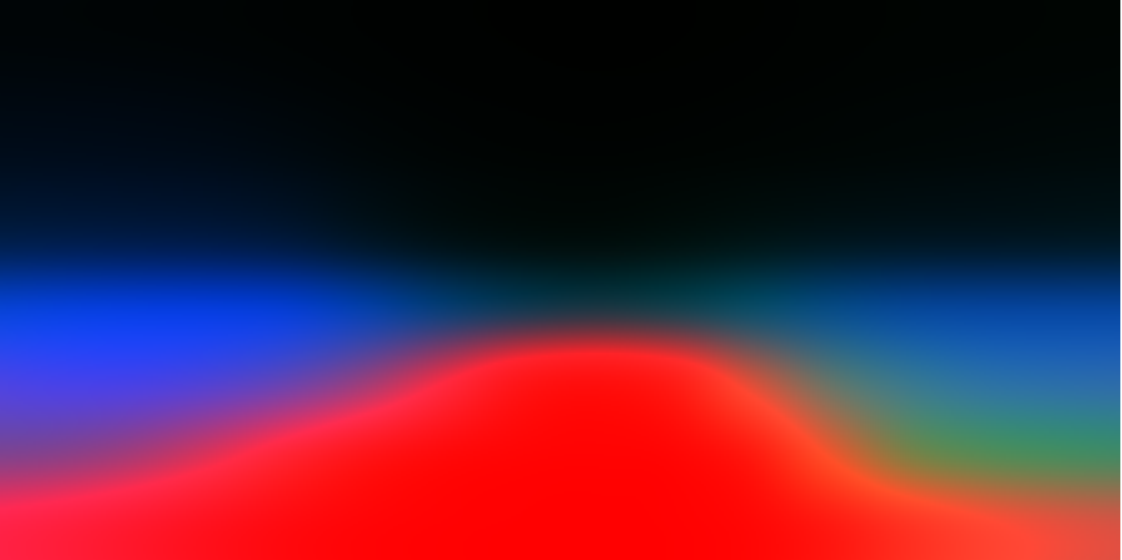}
  \end{minipage}
  \includegraphics[height=\aheight]{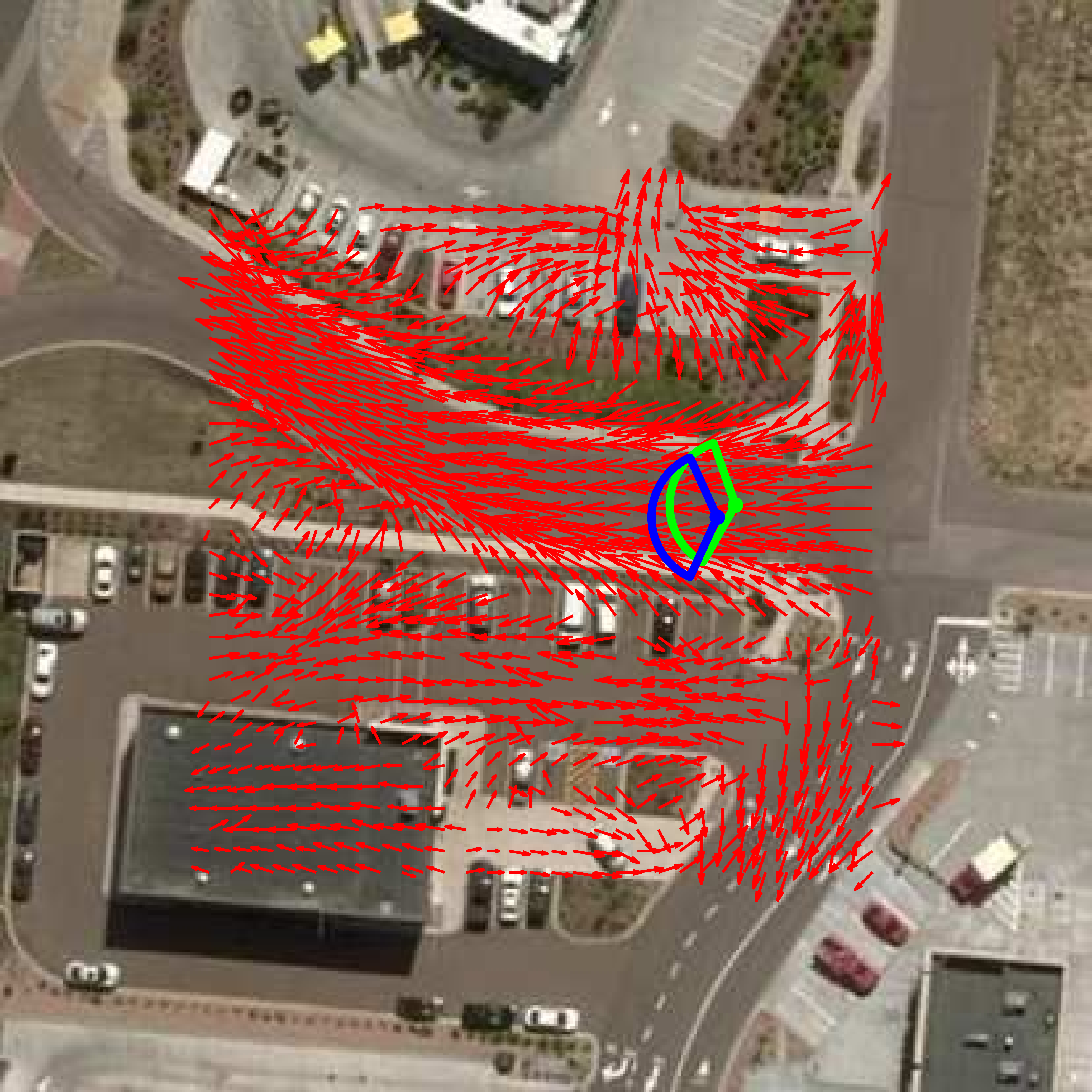} \hfill %
  \begin{minipage}[b]{\gwidth}
  \includegraphics[width=\textwidth]{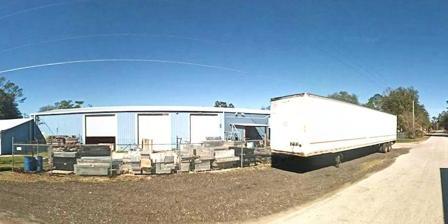}
  \includegraphics[width=\textwidth]{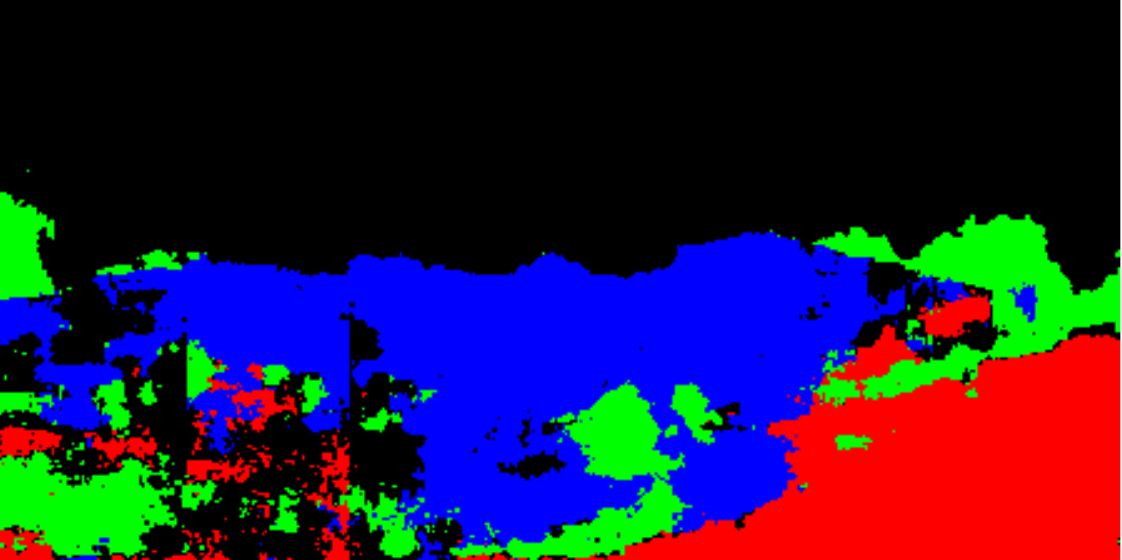}
  \includegraphics[width=\textwidth]{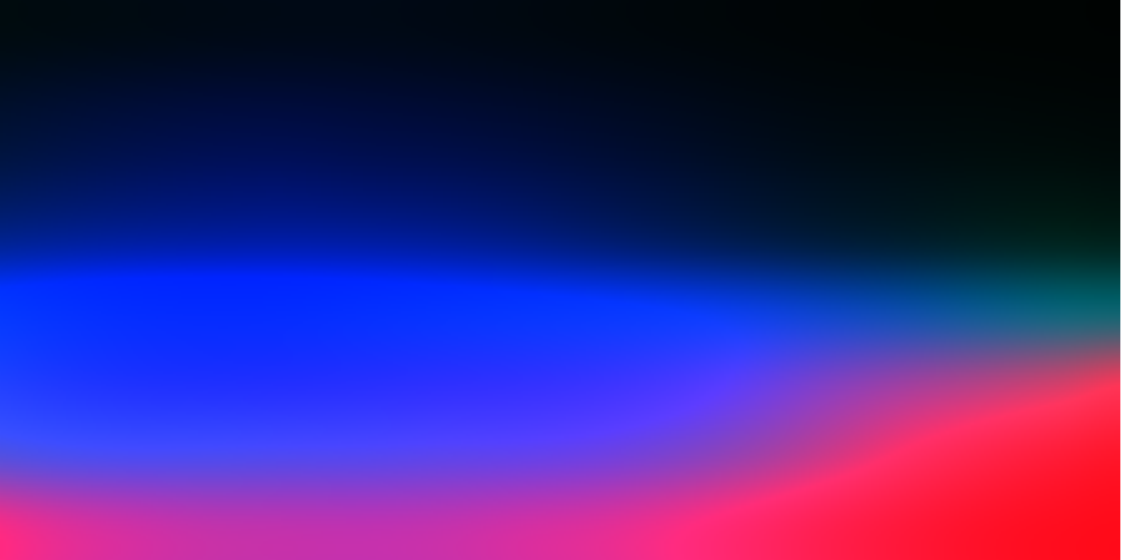}
  \end{minipage}
  \includegraphics[height=\aheight]{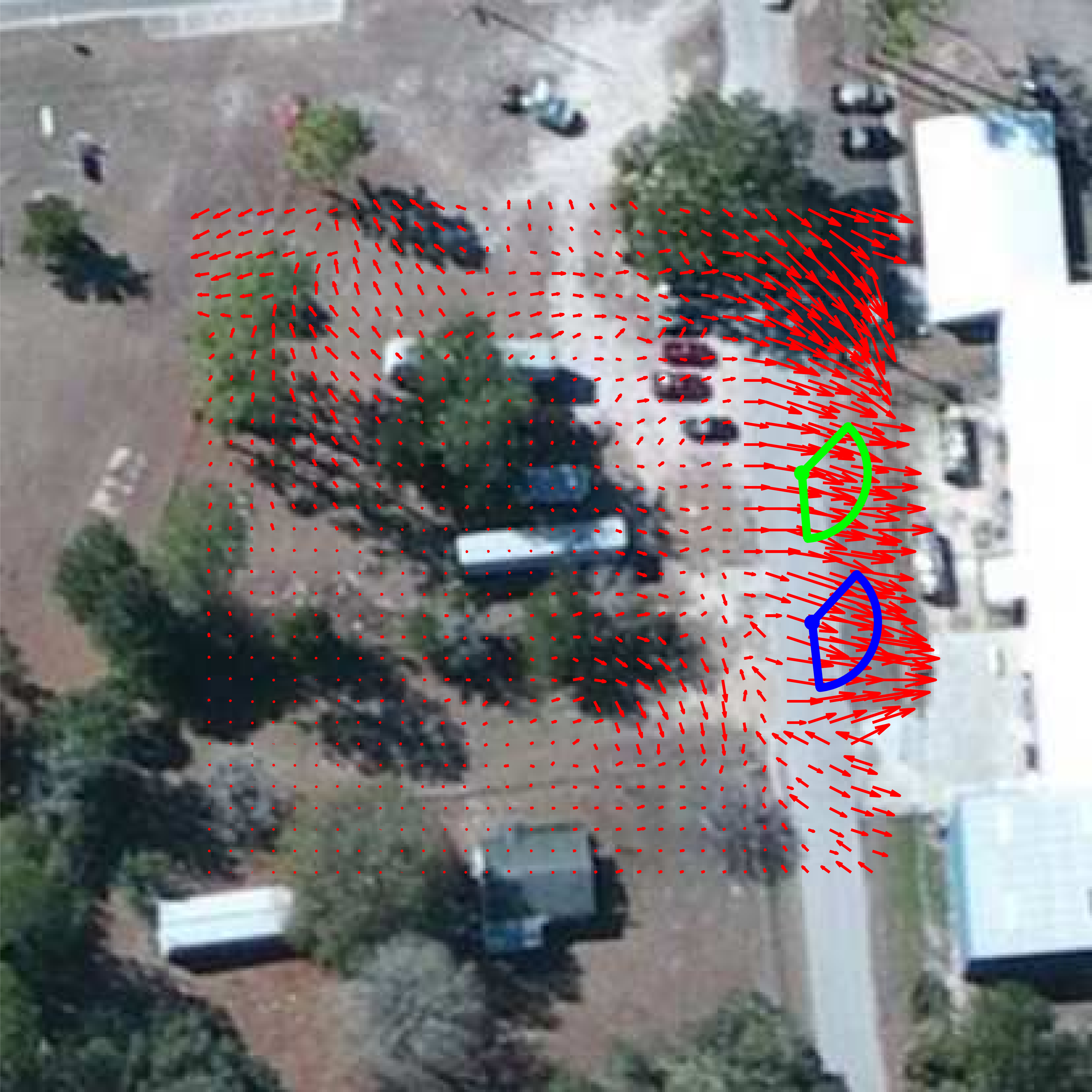} \hfill %
  \begin{minipage}[b]{\gwidth}
  \includegraphics[width=\textwidth]{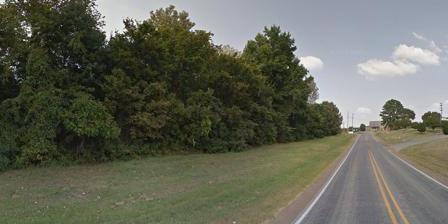}
  \includegraphics[width=\textwidth]{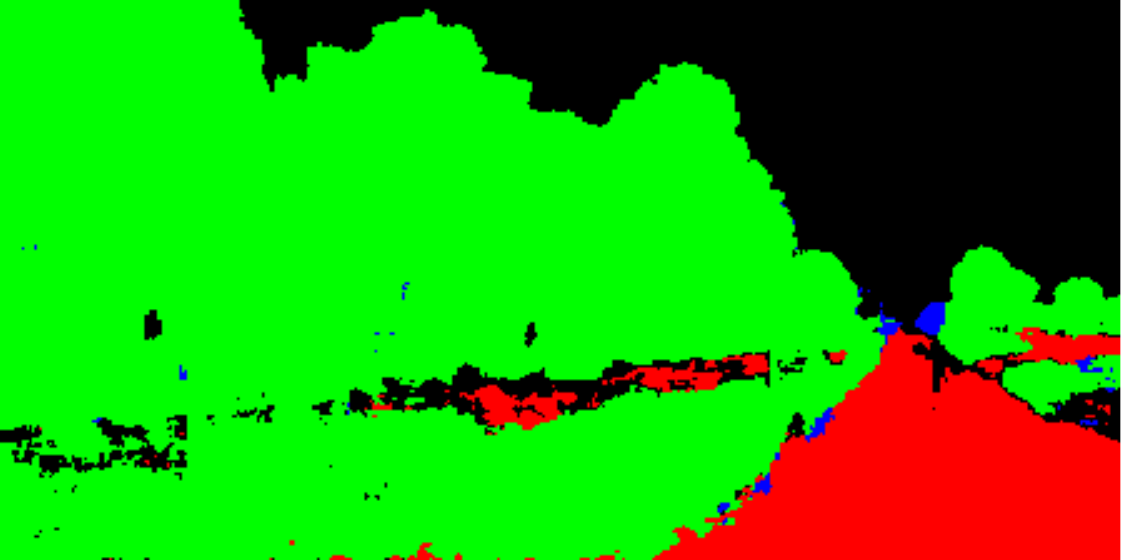}
  \includegraphics[width=\textwidth]{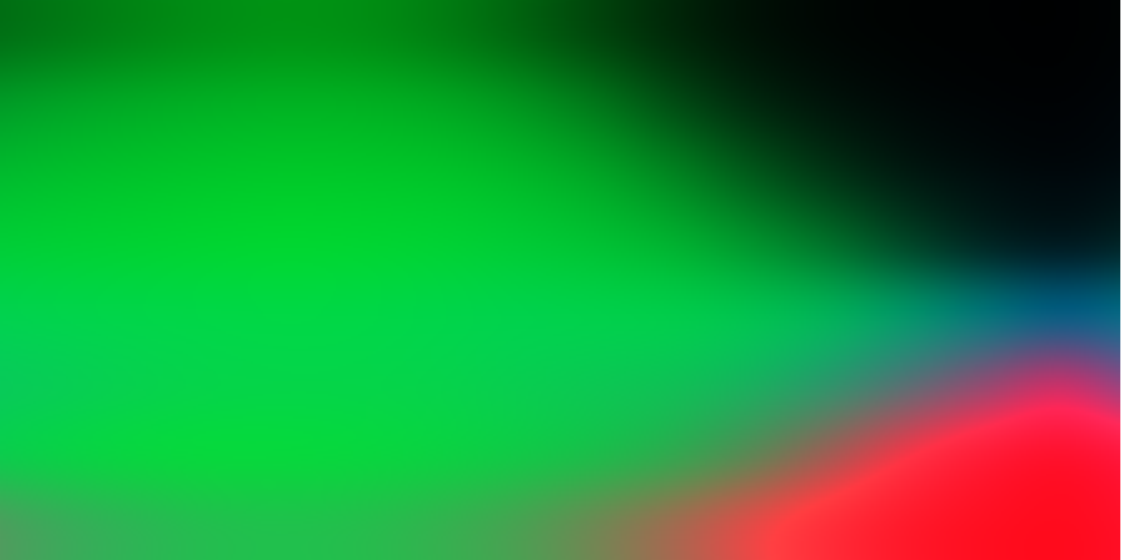}
  \end{minipage}
  \includegraphics[height=\aheight]{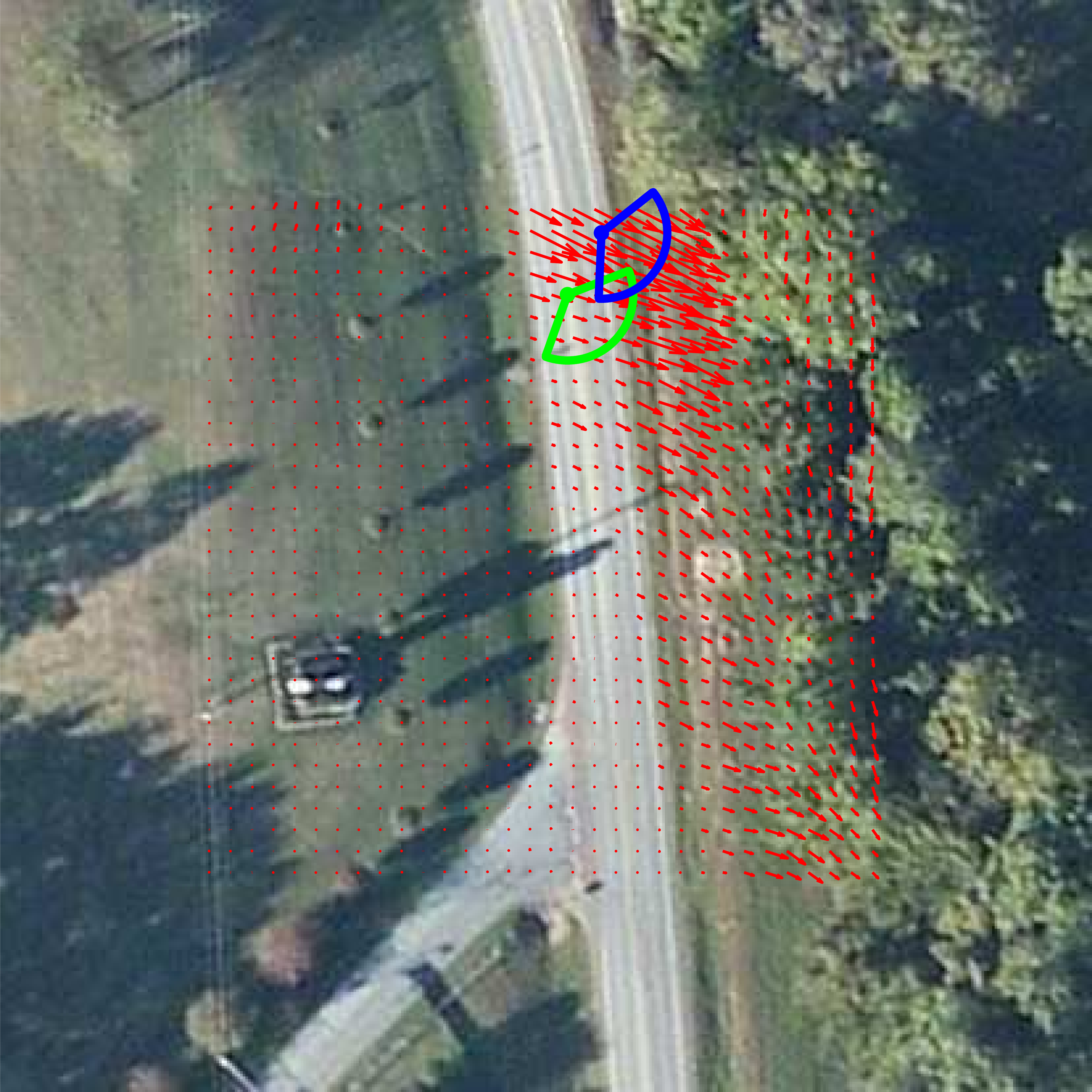} \hfill %
  \begin{minipage}[b]{\gwidth}
  \includegraphics[width=\textwidth]{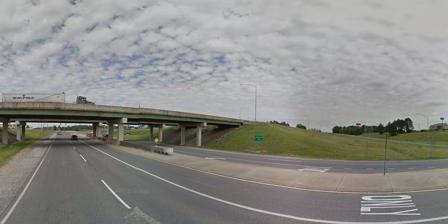}
  \includegraphics[width=\textwidth]{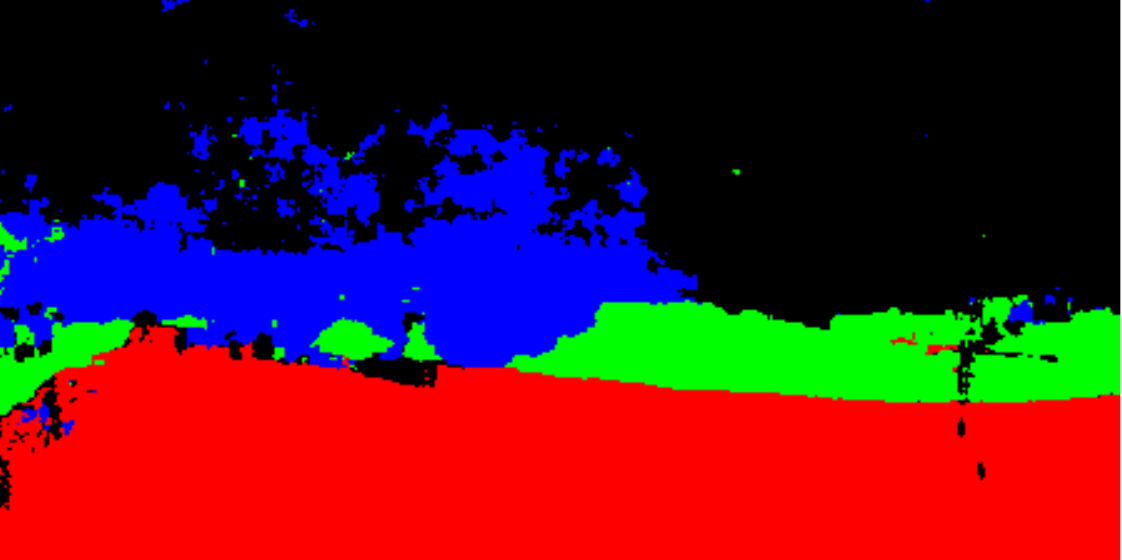}
  \includegraphics[width=\textwidth]{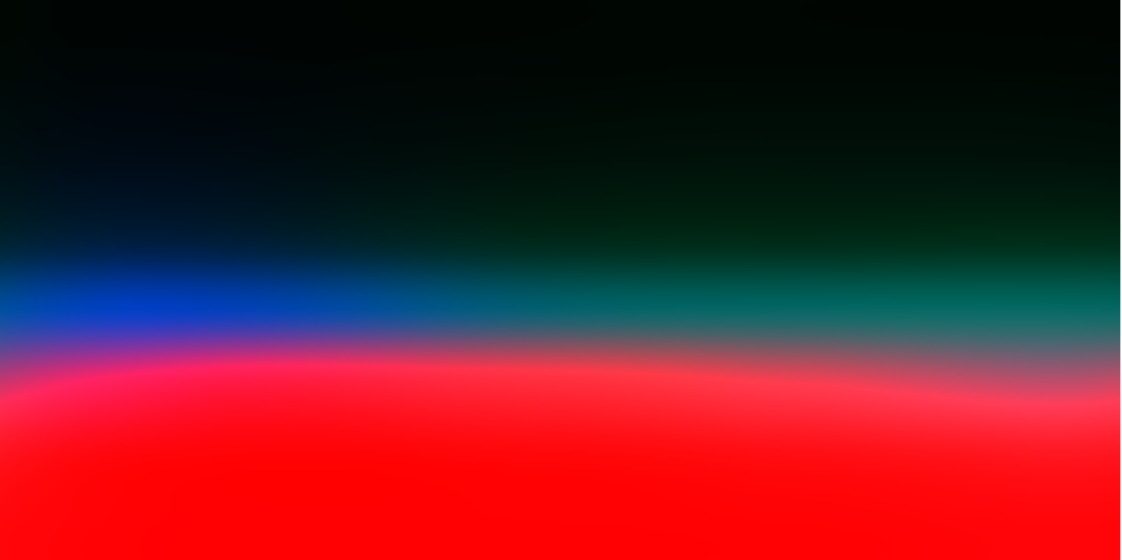}
  \end{minipage}
  \includegraphics[height=\aheight]{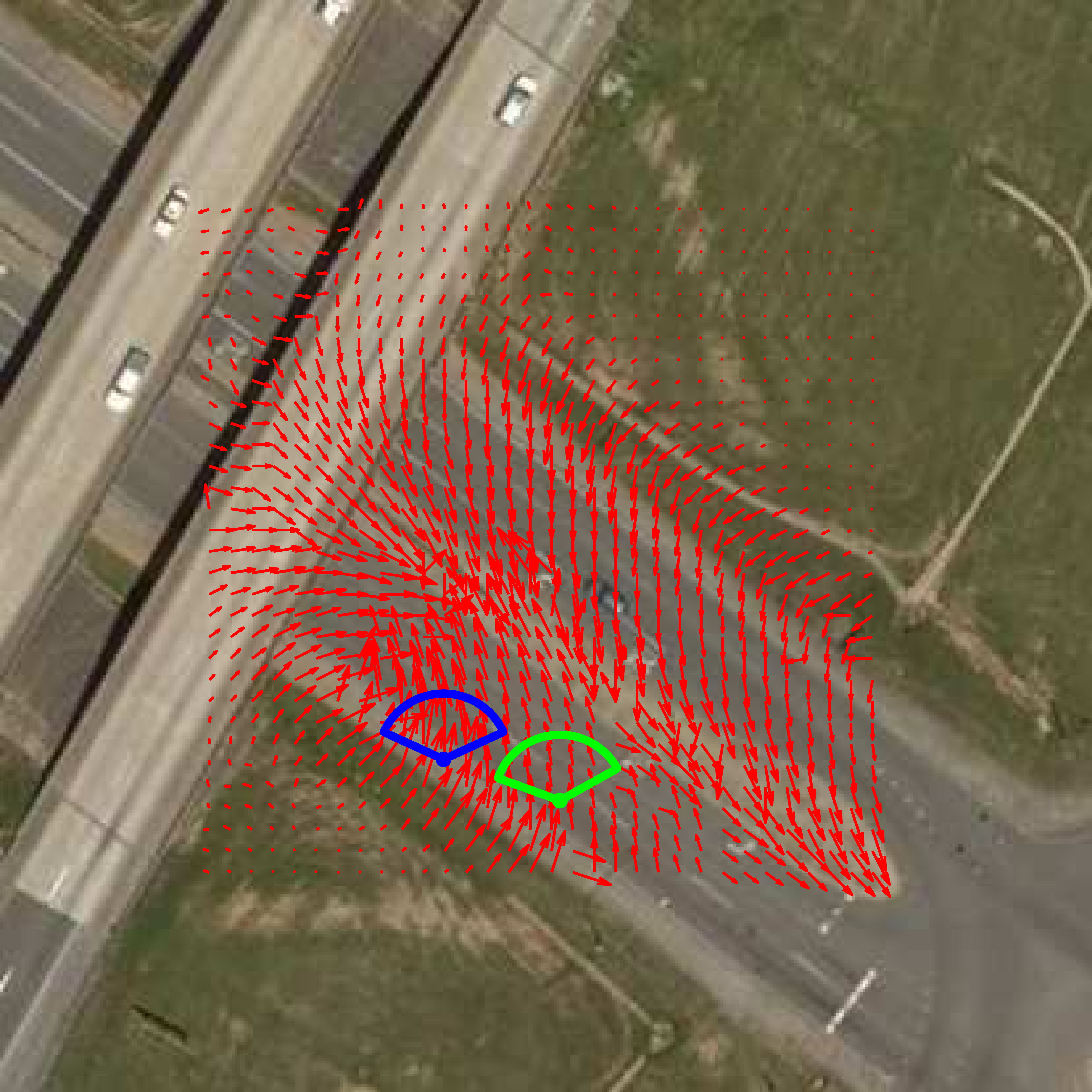} \hfill %
  \begin{minipage}[b]{\gwidth}
  \includegraphics[width=\textwidth]{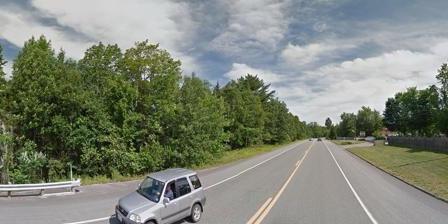}
  \includegraphics[width=\textwidth]{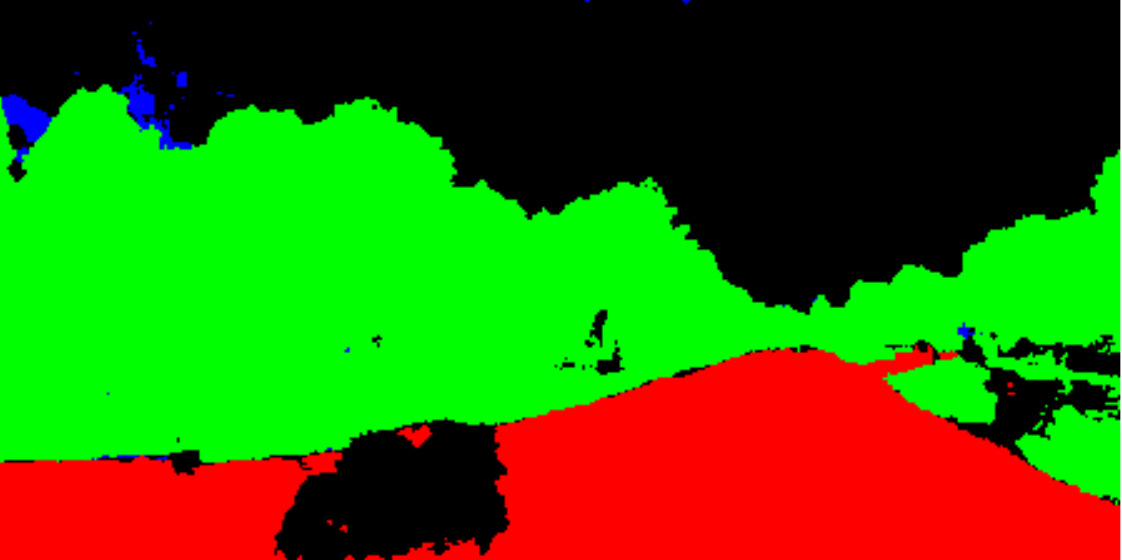}
  \includegraphics[width=\textwidth]{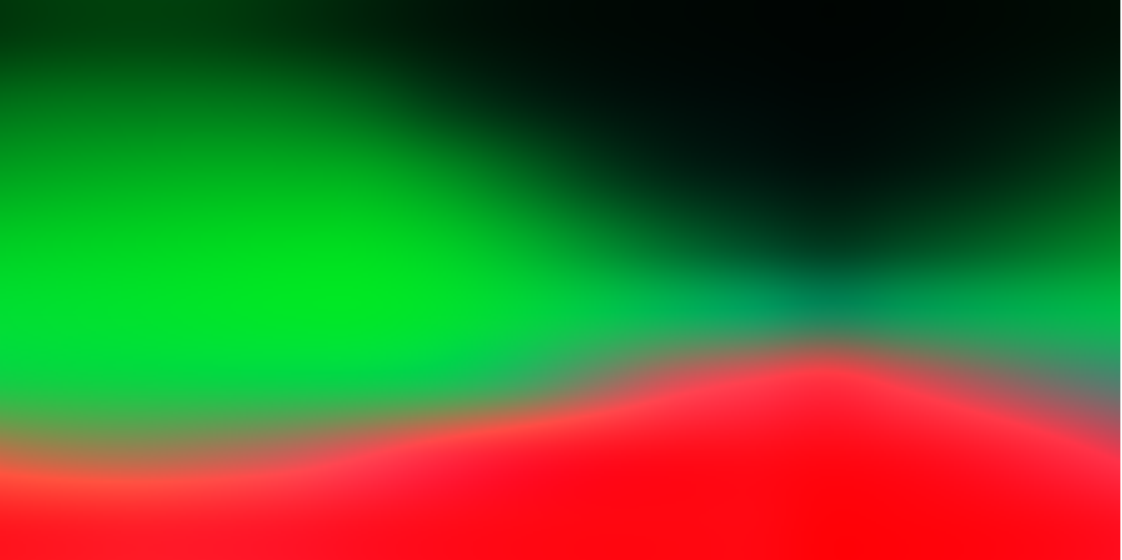}
  \end{minipage}
  \includegraphics[height=\aheight]{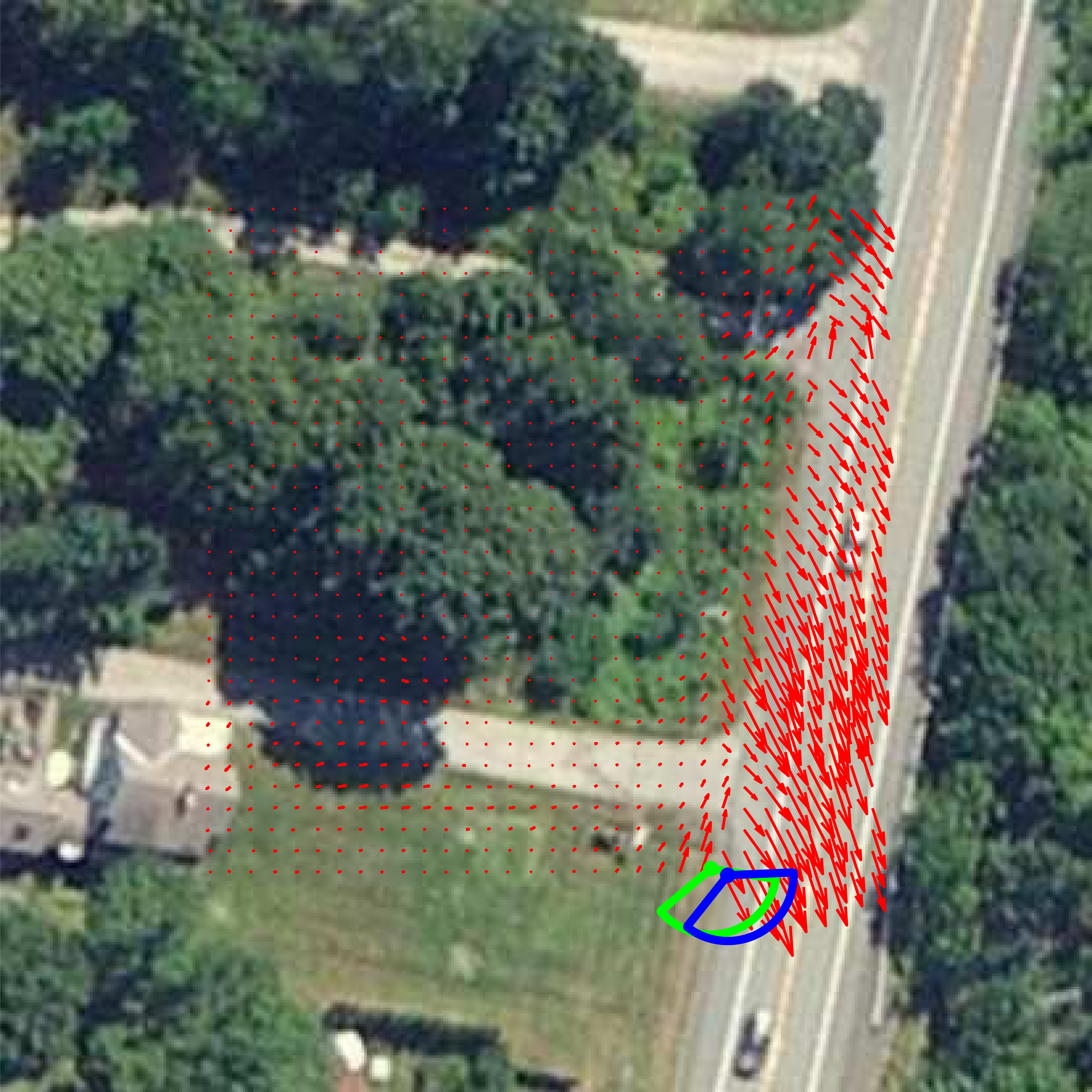} \hfill %
  \begin{minipage}[b]{\gwidth}
  \includegraphics[width=\textwidth]{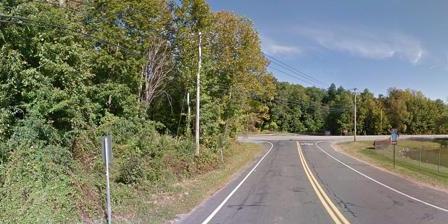}
  \includegraphics[width=\textwidth]{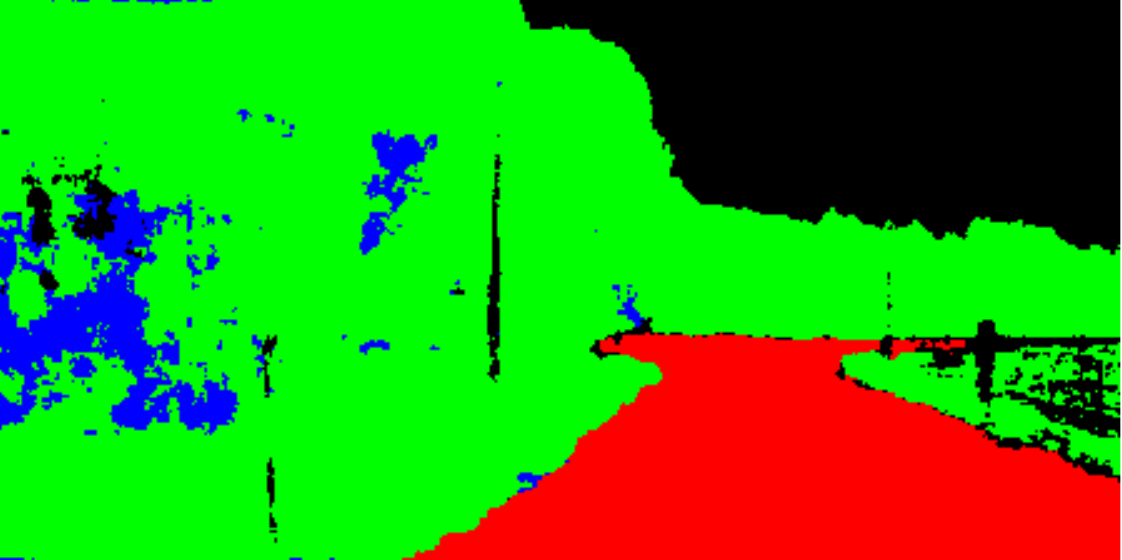}
  \includegraphics[width=\textwidth]{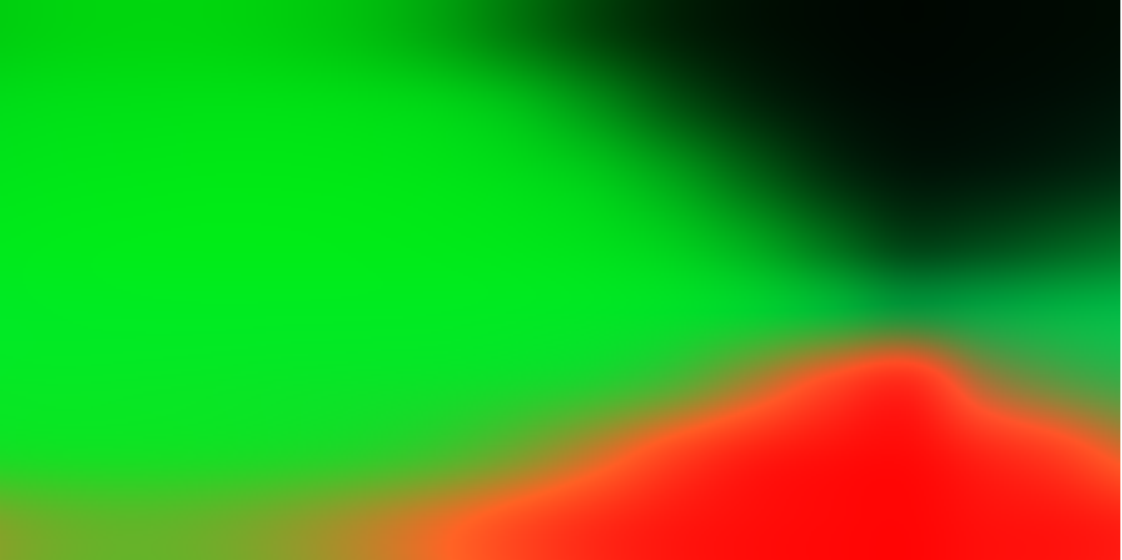}
  \end{minipage}
  \includegraphics[height=\aheight]{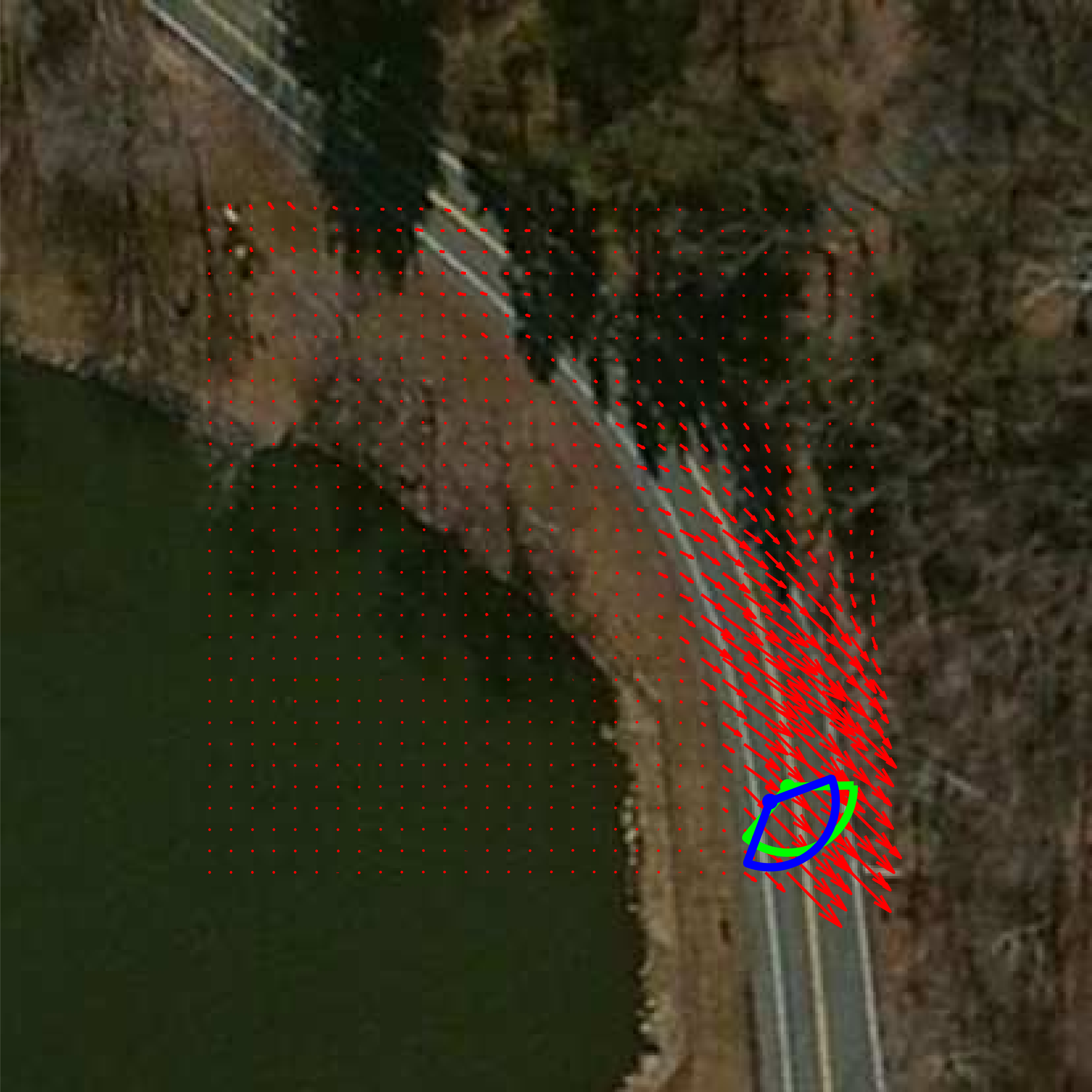} \hfill %
  \begin{minipage}[b]{\gwidth}
  \includegraphics[width=\textwidth]{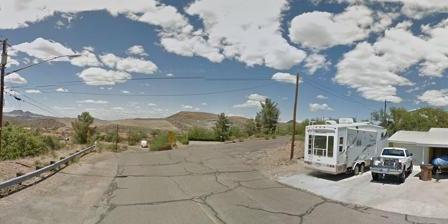}
  \includegraphics[width=\textwidth]{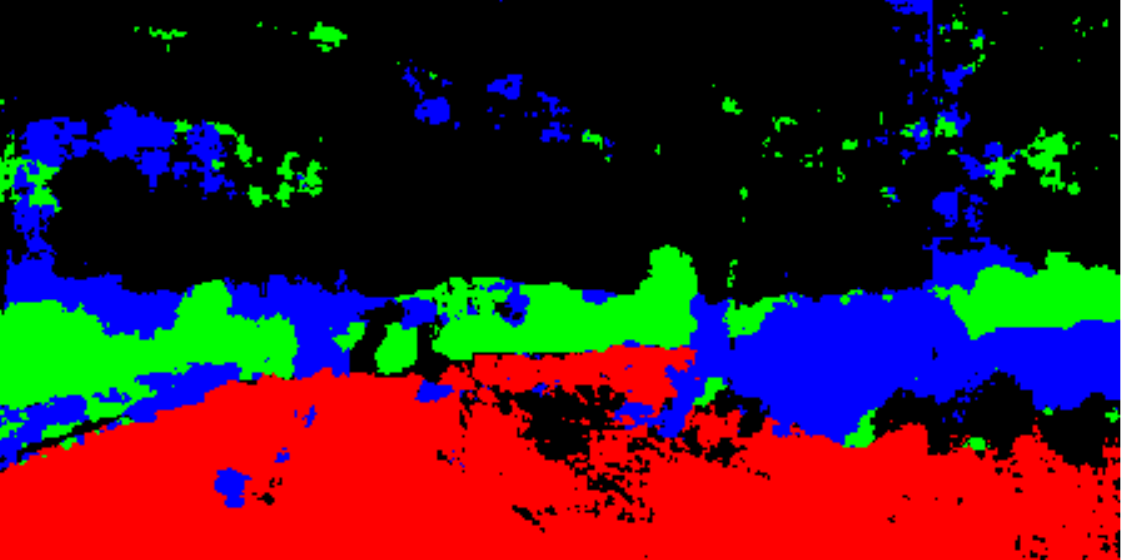}
  \includegraphics[width=\textwidth]{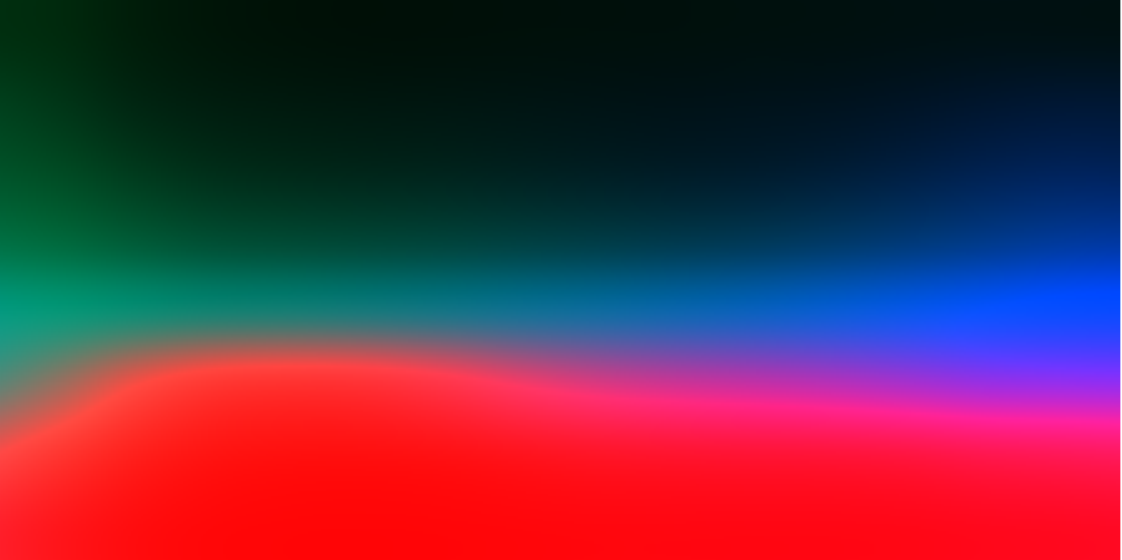}
  \end{minipage}
  \includegraphics[height=\aheight]{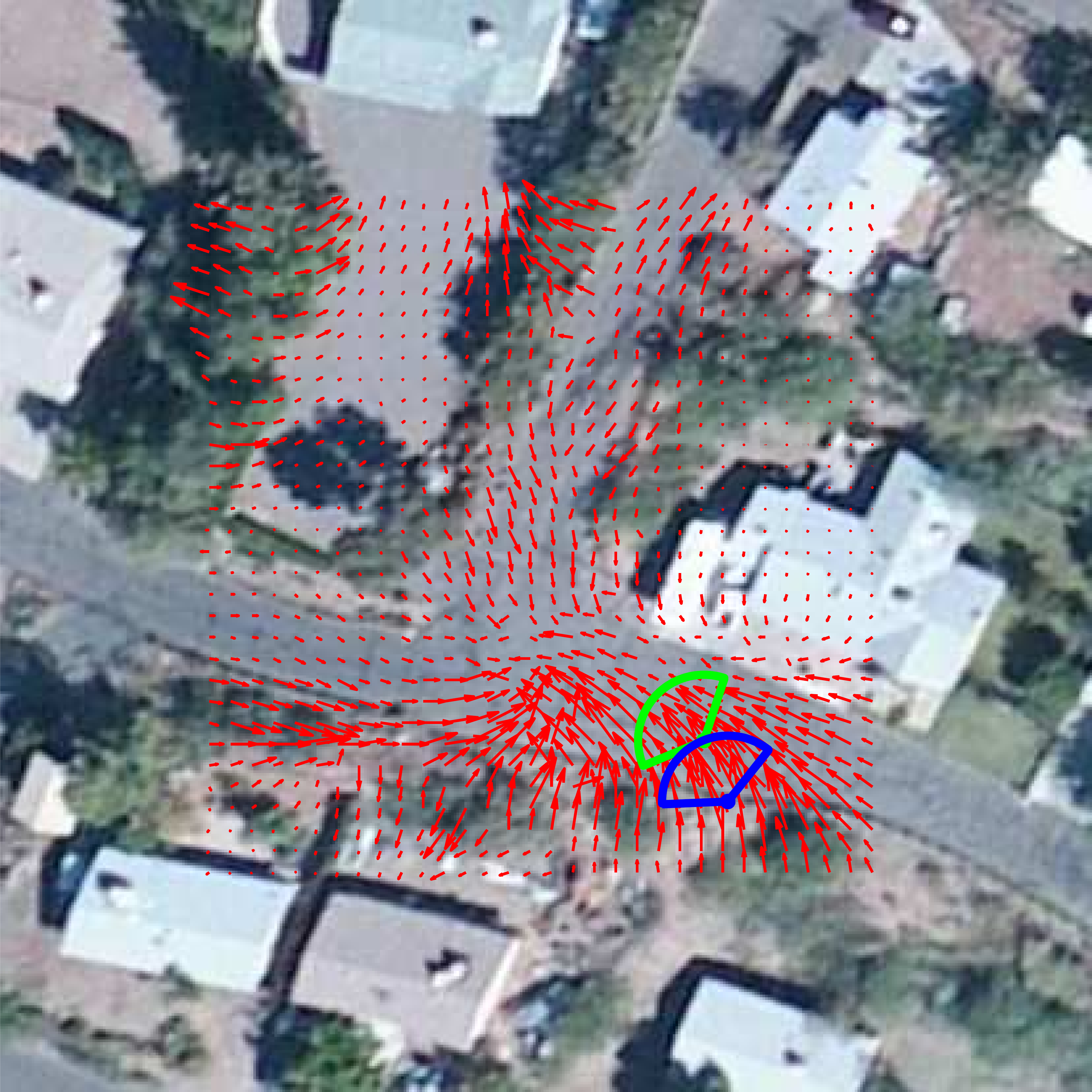} \hfill %
  \begin{minipage}[b]{\gwidth}
  \includegraphics[width=\textwidth]{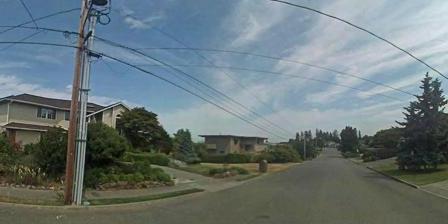}
  \includegraphics[width=\textwidth]{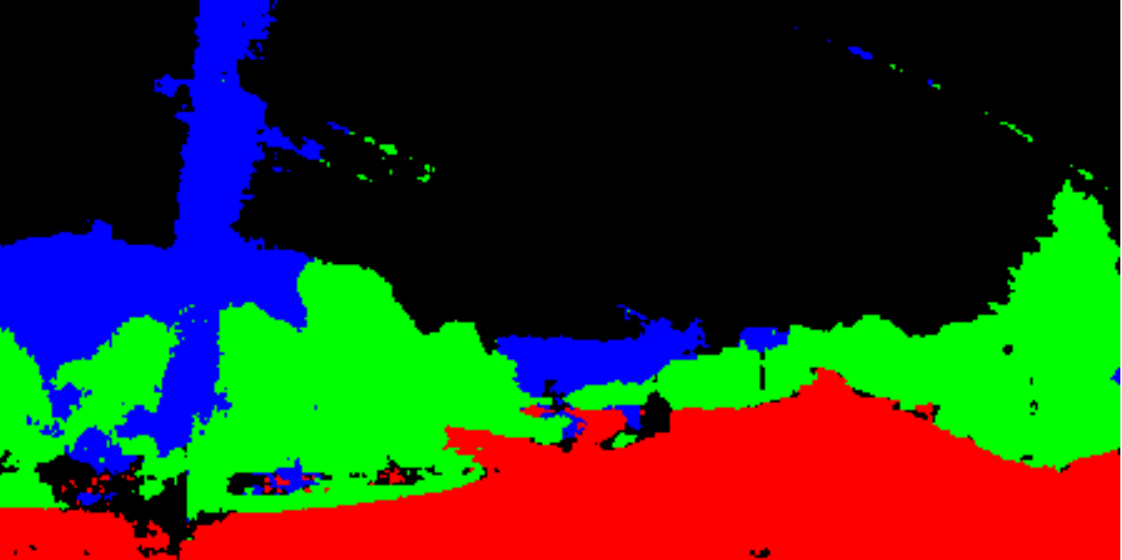}
  \includegraphics[width=\textwidth]{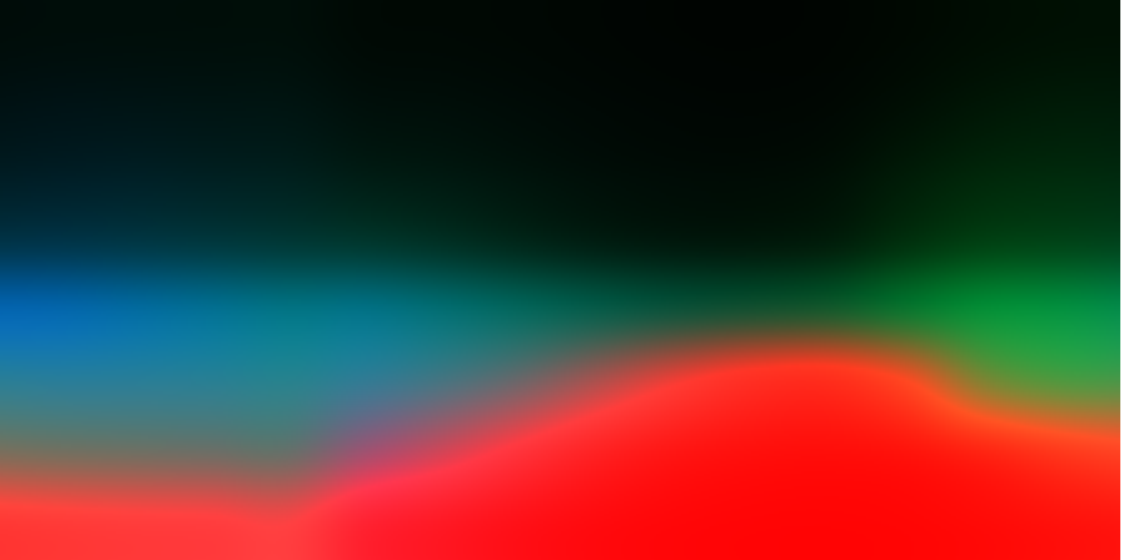}
  \end{minipage}
  \includegraphics[height=\aheight]{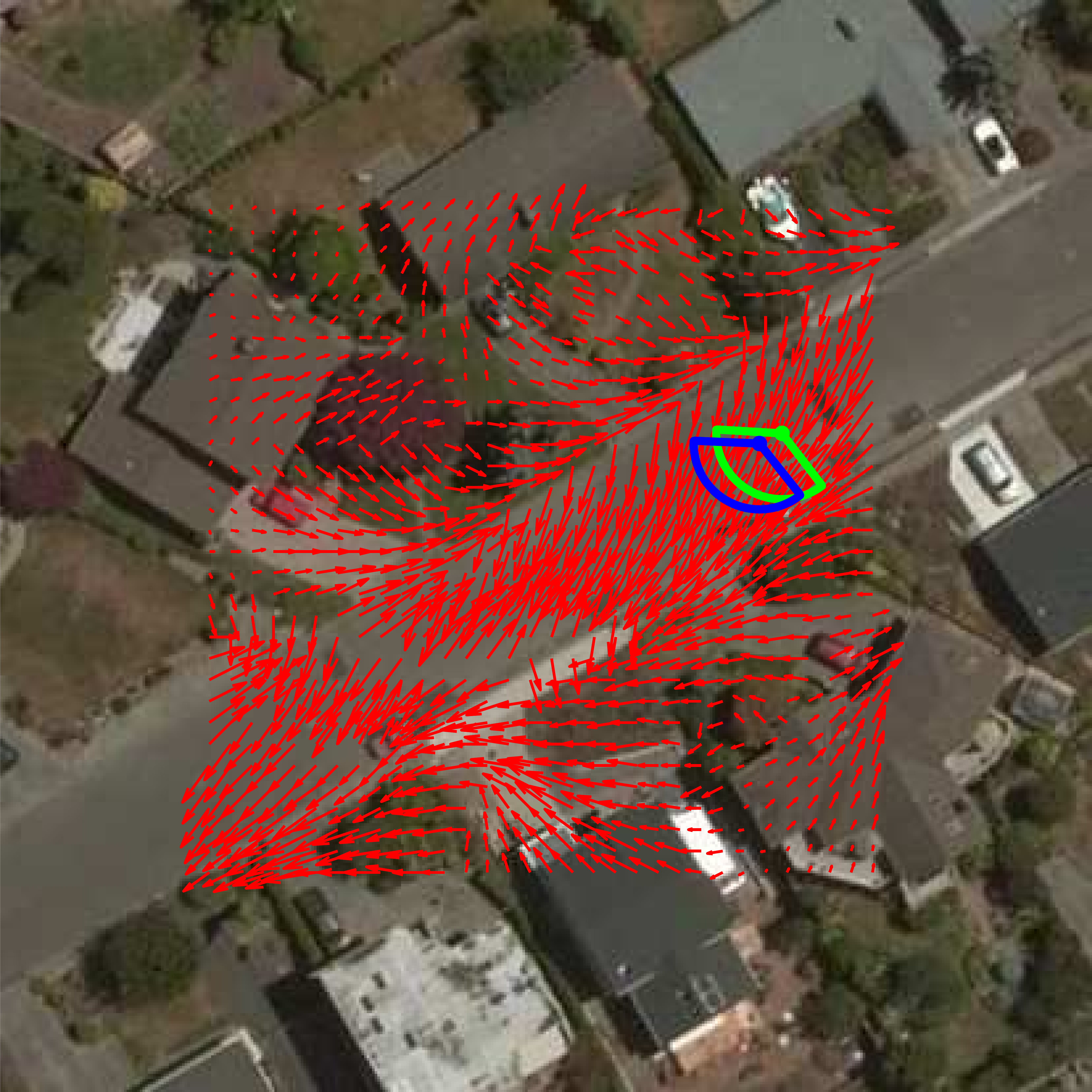} \hfill %
  \begin{minipage}[b]{\gwidth}
  \includegraphics[width=\textwidth]{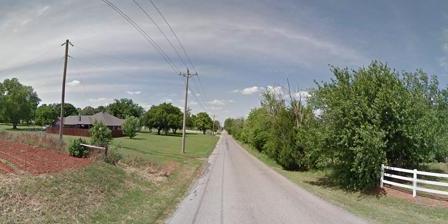}
  \includegraphics[width=\textwidth]{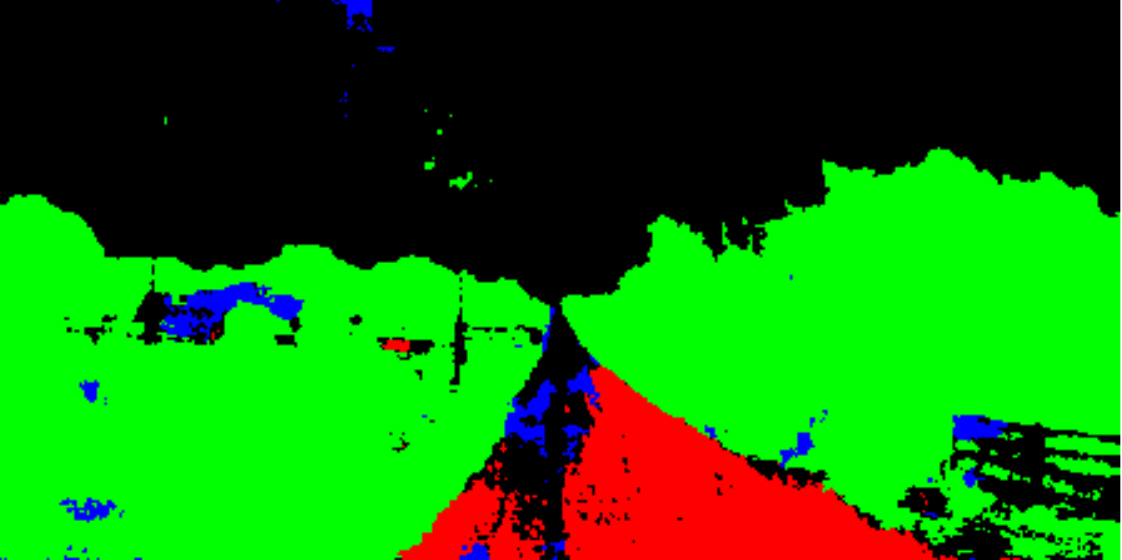}
  \includegraphics[width=\textwidth]{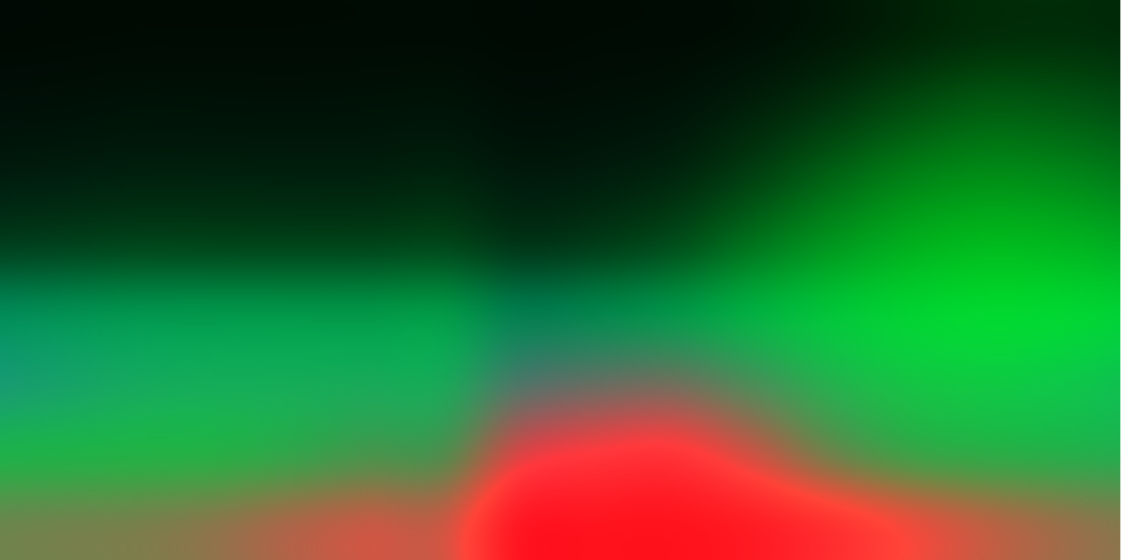}
  \end{minipage}
  \includegraphics[height=\aheight]{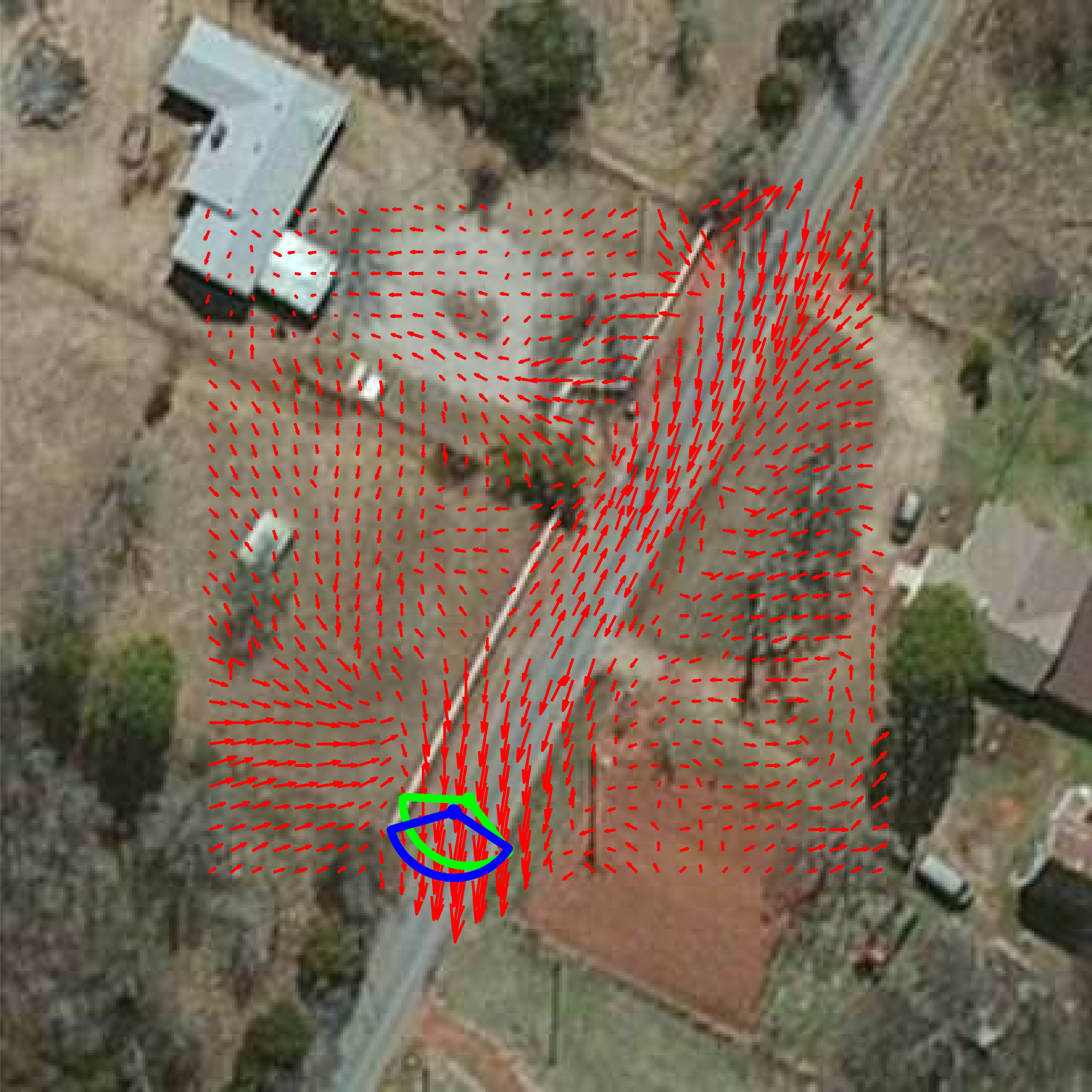} \hfill %
  \begin{minipage}[b]{\gwidth}
  \includegraphics[width=\textwidth]{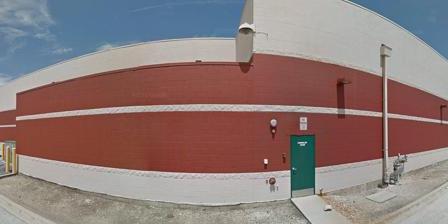}
  \includegraphics[width=\textwidth]{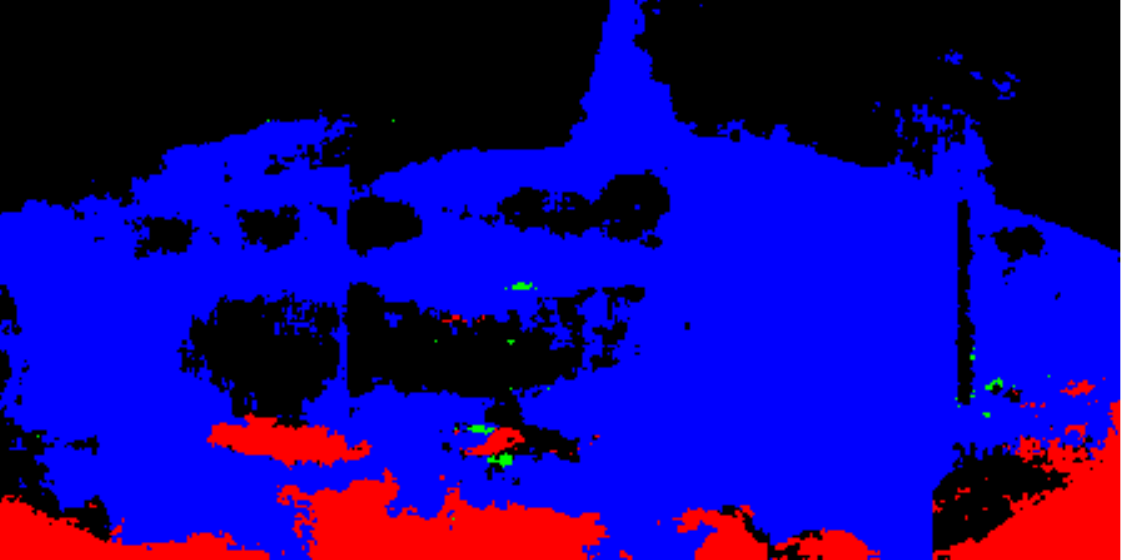}
  \includegraphics[width=\textwidth]{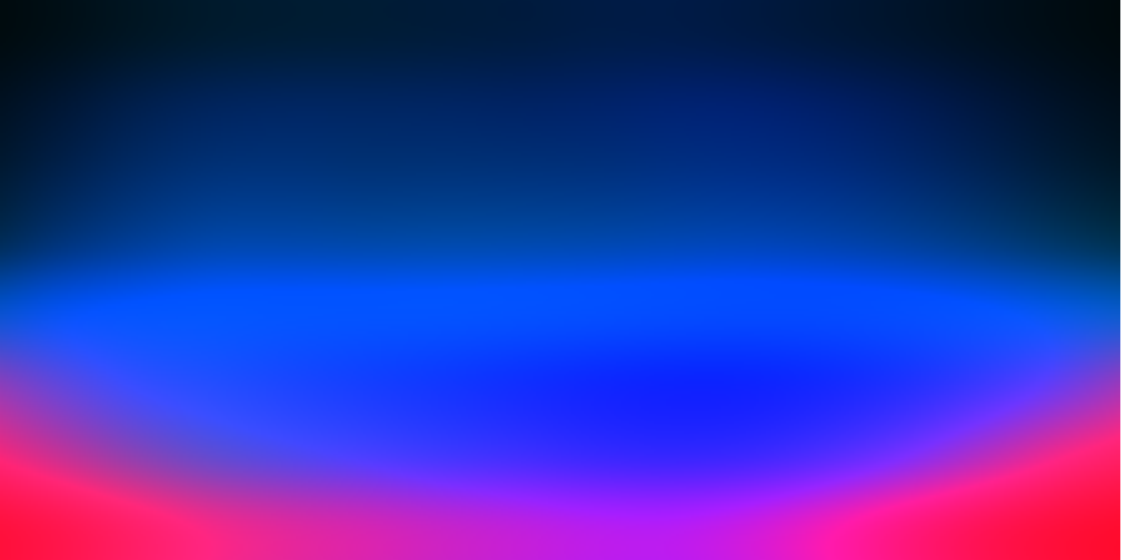}
  \end{minipage}
  \includegraphics[height=\aheight]{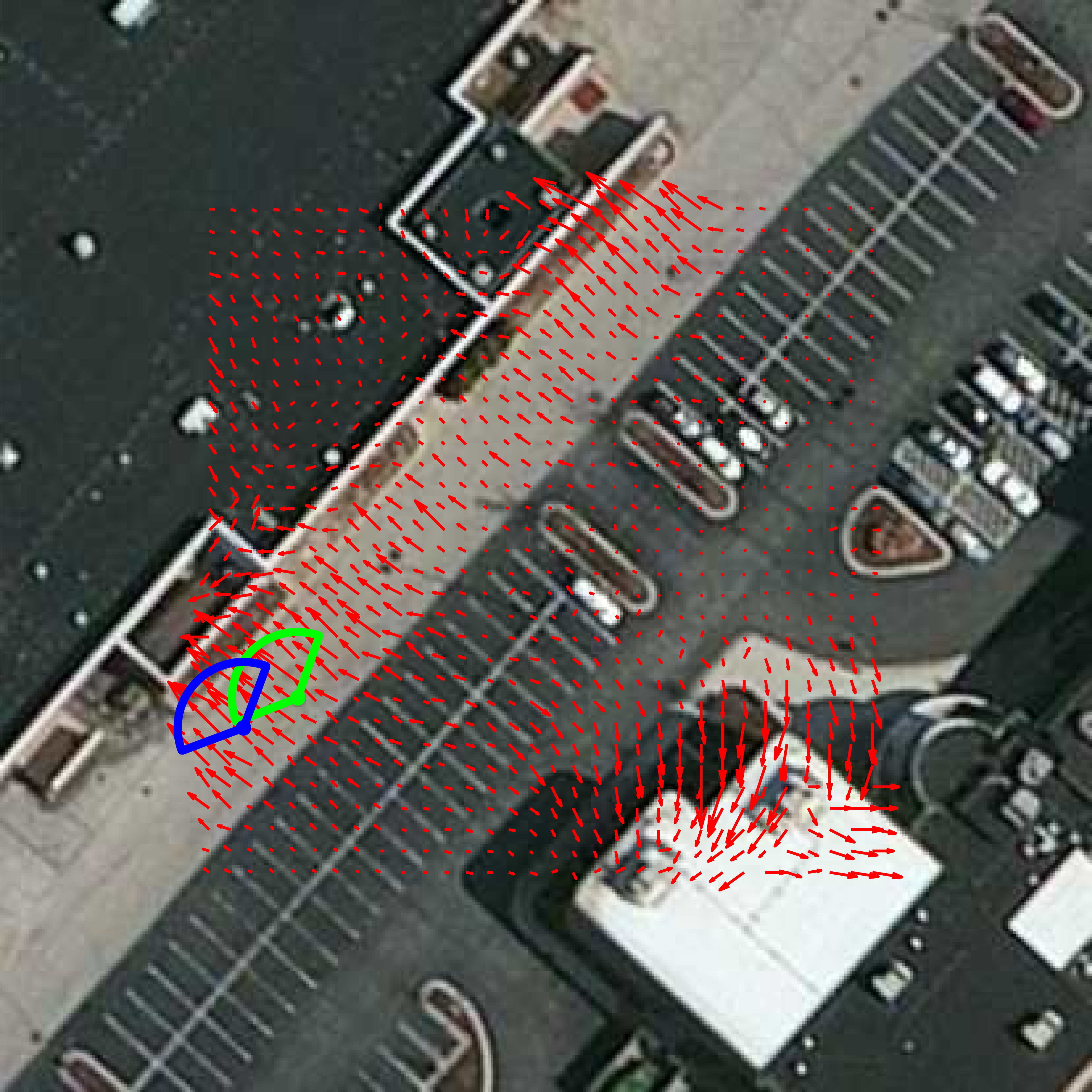} \hfill %
  \begin{minipage}[b]{\gwidth}
  \includegraphics[width=\textwidth]{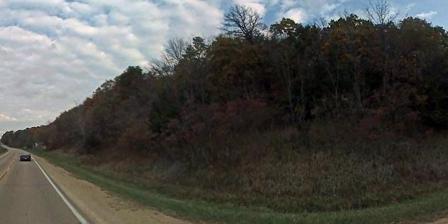}
  \includegraphics[width=\textwidth]{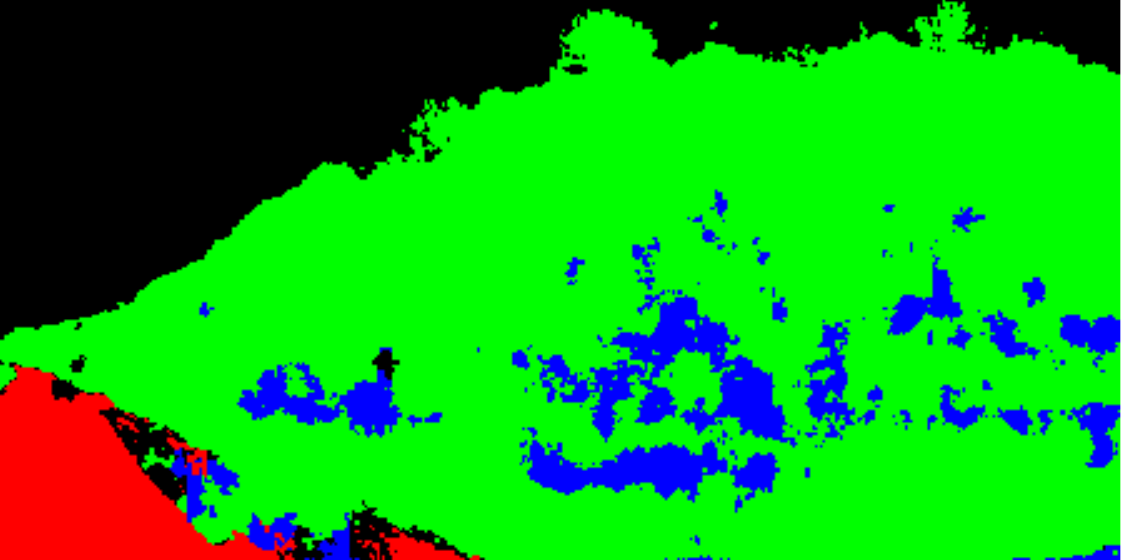}
  \includegraphics[width=\textwidth]{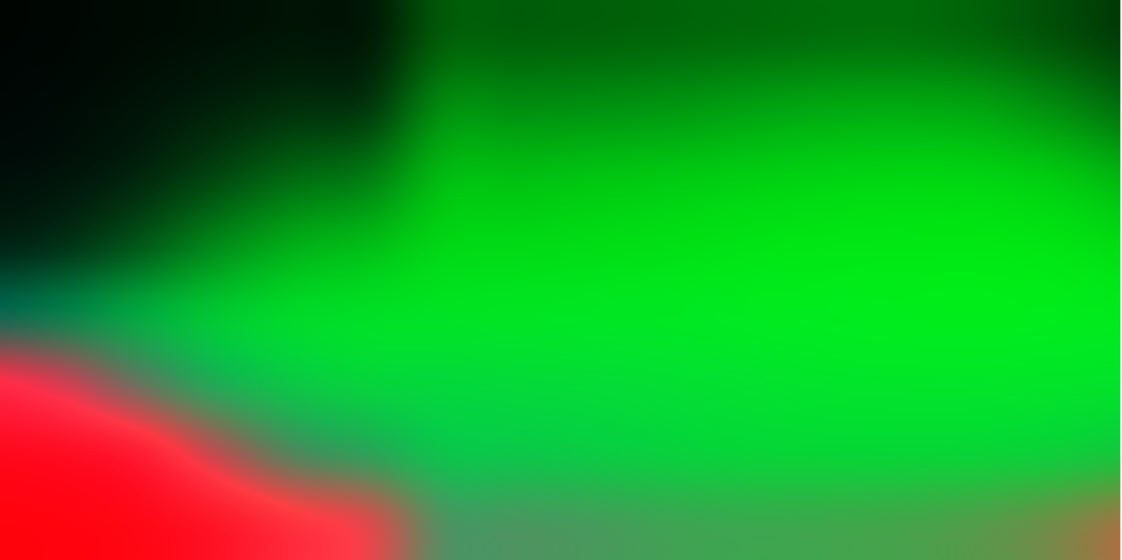}
  \end{minipage}
  \includegraphics[height=\aheight]{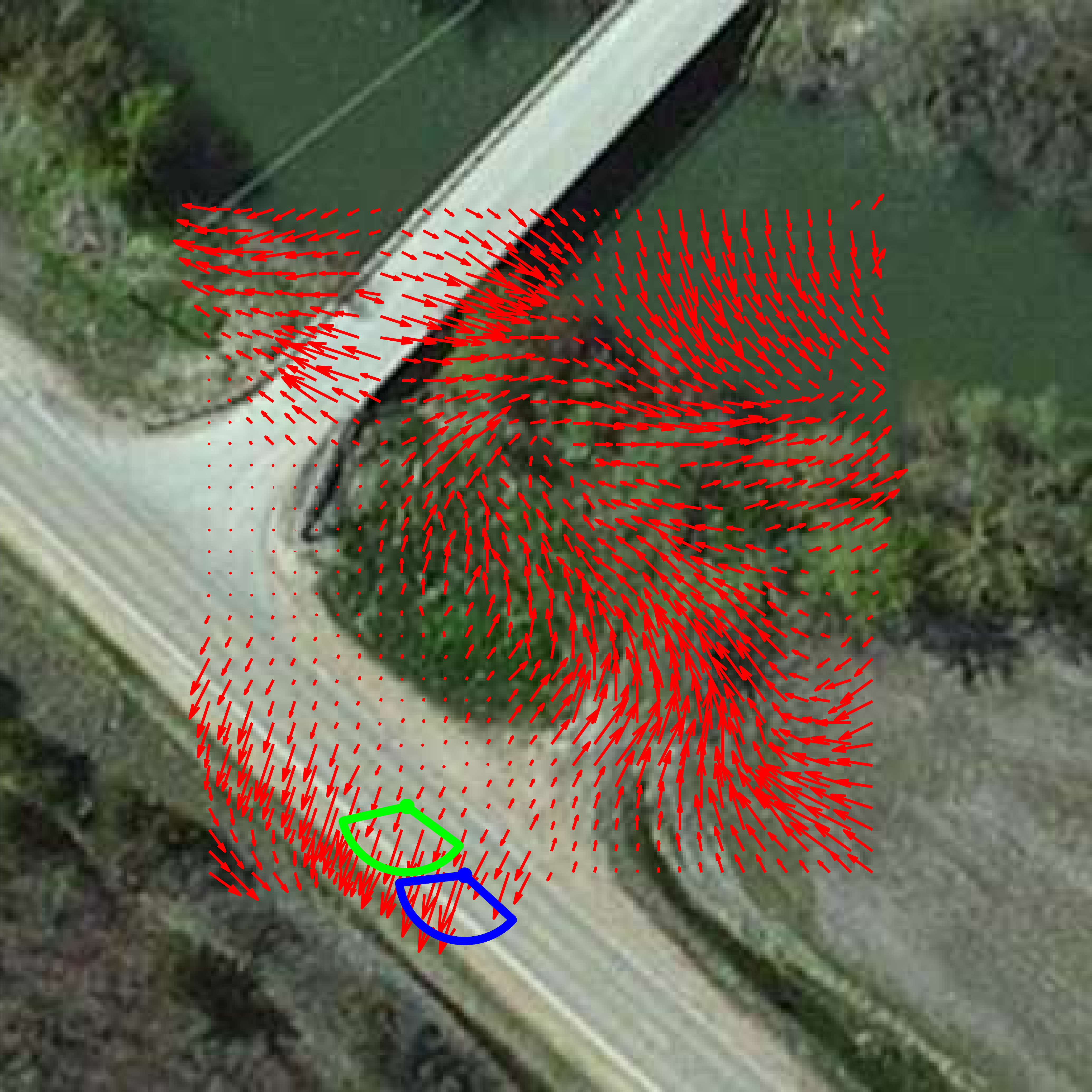} \hfill %
  \begin{minipage}[b]{\gwidth}
  \includegraphics[width=\textwidth]{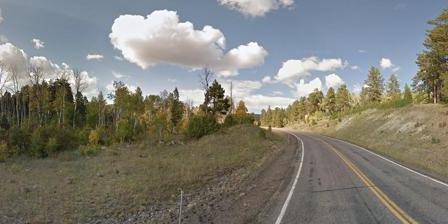}
  \includegraphics[width=\textwidth]{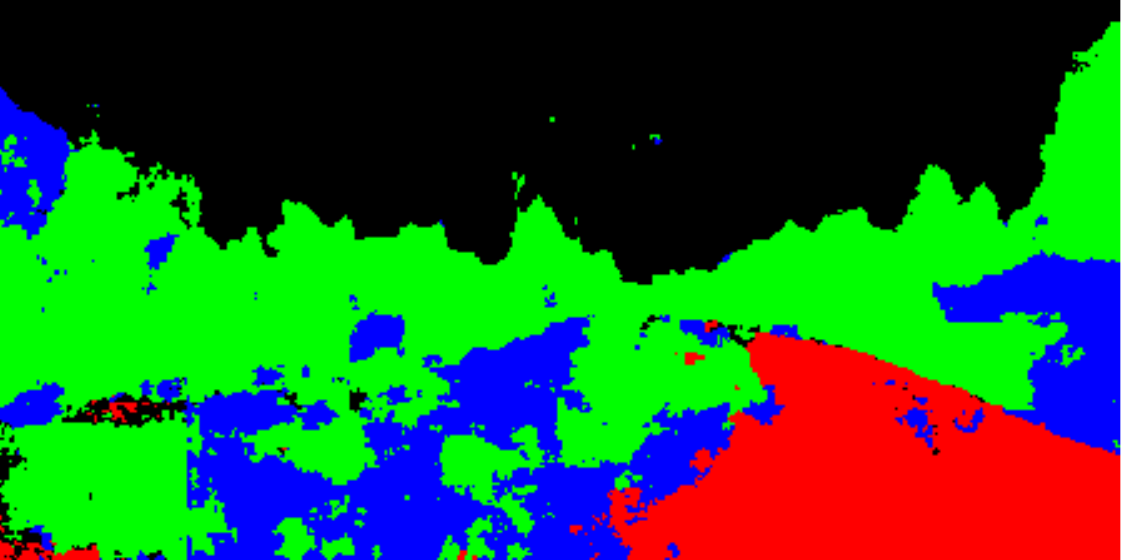}
  \includegraphics[width=\textwidth]{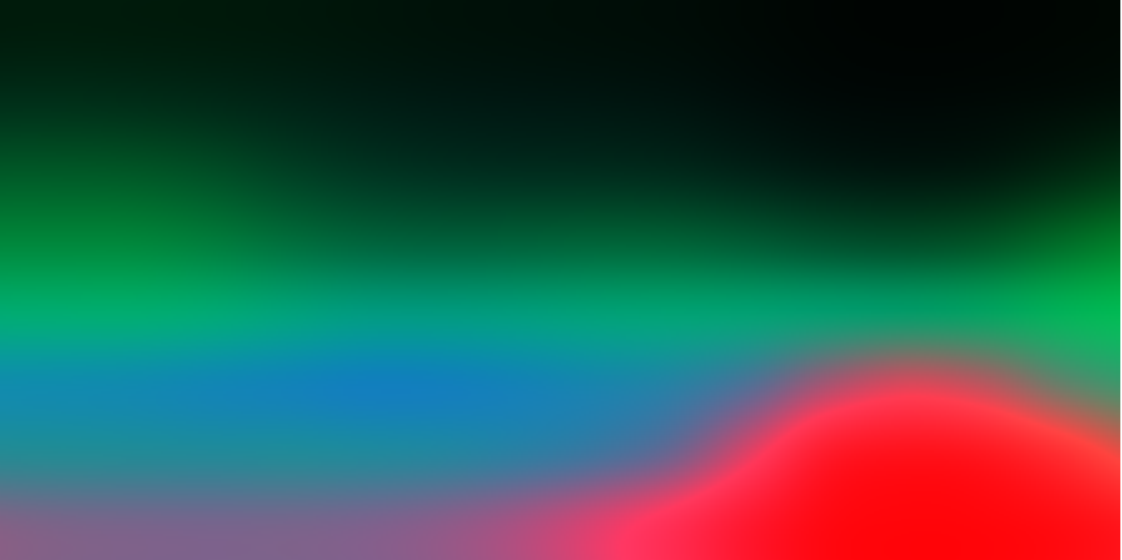}
  \end{minipage}
  \includegraphics[height=\aheight]{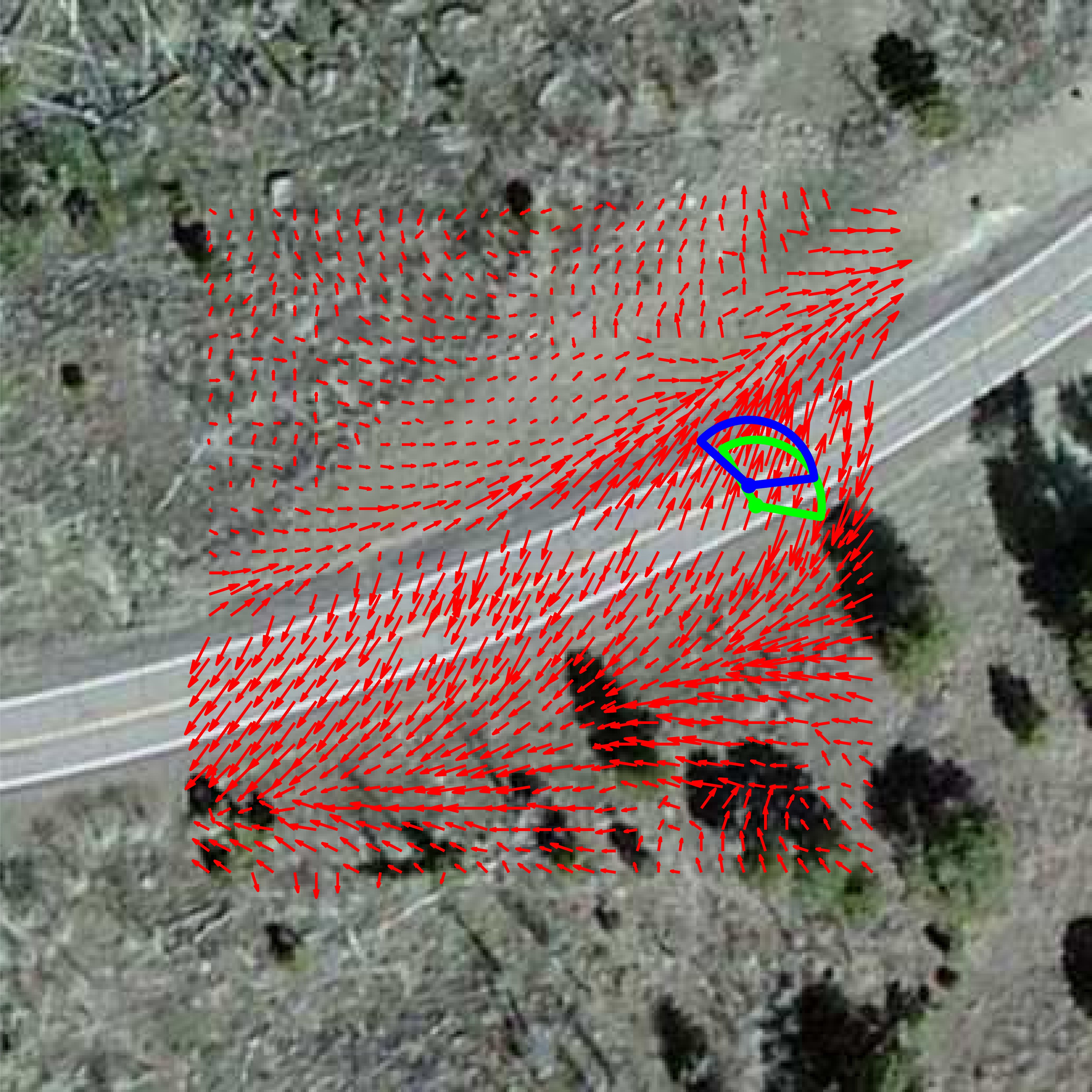} \hfill %
  \begin{minipage}[b]{\gwidth}
  \includegraphics[width=\textwidth]{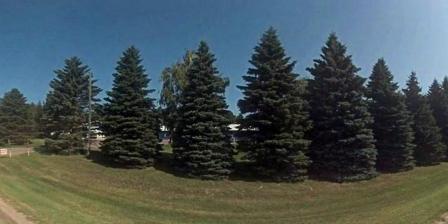}
  \includegraphics[width=\textwidth]{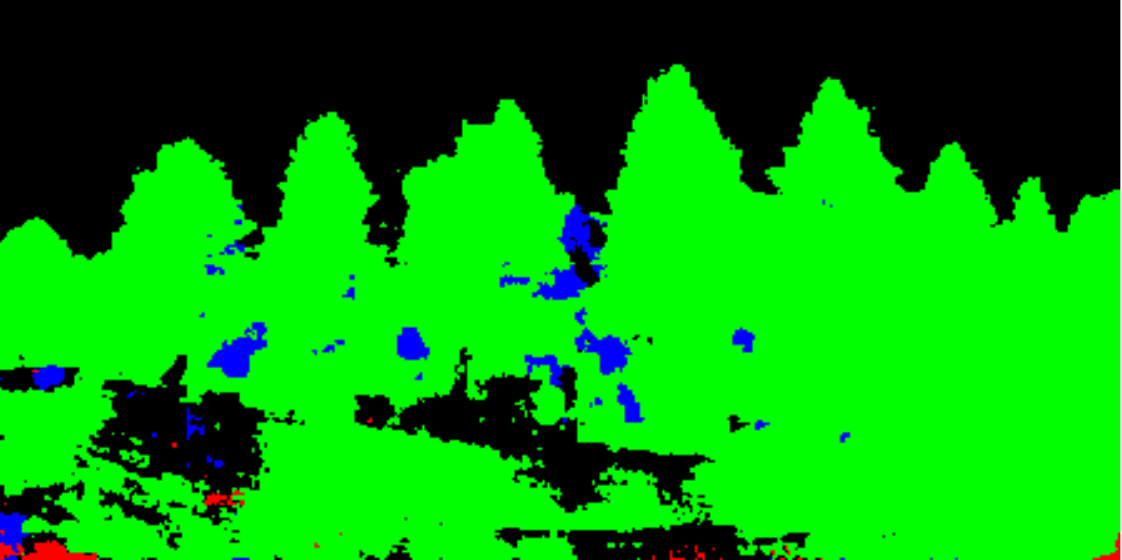}
  \includegraphics[width=\textwidth]{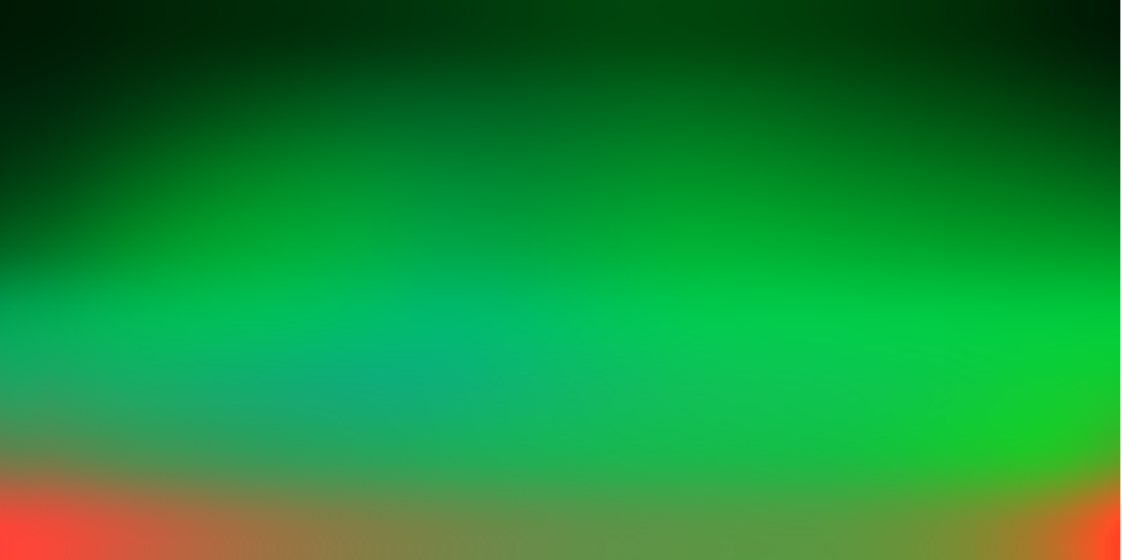}
  \end{minipage}
  \includegraphics[height=\aheight]{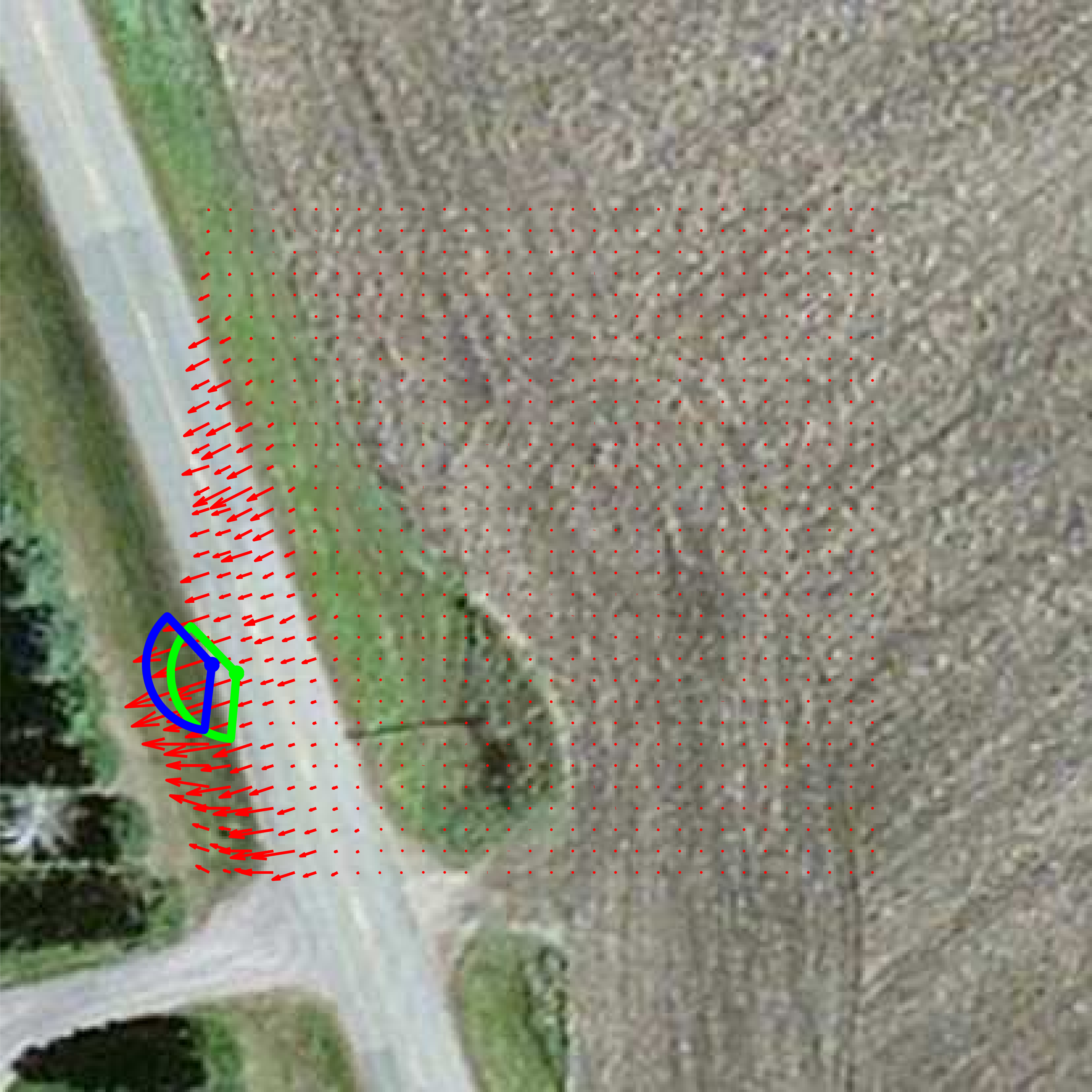} \hfill %
  \begin{minipage}[b]{\gwidth}
  \includegraphics[width=\textwidth]{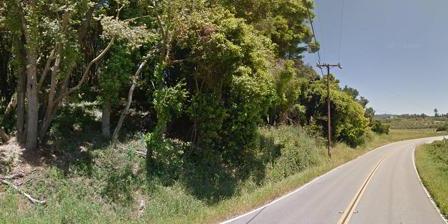}
  \includegraphics[width=\textwidth]{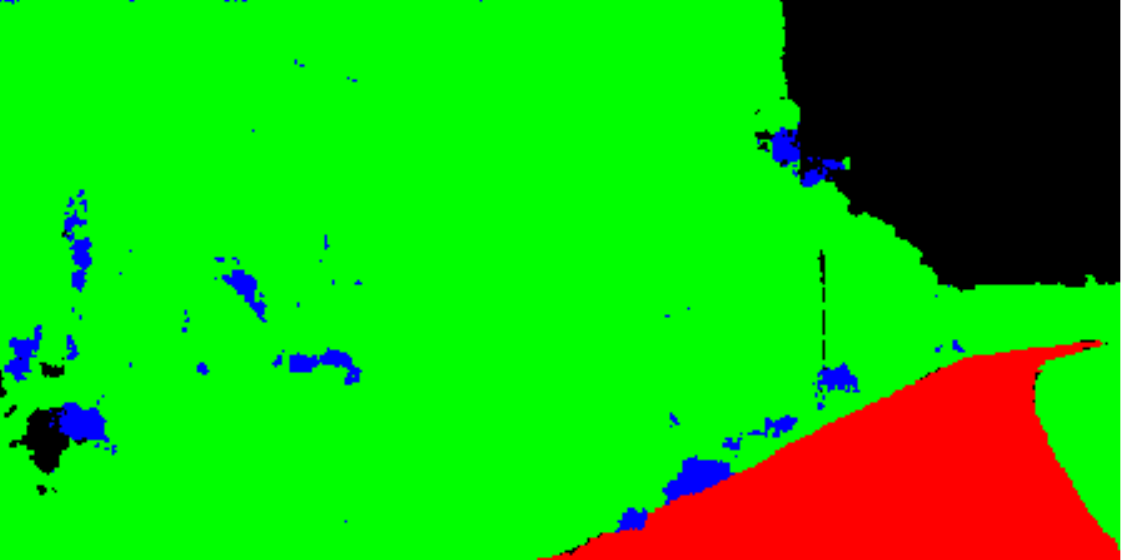}
  \includegraphics[width=\textwidth]{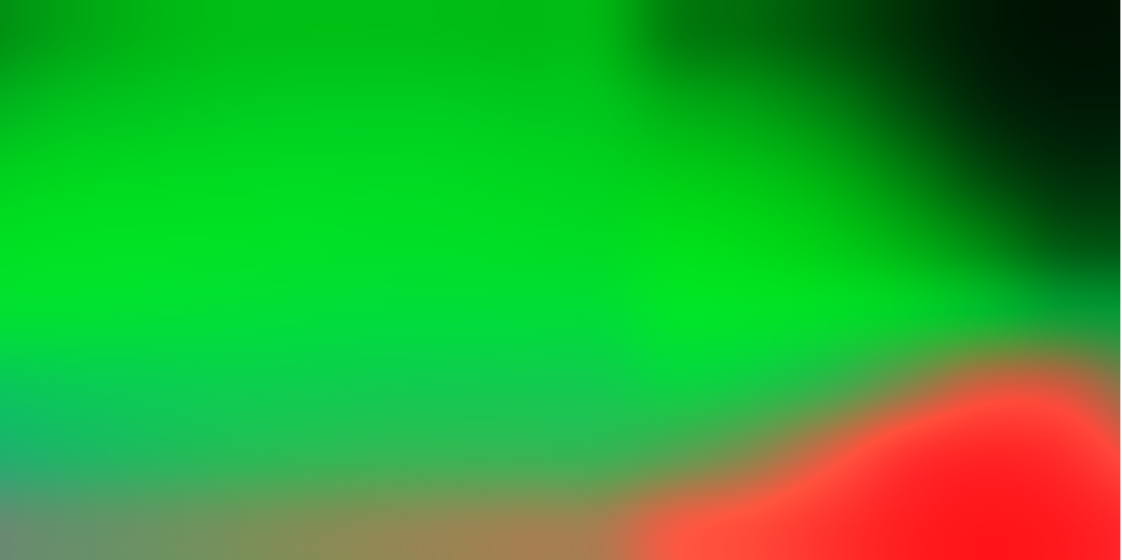}
  \end{minipage}
  \includegraphics[height=\aheight]{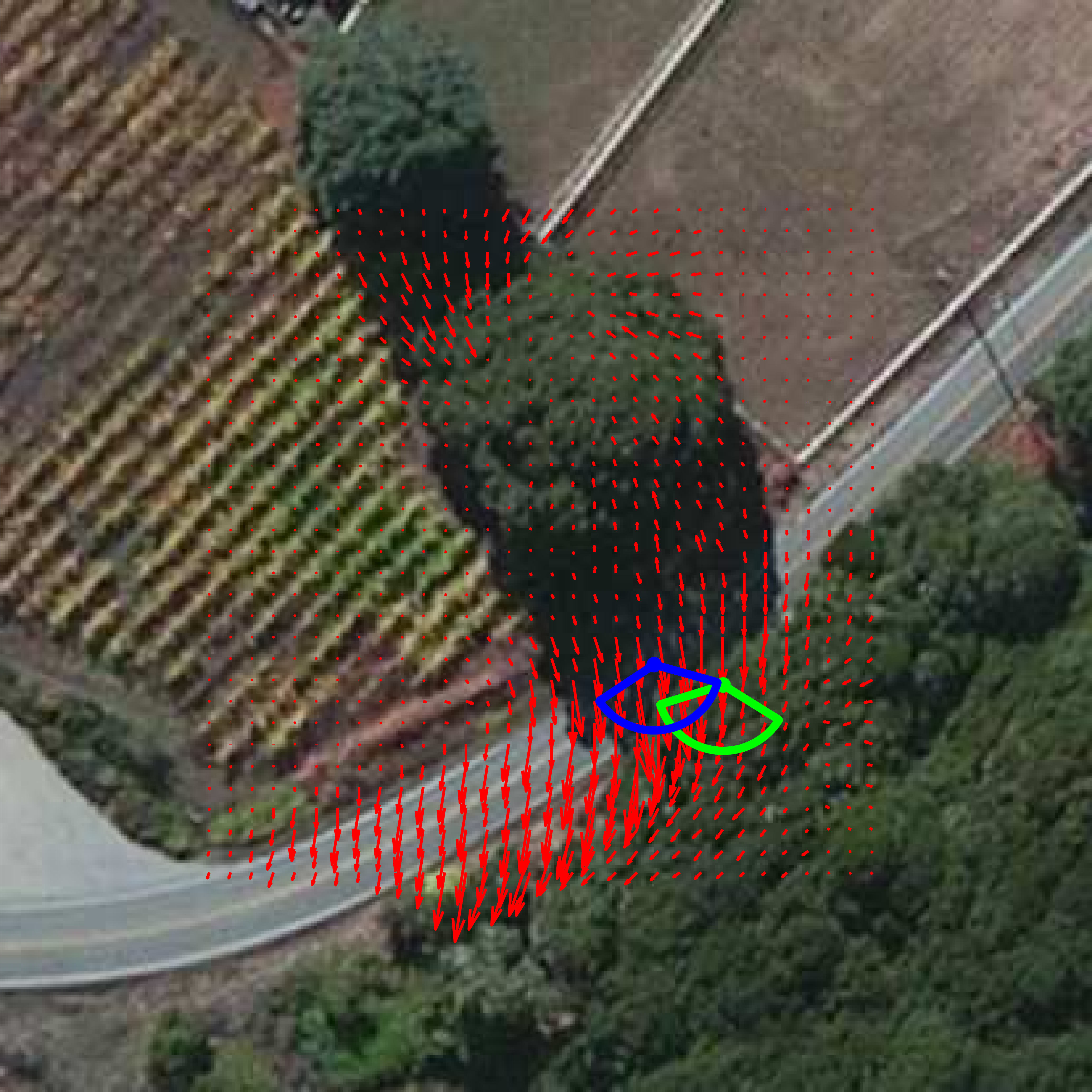} \hfill %
  \begin{minipage}[b]{\gwidth}
  \includegraphics[width=\textwidth]{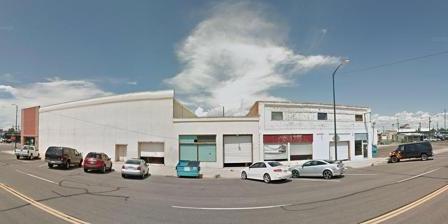}
  \includegraphics[width=\textwidth]{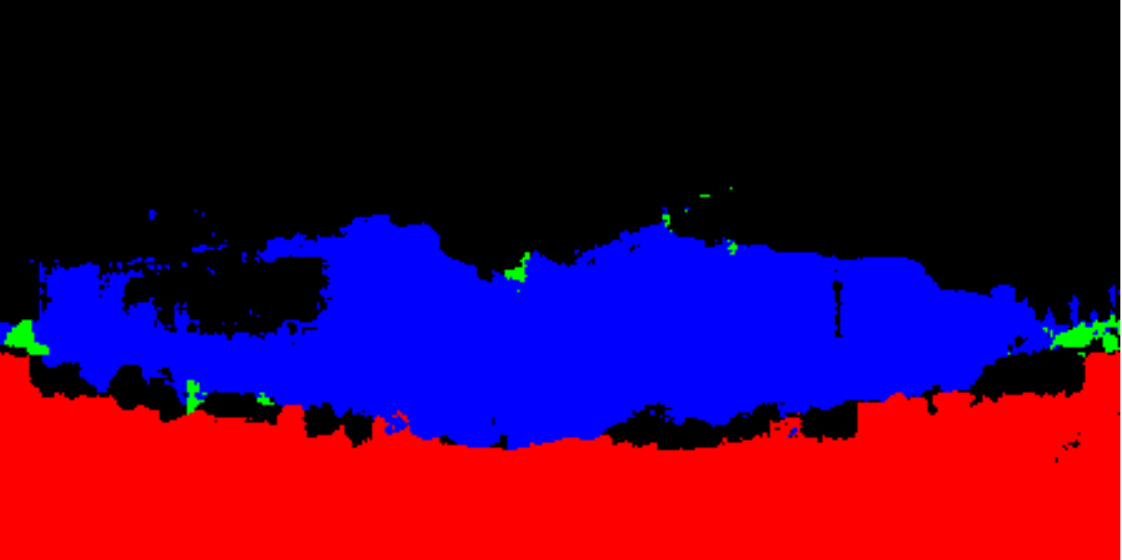}
  \includegraphics[width=\textwidth]{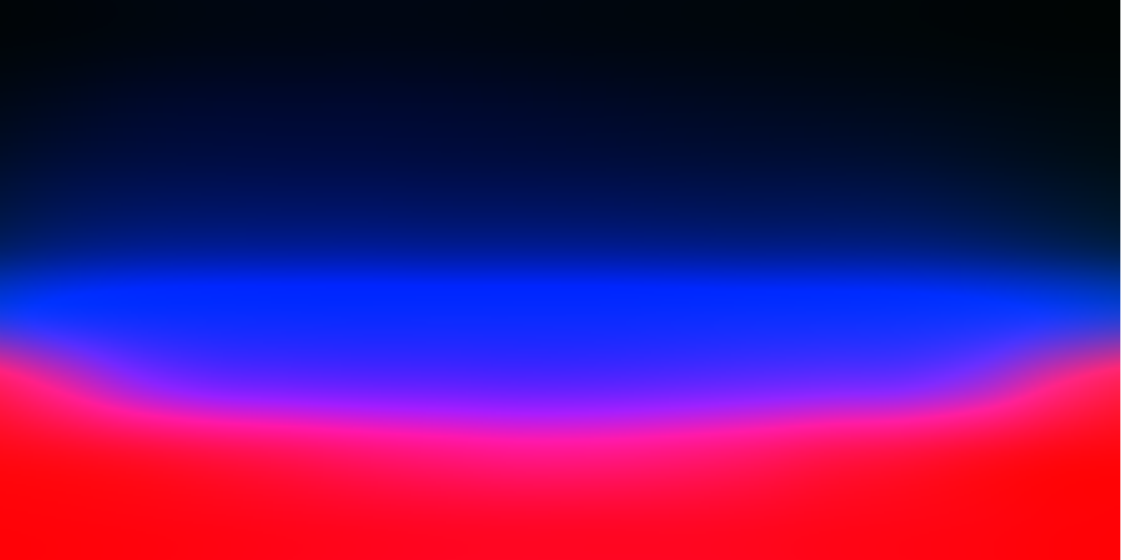}
  \end{minipage}
  \includegraphics[height=\aheight]{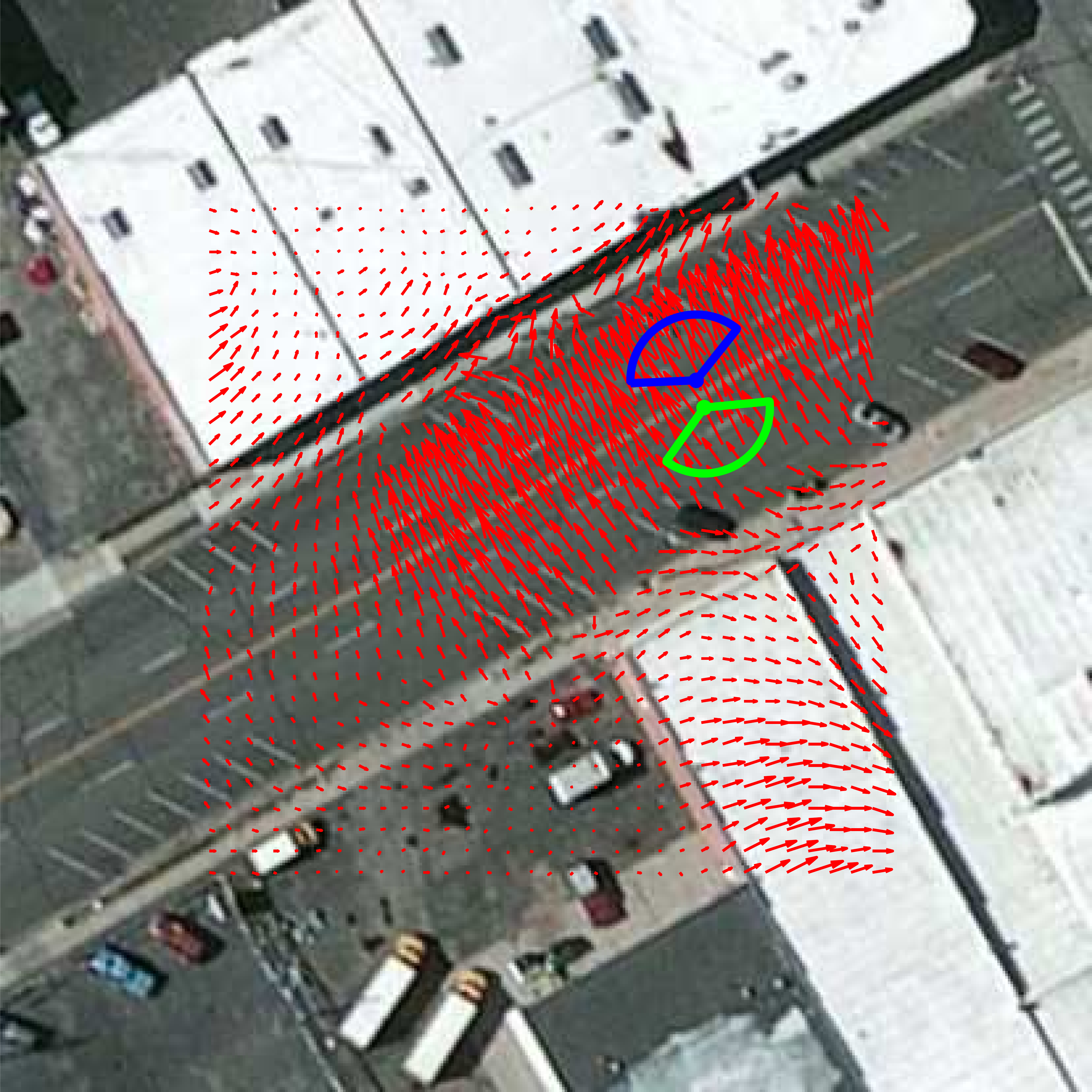} \hfill %
  \begin{minipage}[b]{\gwidth}
  \includegraphics[width=\textwidth]{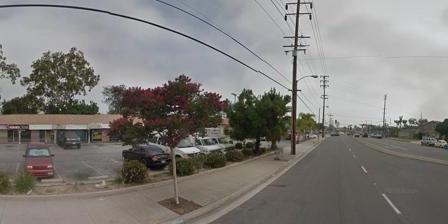}
  \includegraphics[width=\textwidth]{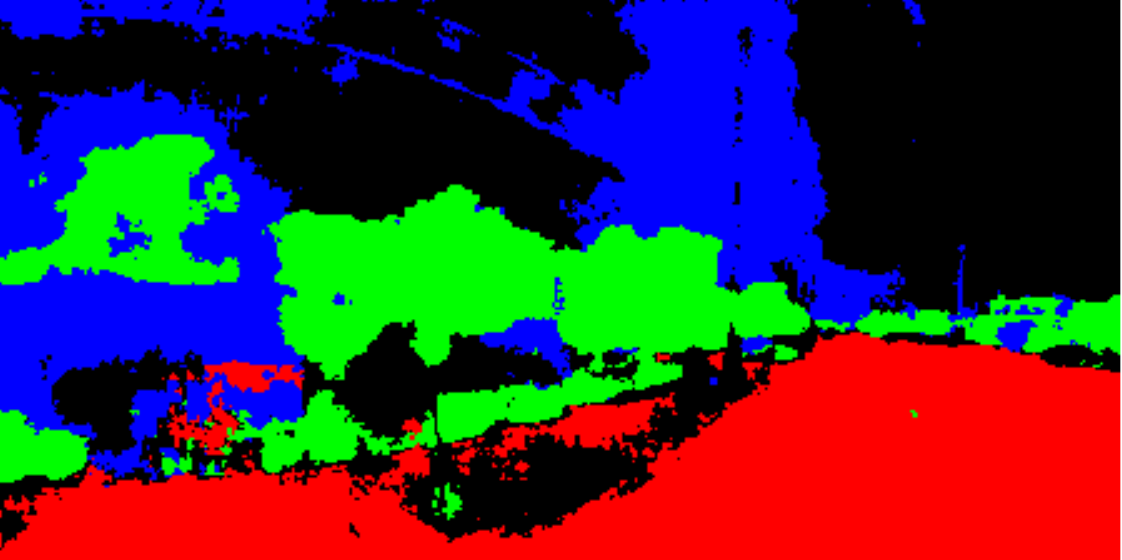}
  \includegraphics[width=\textwidth]{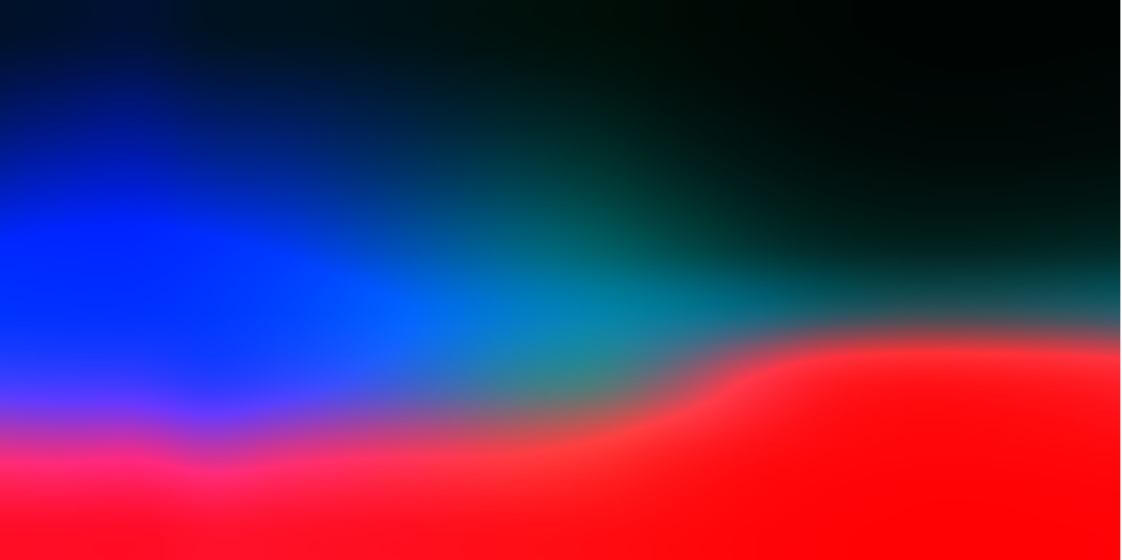}
  \end{minipage}
  \includegraphics[height=\aheight]{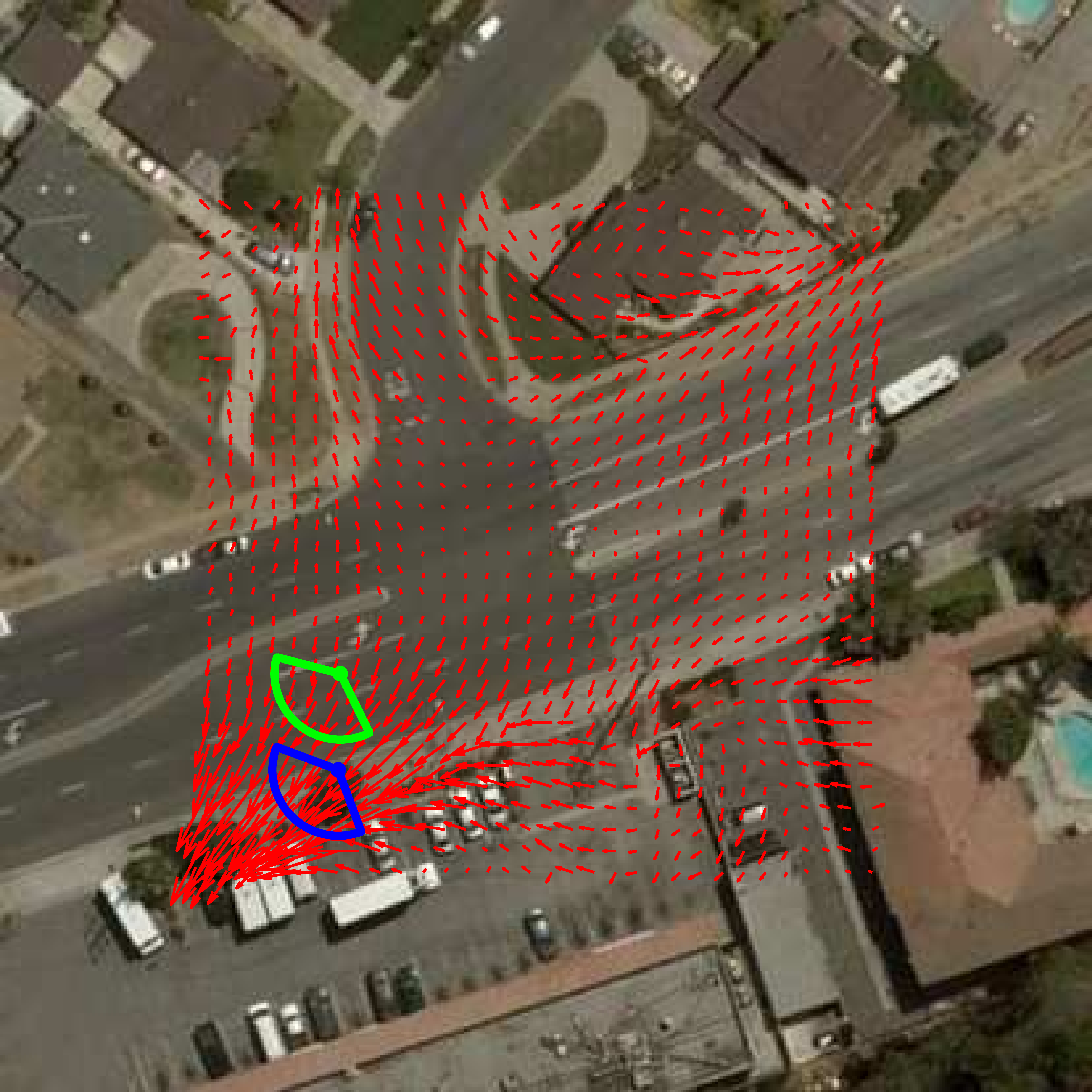} \hfill %
  \begin{minipage}[b]{\gwidth}
  \includegraphics[width=\textwidth]{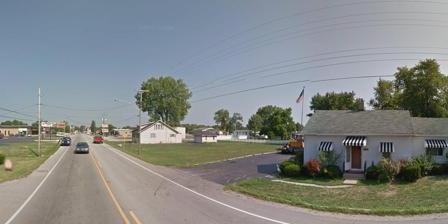}
  \includegraphics[width=\textwidth]{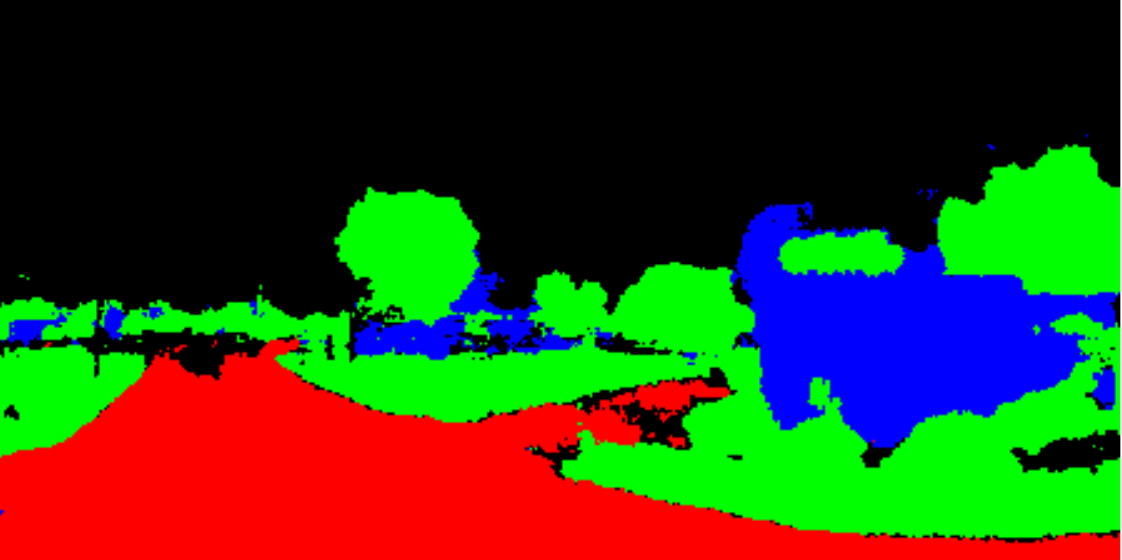}
  \includegraphics[width=\textwidth]{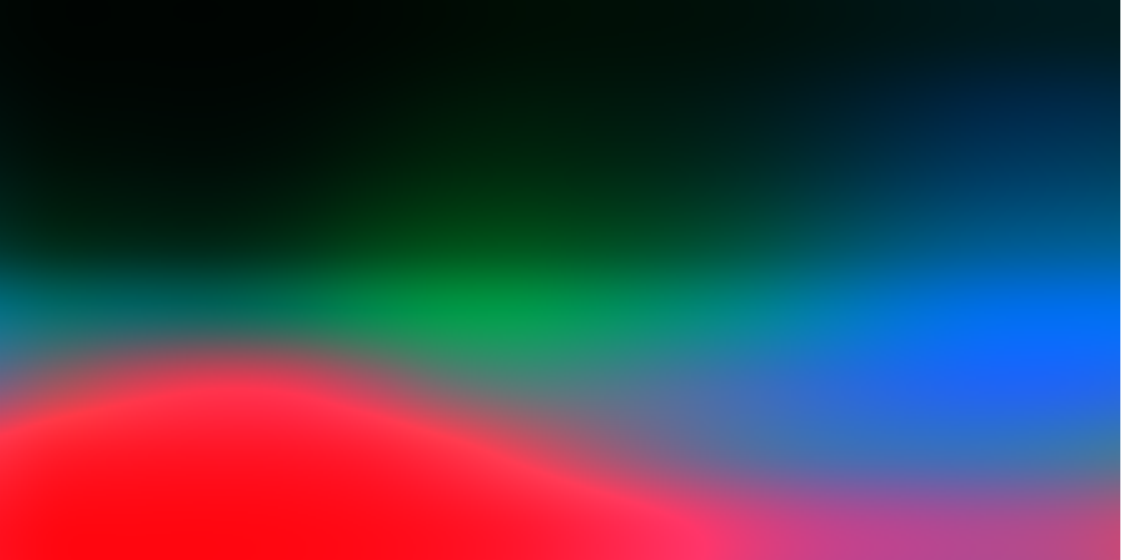}
  \end{minipage}
  \includegraphics[height=\aheight]{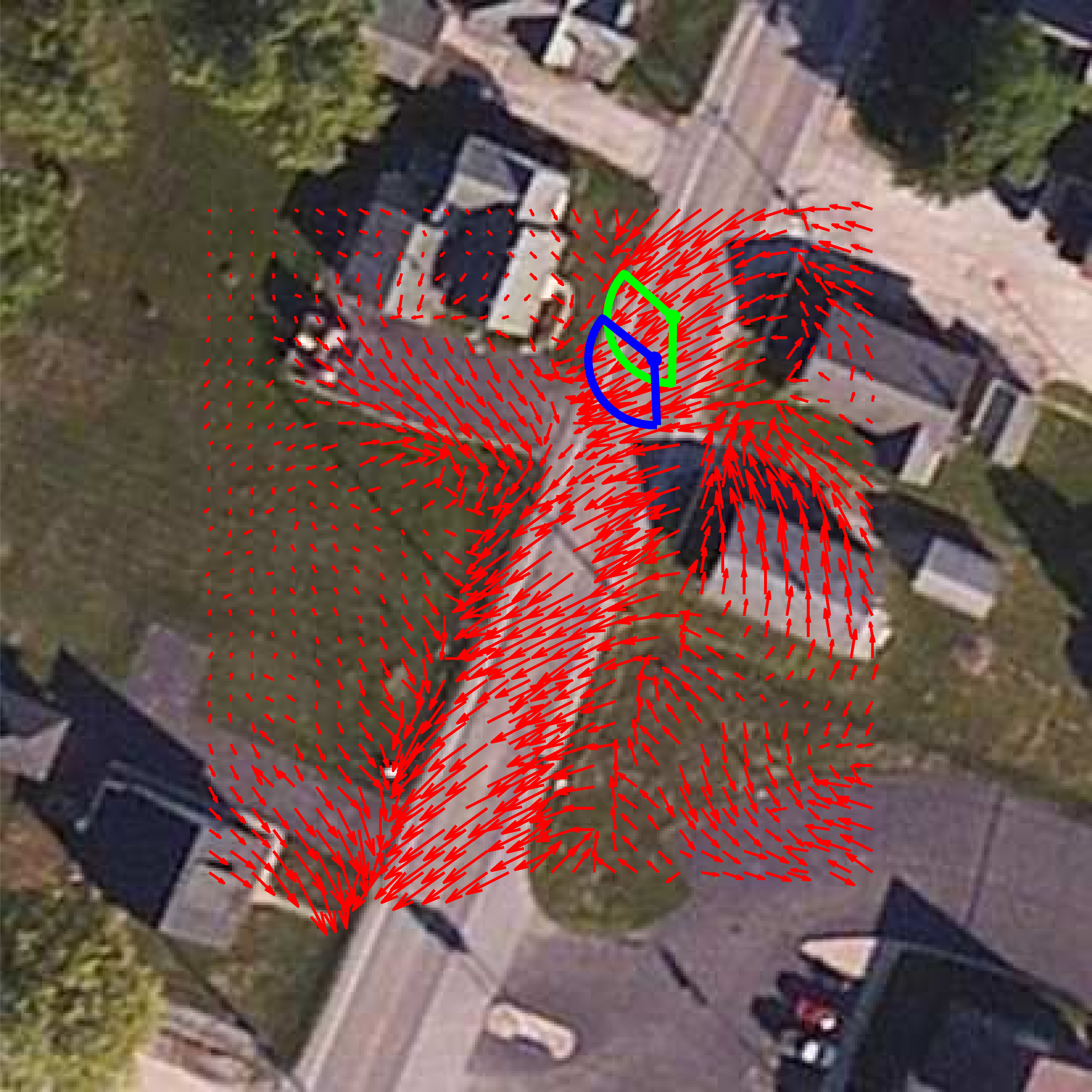}
  
  \caption{Additional fine-grained geocalibration results.  (left)
  Visualized from top to bottom are $I_g$, $L_g$, and $L_{g'}$
  respectively.  We visualize three classes: {\em road} (red), {\em
  vegetation} (green), and {\em man-made} (blue).  (right) Orientation
  flow map (red), where the arrow direction indicates the optimal
  direction at that location and length indicates the magnitude. We
  also show the optimal prediction and the ground-truth frustums in
  blue and green respectively.}
  \label{fig:app:geocali}
\end{figure*}

\begin{figure*}[!t] %
	\setlength{\awidth}{.09125\linewidth}
	\setlength{\gwidth}{.23\linewidth}
	\setlength{\gspace}{1pt}
	\includegraphics[width=\awidth]{gan/aerial1}\hspace*{\gspace} %
	\begin{minipage}[b]{\gwidth}
	\includegraphics[width=\textwidth]{gan/ground1}\vfill
	\vspace*{\gspace}
	\includegraphics[width=\textwidth]{gan/ground1_pred}
	\end{minipage}\vspace*{\gspace}\hfill %
	\includegraphics[width=\awidth]{gan/aerial2}\hspace*{\gspace} %
	\begin{minipage}[b]{\gwidth}
	\includegraphics[width=\textwidth]{gan/ground2}\vfill
	\vspace*{\gspace}
	\includegraphics[width=\textwidth]{gan/ground2_pred}
	\end{minipage}\vspace*{\gspace}\hfill %
	\includegraphics[width=\awidth]{gan/aerial3}\hspace*{\gspace} %
	\begin{minipage}[b]{\gwidth}
	\includegraphics[width=\textwidth]{gan/ground3}\vfill
	\vspace*{\gspace}
	\includegraphics[width=\textwidth]{gan/ground3_pred}
	\end{minipage}\vspace*{\gspace}\hfill %
	
	\includegraphics[width=\awidth]{gan/aerial4}\hspace*{\gspace} %
	\begin{minipage}[b]{\gwidth}
	\includegraphics[width=\textwidth]{gan/ground4}\vfill
	\vspace*{\gspace}
	\includegraphics[width=\textwidth]{gan/ground4_pred}
	\end{minipage}\vspace*{\gspace}\hfill %
	\includegraphics[width=\awidth]{gan/aerial5}\hspace*{\gspace} %
	\begin{minipage}[b]{\gwidth}
	\includegraphics[width=\textwidth]{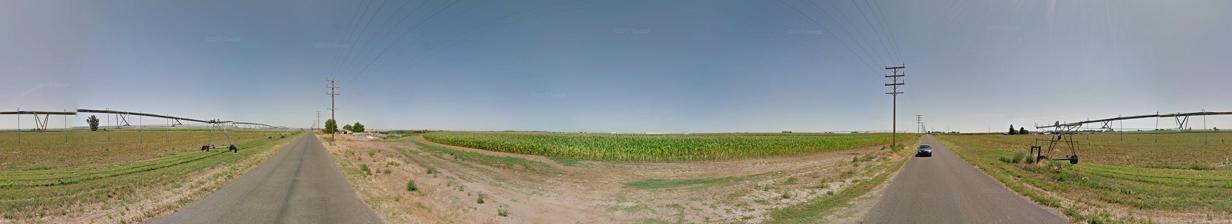}\vfill
	\vspace*{\gspace}
	\includegraphics[width=\textwidth]{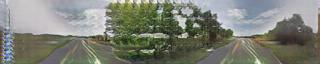}
	\end{minipage}\vspace*{\gspace}\hfill %
	\includegraphics[width=\awidth]{gan/aerial6}\hspace*{\gspace} %
	\begin{minipage}[b]{\gwidth}
	\includegraphics[width=\textwidth]{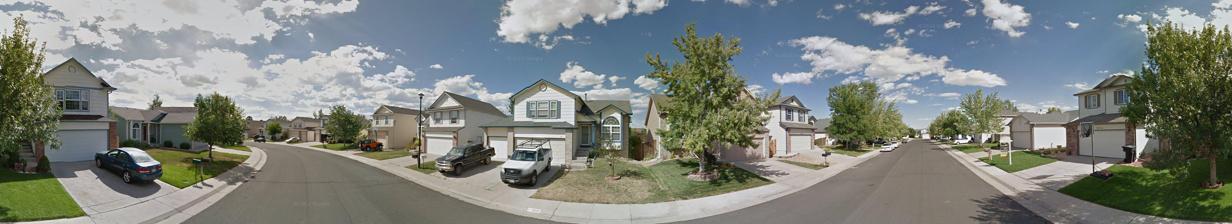}\vfill
	\vspace*{\gspace}
	\includegraphics[width=\textwidth]{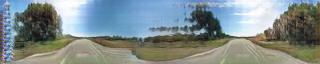}
	\end{minipage}\vspace*{\gspace}\hfill %
	
	\includegraphics[width=\awidth]{gan/aerial7}\hspace*{\gspace} %
	\begin{minipage}[b]{\gwidth}
	\includegraphics[width=\textwidth]{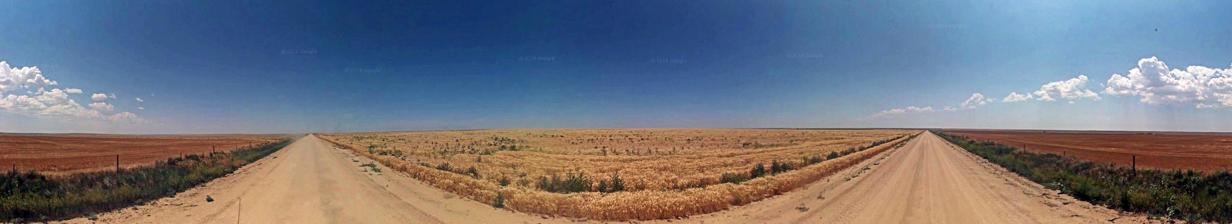}\vfill
	\vspace*{\gspace}
	\includegraphics[width=\textwidth]{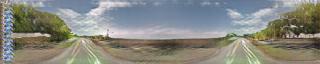}
	\end{minipage}\vspace*{\gspace}\hfill %
	\includegraphics[width=\awidth]{gan/aerial8}\hspace*{\gspace} %
	\begin{minipage}[b]{\gwidth}
	\includegraphics[width=\textwidth]{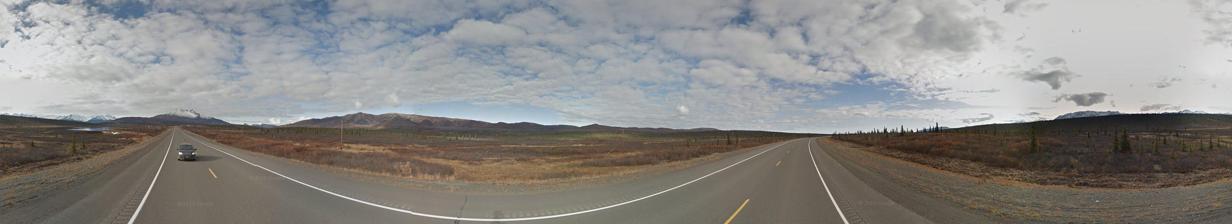}\vfill
	\vspace*{\gspace}
	\includegraphics[width=\textwidth]{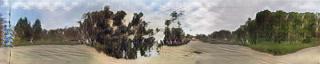}
	\end{minipage}\vspace*{\gspace}\hfill %
	\includegraphics[width=\awidth]{gan/aerial9}\hspace*{\gspace} %
	\begin{minipage}[b]{\gwidth}
	\includegraphics[width=\textwidth]{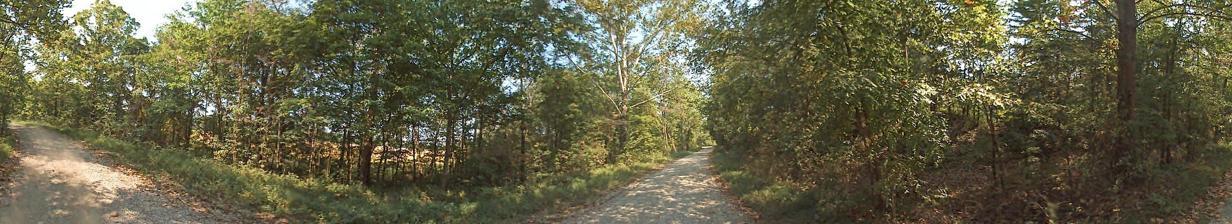}\vfill
	\vspace*{\gspace}
	\includegraphics[width=\textwidth]{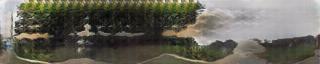}
	\end{minipage}\vspace*{\gspace}\hfill %
	
	\includegraphics[width=\awidth]{gan/aerial10}\hspace*{\gspace} %
	\begin{minipage}[b]{\gwidth}
	\includegraphics[width=\textwidth]{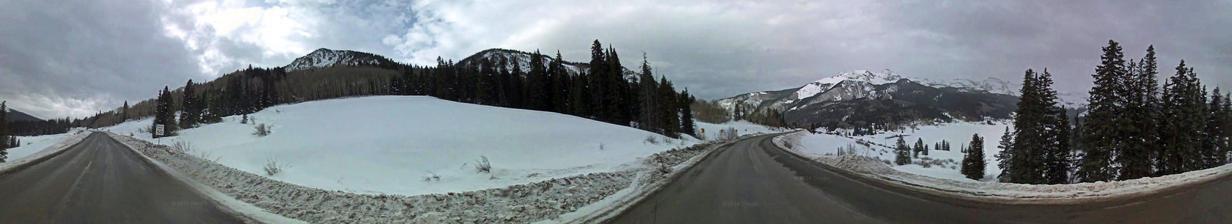}\vfill
	\vspace*{\gspace}
	\includegraphics[width=\textwidth]{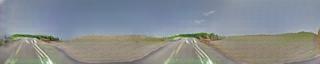}
	\end{minipage}\vspace*{\gspace}\hfill %
	\includegraphics[width=\awidth]{gan/aerial11}\hspace*{\gspace} %
	\begin{minipage}[b]{\gwidth}
	\includegraphics[width=\textwidth]{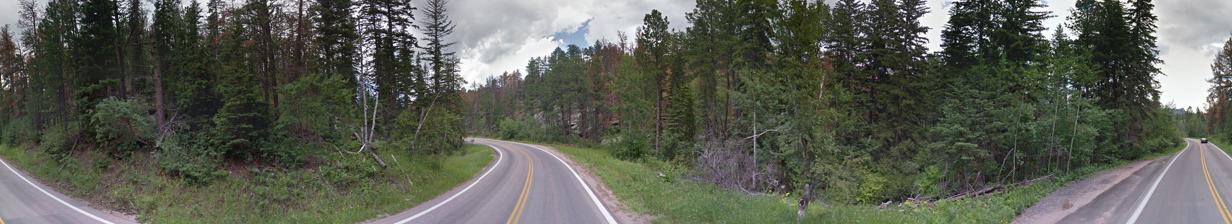}\vfill
	\vspace*{\gspace}
	\includegraphics[width=\textwidth]{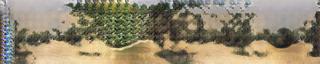}
	\end{minipage}\vspace*{\gspace}\hfill %
	\includegraphics[width=\awidth]{gan/aerial12}\hspace*{\gspace} %
	\begin{minipage}[b]{\gwidth}
	\includegraphics[width=\textwidth]{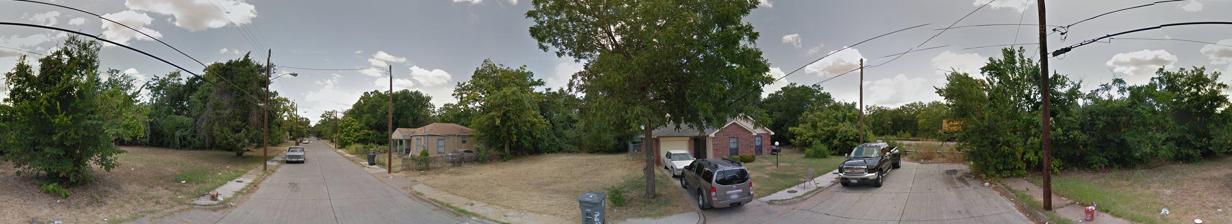}\vfill
	\vspace*{\gspace}
	\includegraphics[width=\textwidth]{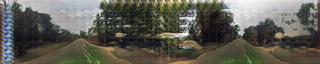}
	\end{minipage}\vspace*{\gspace}\hfill %
	
	\includegraphics[width=\awidth]{gan/aerial13}\hspace*{\gspace} %
	\begin{minipage}[b]{\gwidth}
	\includegraphics[width=\textwidth]{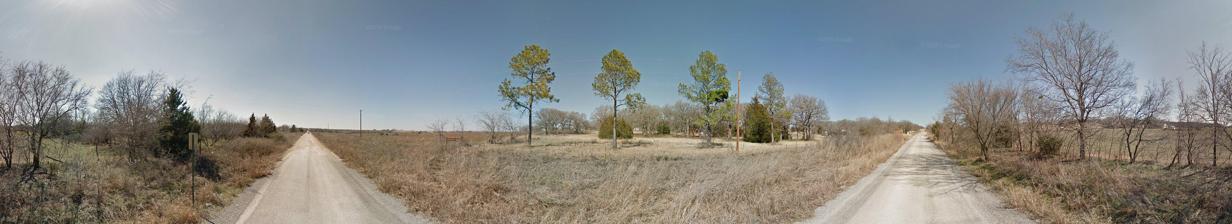}\vfill
	\vspace*{\gspace}
	\includegraphics[width=\textwidth]{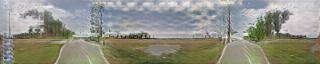}
	\end{minipage}\vspace*{\gspace}\hfill %
	\includegraphics[width=\awidth]{gan/aerial14}\hspace*{\gspace} %
	\begin{minipage}[b]{\gwidth}
	\includegraphics[width=\textwidth]{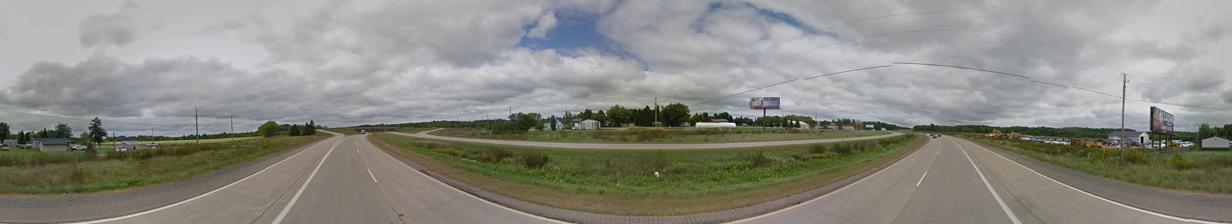}\vfill
	\vspace*{\gspace}
	\includegraphics[width=\textwidth]{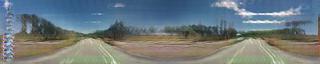}
	\end{minipage}\vspace*{\gspace}\hfill %
	\includegraphics[width=\awidth]{gan/aerial15}\hspace*{\gspace} %
	\begin{minipage}[b]{\gwidth}
	\includegraphics[width=\textwidth]{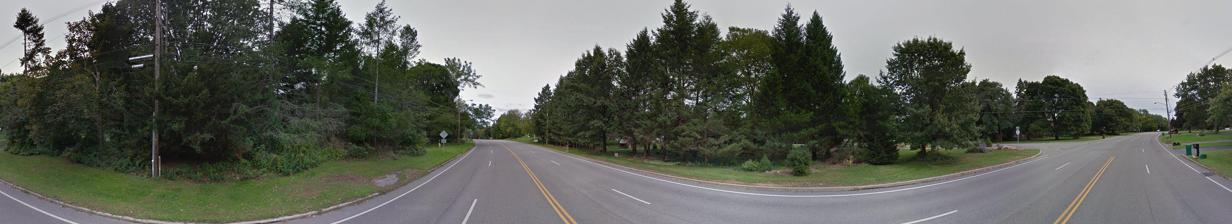}\vfill
	\vspace*{\gspace}
	\includegraphics[width=\textwidth]{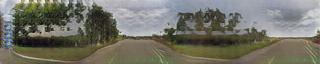}
	\end{minipage}\vspace*{\gspace}\hfill %
	
	\includegraphics[width=\awidth]{gan/aerial16}\hspace*{\gspace} %
	\begin{minipage}[b]{\gwidth}
	\includegraphics[width=\textwidth]{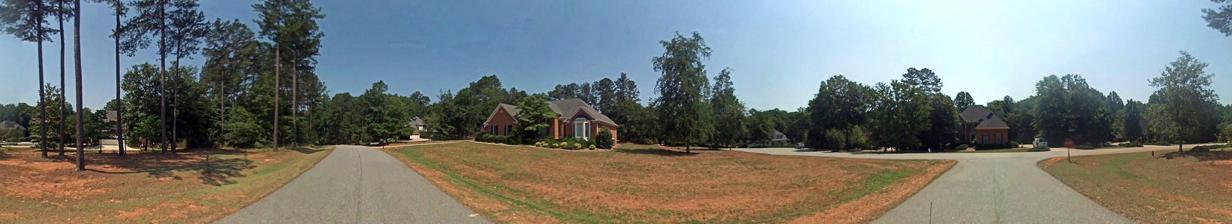}\vfill
	\vspace*{\gspace}
	\includegraphics[width=\textwidth]{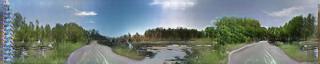}
	\end{minipage}\vspace*{\gspace}\hfill %
	\includegraphics[width=\awidth]{gan/aerial17}\hspace*{\gspace} %
	\begin{minipage}[b]{\gwidth}
	\includegraphics[width=\textwidth]{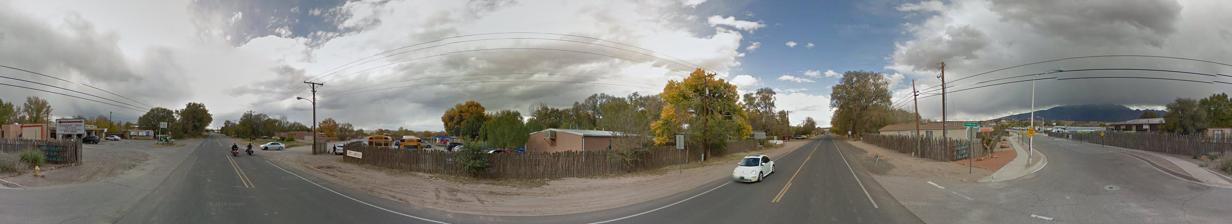}\vfill
	\vspace*{\gspace}
	\includegraphics[width=\textwidth]{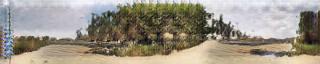}
	\end{minipage}\vspace*{\gspace}\hfill %
	\includegraphics[width=\awidth]{gan/aerial18}\hspace*{\gspace} %
	\begin{minipage}[b]{\gwidth}
	\includegraphics[width=\textwidth]{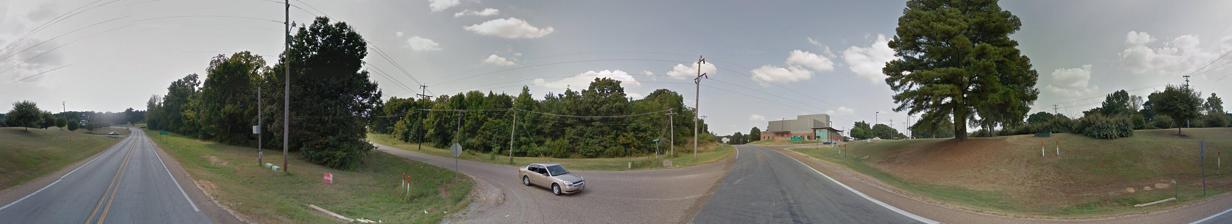}\vfill
	\vspace*{\gspace}
	\includegraphics[width=\textwidth]{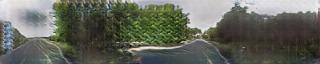}
	\end{minipage}\vspace*{\gspace}\hfill %
	
	\includegraphics[width=\awidth]{gan/aerial19}\hspace*{\gspace} %
	\begin{minipage}[b]{\gwidth}
	\includegraphics[width=\textwidth]{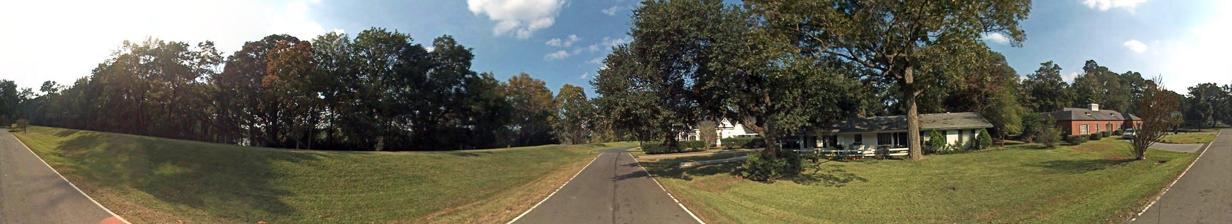}\vfill
	\vspace*{\gspace}
	\includegraphics[width=\textwidth]{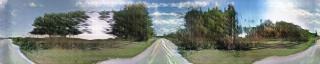}
	\end{minipage}\vspace*{\gspace}\hfill %
	\includegraphics[width=\awidth]{gan/aerial20}\hspace*{\gspace} %
	\begin{minipage}[b]{\gwidth}
	\includegraphics[width=\textwidth]{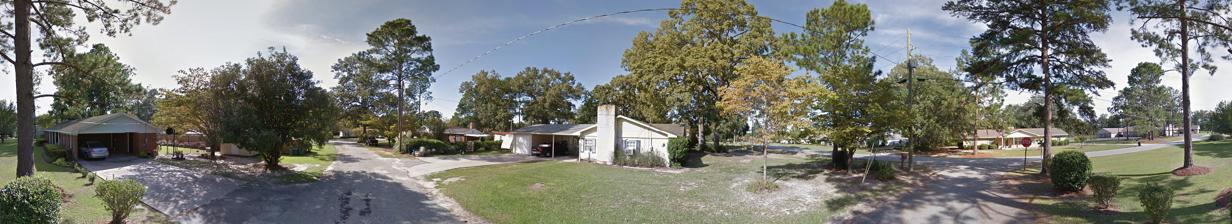}\vfill
	\vspace*{\gspace}
	\includegraphics[width=\textwidth]{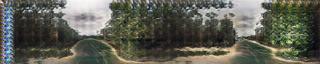}
	\end{minipage}\vspace*{\gspace}\hfill %
	\includegraphics[width=\awidth]{gan/aerial21}\hspace*{\gspace} %
	\begin{minipage}[b]{\gwidth}
	\includegraphics[width=\textwidth]{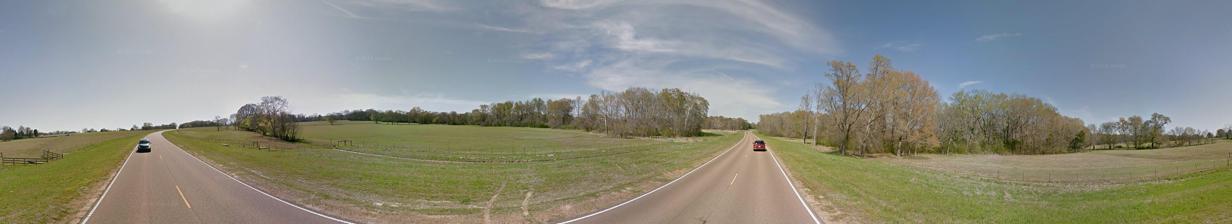}\vfill
	\vspace*{\gspace}
	\includegraphics[width=\textwidth]{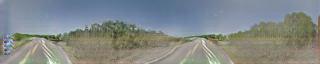}
	\end{minipage}\vspace*{\gspace}\hfill %
	
	\includegraphics[width=\awidth]{gan/aerial22}\hspace*{\gspace} %
	\begin{minipage}[b]{\gwidth}
	\includegraphics[width=\textwidth]{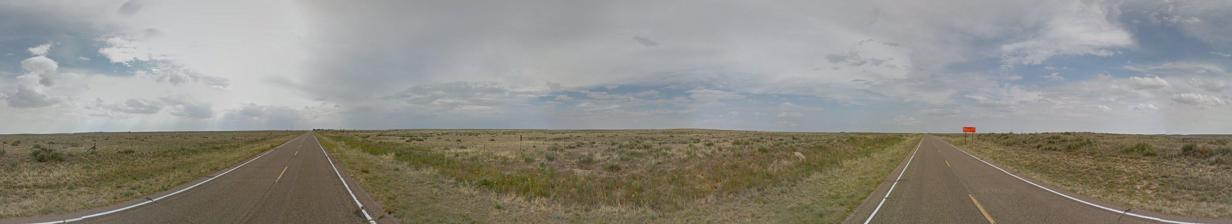}\vfill
	\vspace*{\gspace}
	\includegraphics[width=\textwidth]{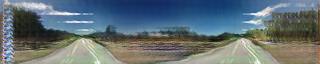}
	\end{minipage}\vspace*{\gspace}\hfill %
	\includegraphics[width=\awidth]{gan/aerial23}\hspace*{\gspace} %
	\begin{minipage}[b]{\gwidth}
	\includegraphics[width=\textwidth]{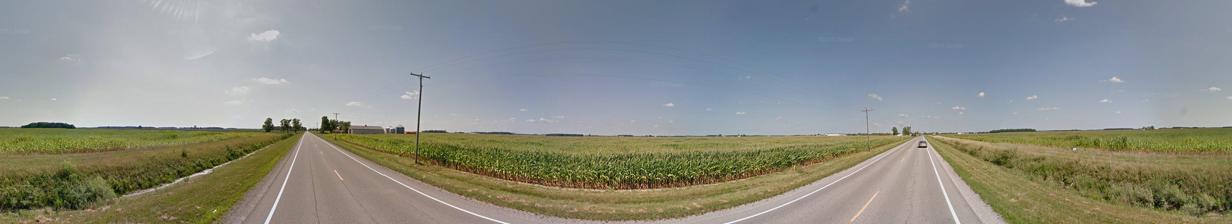}\vfill
	\vspace*{\gspace}
	\includegraphics[width=\textwidth]{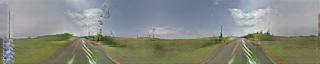}
	\end{minipage}\vspace*{\gspace}\hfill %
	\includegraphics[width=\awidth]{gan/aerial24}\hspace*{\gspace} %
	\begin{minipage}[b]{\gwidth}
	\includegraphics[width=\textwidth]{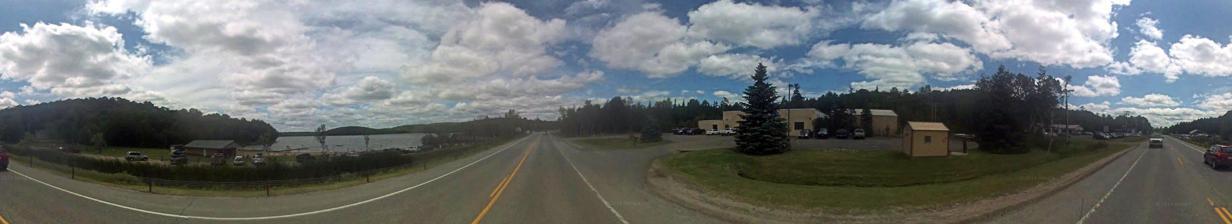}\vfill
	\vspace*{\gspace}
	\includegraphics[width=\textwidth]{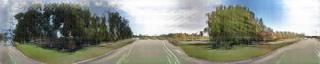}
	\end{minipage}\vspace*{\gspace}\hfill %
	
  \caption{Randomly sampled test image results for synthesizing
  ground-level views. Each row shows an aerial image (left), its
  corresponding ground-level panorama (top-right), and predicted
  ground-level panorama (bottom-right).}
  \label{fig:app:gan}
\end{figure*}

The appendix contains additional details and qualitative results from
our experiments. 
\figref{app:transfmat} demonstrates how we visualize the transformation
matrix in different ways. 
\figref{app:weakly} shows randomly selected qualitative results from our
weakly supervised learning task. 
\figref{app:geocali} shows additional fine-grained geocalibration results.
\tabref{dis} and \tabref{gen} describe the complete network structure
for the ground image synthesis application and additional qualitative
results are shown in \figref{app:gan}. %

\clearpage

\clearpage
\begin{table*}[!ht]
	
	\setlength\tabcolsep{3.4pt}
	\begin{minipage}{.48\linewidth} %
	\caption{Deep generator network architecture. All deconvolutions use a 
		stride of 2. $f$ is the extracted cross-view feature, and $z$ is 
		Gaussian noise.} %
	\begin{tabular}{llll} %
		\toprule
		\textbf{Input} & \textbf{Input Shape} & \textbf{Operation} & 
		\textbf{Output Shape} \\
		\midrule
		$f$ & 8 $\times$ 40 $\times$ 512 & 1 $\times$ 1 conv. & 8 
		$\times$ 40 $\times$ 448 \\
		$z$ & 8 $\times$ 40 $\times$ 64 & concat. & 8 $\times$ 40 $\times$ 
		512 \\
		\ & 8 $\times$ 40 $\times$ 512 & 5 $\times$ 5 deconv. & 16 $\times$ 80 
		$\times$ 
		256 \\
		\ & 16 $\times$ 80 $\times$ 256 & 5 $\times$ 5 deconv. & 32 $\times$ 
		160 
		$\times$ 128 \\
		\ & 32 $\times$ 160 $\times$ 128 & 5 $\times$ 5 deconv. & 64 $\times$ 
		320 
		$\times$ 64 \\
		\ & 64 $\times$ 320 $\times$ 64 & 1 $\times$ 1 conv. & 64 
		$\times$ 320 $\times$ 	32 \\
		\ & 64 $\times$ 320 $\times$ 32 & 1 $\times$ 1 conv. & 64 
		$\times$ 320 $\times$ 3 \\
		\bottomrule
	\end{tabular}%
	\label{tab:dis}
	\end{minipage} %
	\hspace{12pt} %
	\setlength\tabcolsep{5.65pt} %
	\begin{minipage}{0.48\textwidth} %
	\centering
	\caption{Deep energy network architecture. All 3 $\times$ 3 convolutions 
	use a stride of 2. $I_f = G(f, z)$, where $G$ is the deep generator and $f, 
	z$ are its parameters.}%
	\noindent \begin{tabular}{llll}%
		\toprule
		\textbf{Input} & \textbf{Input Shape} & \textbf{Operation} & 
		\textbf{Output Shape} \\
		\midrule
		$I_{f}$ & 64 $\times$ 320 $\times$ 3 & 3 $\times$ 3 conv. & 32 $\times$ 
		160 $\times$ 64 \\
		\ & 32 $\times$ 160 $\times$ 64 & 3 $\times$ 3 conv.& 16 $\times$ 80 
		$\times$ 128 \\
		\ & 16 $\times$ 80 $\times$ 128 & 3 $\times$ 3 conv.& 8 $\times$ 40 
		$\times$ 
		256 \\
		$f$ & 8 $\times$ 40 $\times$ 512 & concat. & 8 $\times$ 40 
		$\times$ 768 \\
		\ & 8 $\times$ 40 $\times$ 768 & 1 $\times$ 1 conv. & 8 
		$\times$ 40 $\times$ 32 \\
		\ & 8 $\times$ 40 $\times$ 32 & 1 $\times$ 1 conv. & 8 
		$\times$ 40 $\times$ 3 \\
		\ & 8 $\times$ 40 $\times$ 3 & energy term& 1 $\times$ 1 \\
				
		\bottomrule
	\end{tabular}
	\label{tab:gen}
	\end{minipage}
\end{table*} %

\end{document}